\documentclass[acmsmall,nonacm]{acmart}
\AtBeginDocument{%
  \providecommand\BibTeX{{%
    \normalfont B\kern-0.5em{\scshape i\kern-0.25em b}\kern-0.8em\TeX}}}

\usepackage[utf8]{inputenc} 
\usepackage[T1]{fontenc}    
\usepackage{url}            
\usepackage{booktabs}       
\usepackage{amsmath}
\usepackage{mathtools}
\usepackage{amsfonts}       
\usepackage{nicefrac}       
\usepackage{microtype}      
\usepackage{lipsum}
\usepackage{graphicx}
\usepackage{graphbox}
\usepackage{subfig}
\usepackage[ruled,vlined,linesnumbered]{algorithm2e}
\usepackage{multirow}
\usepackage{caption}
\usepackage{array}
\usepackage{pgfplots}
\usepackage{dsfont}
\usepackage{wrapfig}
\newcommand{\somecomment}[1]{}
\newcommand{\ccite}[1]{}
\newcommand\ddfrac[2]{\frac{\displaystyle #1}{\displaystyle #2}}
\newcolumntype{L}{>{\centering\arraybackslash}m{3cm}}

\DeclareMathOperator*{\argmin}{argmin}

\begin{document}

\title{Efficient Deep Learning: A Survey on Making Deep Learning Models Smaller, Faster, and Better}

\author{Gaurav Menghani}
\email{gmenghani@google.com}
\orcid{0000-0003-2912-2522}
\affiliation{%
  \institution{Google Research}
  \city{Mountain View}
  \state{California}
  \country{USA}
  \postcode{95054}
}


\begin{abstract}
  Deep Learning has revolutionized the fields of computer vision, natural language understanding, speech recognition, information retrieval and more. However, with the progressive improvements in deep learning models, their number of parameters, latency, resources required to train, etc. have all have increased significantly. Consequently, it has become important to pay attention to these footprint metrics of a model as well, not just its quality. We present and motivate the problem of efficiency in deep learning, followed by a thorough survey of the five core areas of model efficiency (spanning modeling techniques, infrastructure, and hardware) and the seminal work there. We also present an experiment-based guide along with code, for practitioners to optimize their model training and deployment. We believe this is the first comprehensive survey in the efficient deep learning space that covers the landscape of model efficiency from modeling techniques to hardware support. Our hope is that this survey would provide the reader with the mental model and the necessary understanding of the field to apply generic efficiency techniques to immediately get significant improvements, and also equip them with ideas for further research and experimentation to achieve additional gains.
\end{abstract}




\maketitle

\section{Introduction}
Deep Learning with neural networks has been the dominant methodology of training new machine learning models for the past decade. Its rise to prominence is often attributed to the ImageNet competition \cite{Deng2009} in 2012. That year, a University of Toronto team submitted a deep convolutional network (AlexNet \cite{Krizhevsky2012}, named after the lead developer Alex Krizhevsky), performed 41\% better than the next best submission. As a result of this trailblazing work, there was a race to create deeper networks with an ever increasing number of parameters and complexity. Several model architectures such as VGGNet \cite{Simonyan2014}, Inception \cite{Szegedy2015}, ResNet \cite{He2016} etc. successively beat previous records at ImageNet competitions in the subsequent years, while also increasing in their footprint (model size, latency, etc.)
 
\begin{figure}%
    \centering
    \subfloat[\centering Computer Vision Models]{{\includegraphics[width=7.5cm]{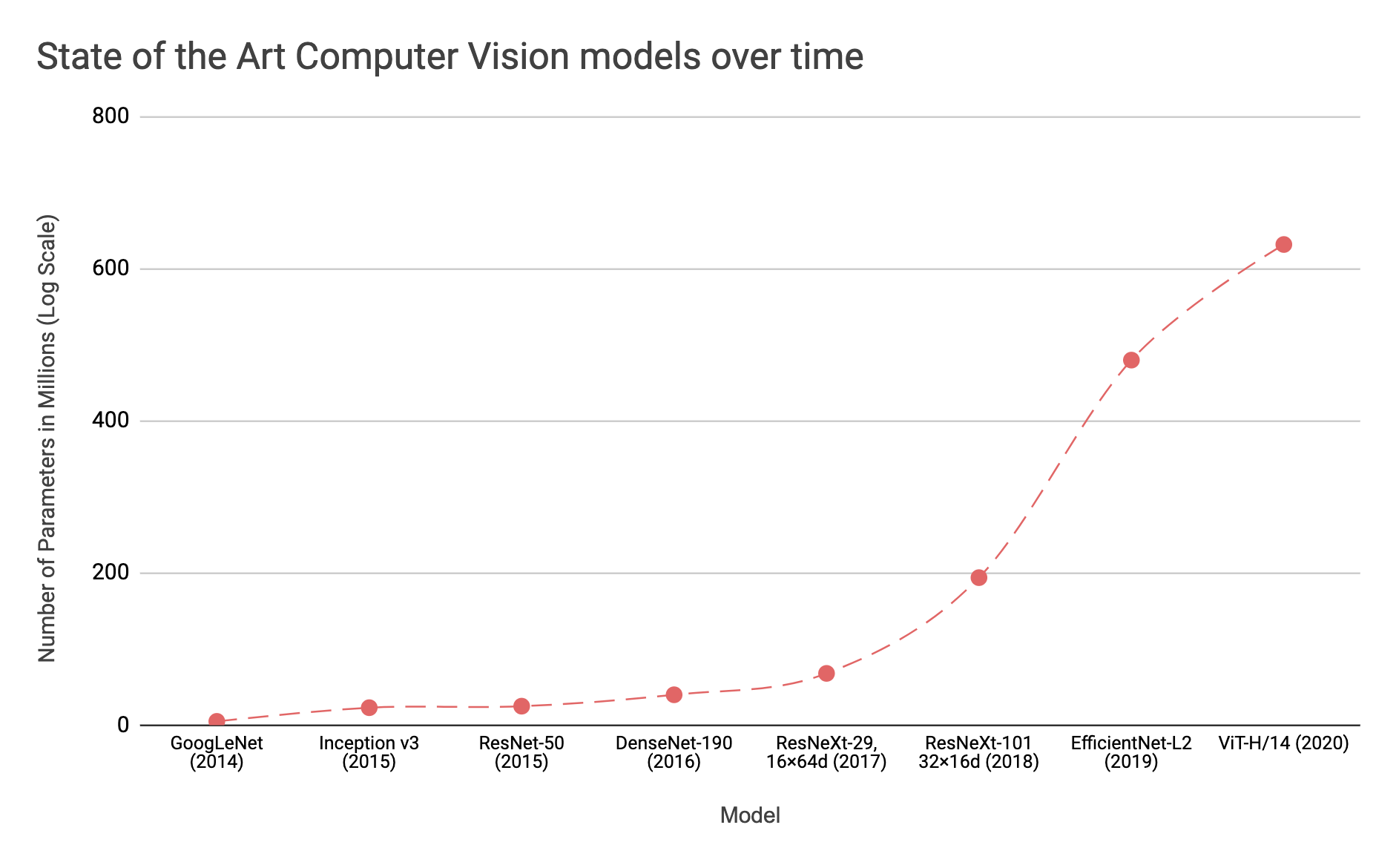} }}%
    \qquad
    \subfloat[\centering Natural Language Models]{{\includegraphics[width=7.5cm]{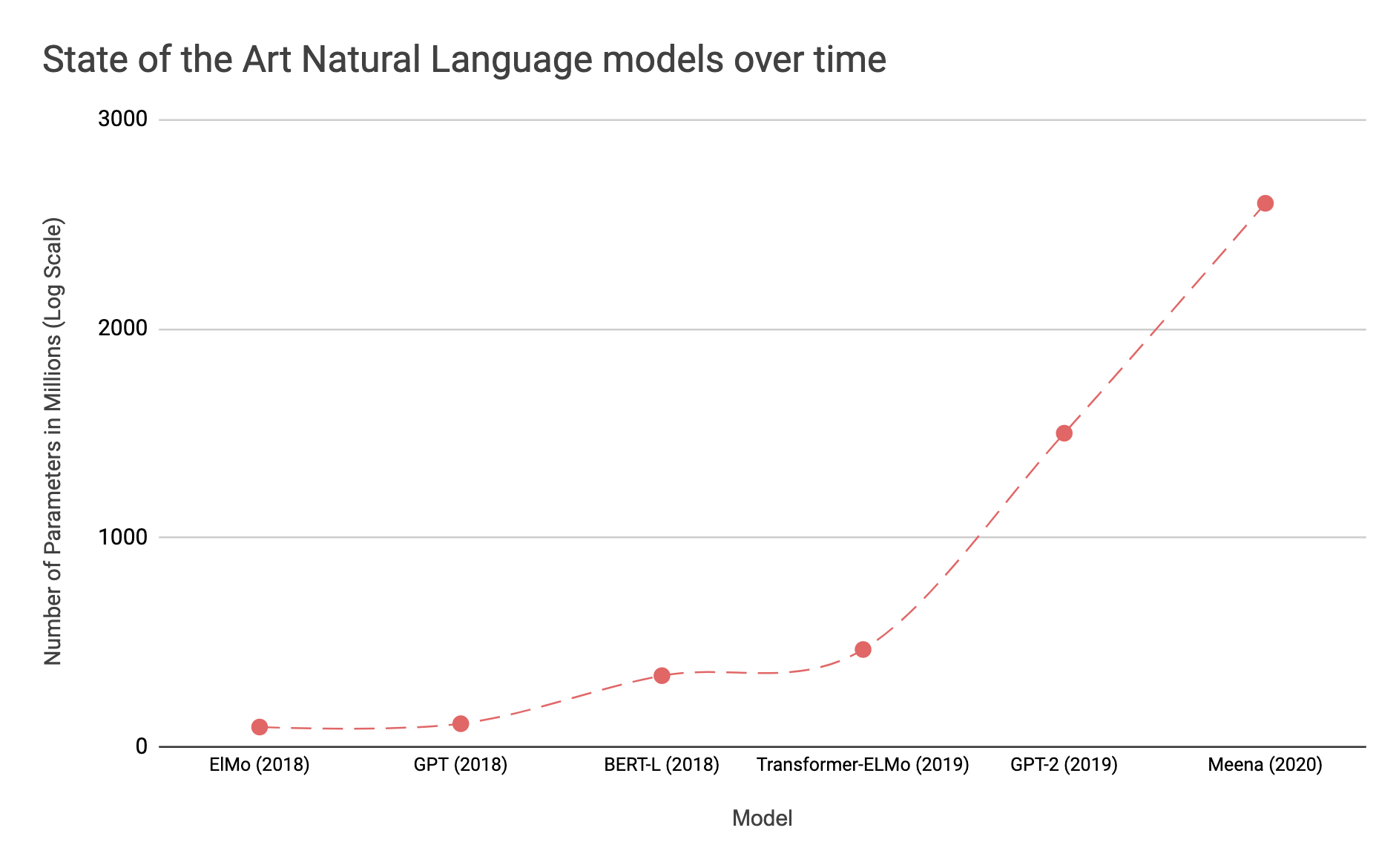} }}%
    \caption{Growth in the number of parameters in Computer Vision models over time. \cite{PapersWithCode2021}}%
    \label{fig:parameter-growth}%
\end{figure}

This effect has also been noted in natural language understanding (NLU), where the Transformer \cite{Vaswani2017} architecture based on primarily Attention layers, spurred the development of general purpose language encoders like BERT \cite{Devlin2018}, GPT-3 \cite{Brown2020}, etc. BERT specifically beat 11 NLU benchmarks when it was published. GPT-3 has also been used in several places in the industry via its API. The common aspect amongst these domains is the rapid growth in the model footprint (Refer to Figure \ref{fig:parameter-growth}), and the cost associated with training and deploying them.

Since deep learning research has been focused on improving the state of the art, progressive improvements on benchmarks like image classification, text classification, etc. have been correlated with an increase in the network complexity, number of parameters, the amount of training resources required to train the network, prediction latency, etc. For instance, GPT-3 comprises of 175 billion parameters, and costs millions of dollars to train just one iteration (\cite{Brown2020}). This excludes the cost of experimentation / trying combinations of different hyper-parameters, which is also computationally expensive.

While these models perform well on the tasks they are trained on, they might not necessarily be efficient enough for direct deployment in the real world. A deep learning practitioner might face the following challenges when training or deploying a model.

\begin{itemize}
\item \textbf{Sustainable Server-Side Scaling}: Training and deploying large deep learning models is costly. While training could be a one-time cost (or could be free if one is using a pre-trained model), deploying and letting inference run for over a long period of time could still turn out to be expensive in terms of consumption of server-side RAM, CPU, etc.. There is also a very real concern around the carbon footprint of datacenters even for organizations like Google, Facebook, Amazon, etc. which spend several billion dollars each per year in capital expenditure on their data-centers.

\item \textbf{Enabling On-Device Deployment}: Certain deep learning applications need to run realtime on IoT and smart devices (where the model inference happens directly on the device), for a multitude of reasons (privacy, connectivity, responsiveness). 
Thus, it becomes imperative to optimize the models for the target devices.

\item \textbf{Privacy \& Data Sensitivity}: Being able to use as little data as possible for training is critical when the user-data might be sensitive. Hence, efficiently training models with a fraction of the data means lesser data-collection required.

\item \textbf{New Applications}: Certain new applications offer new constraints (around model quality or footprint) that existing off-the-shelf models might not be able to address.

\item \textbf{Explosion of Models}: While a singular model might work well, training and/or deploying multiple models on the same infrastructure (colocation) for different applications might end up exhausting the available resources.
\end{itemize}

\subsection{Efficient Deep Learning}
The common theme around the above challenges is \emph{efficiency}. We can break it down further as follows:

\begin{itemize}
\item \textbf{Inference Efficiency}:  This primarily deals with questions that someone deploying a model for inference (computing the model outputs for a given input), would ask. Is the model small? Is it fast, etc.? More concretely, how many parameters does the model have, what is the disk size, RAM consumption during inference, inference latency, etc.

\item \textbf{Training Efficiency}: This involves questions someone training a model would ask, such as How long does the model take to train? How many devices? Can the model fit in memory?, etc. It might also include questions like, how much data would the model need to achieve the desired performance on the given task?
\end{itemize}

If we were to be given two models, performing equally well on a given task, we might want to choose a model which does better in either one, or ideally both of the above aspects. If one were to be deploying a model on devices where inference is constrained (such as mobile and embedded devices), or expensive (cloud servers), it might be worth paying attention to inference efficiency. Similarly, if one is training a large model from scratch on either with limited or costly training resources, developing models that are designed for training efficiency would help. 

\begin{figure}[h]
  \centering
  \includegraphics[width=5.5cm]{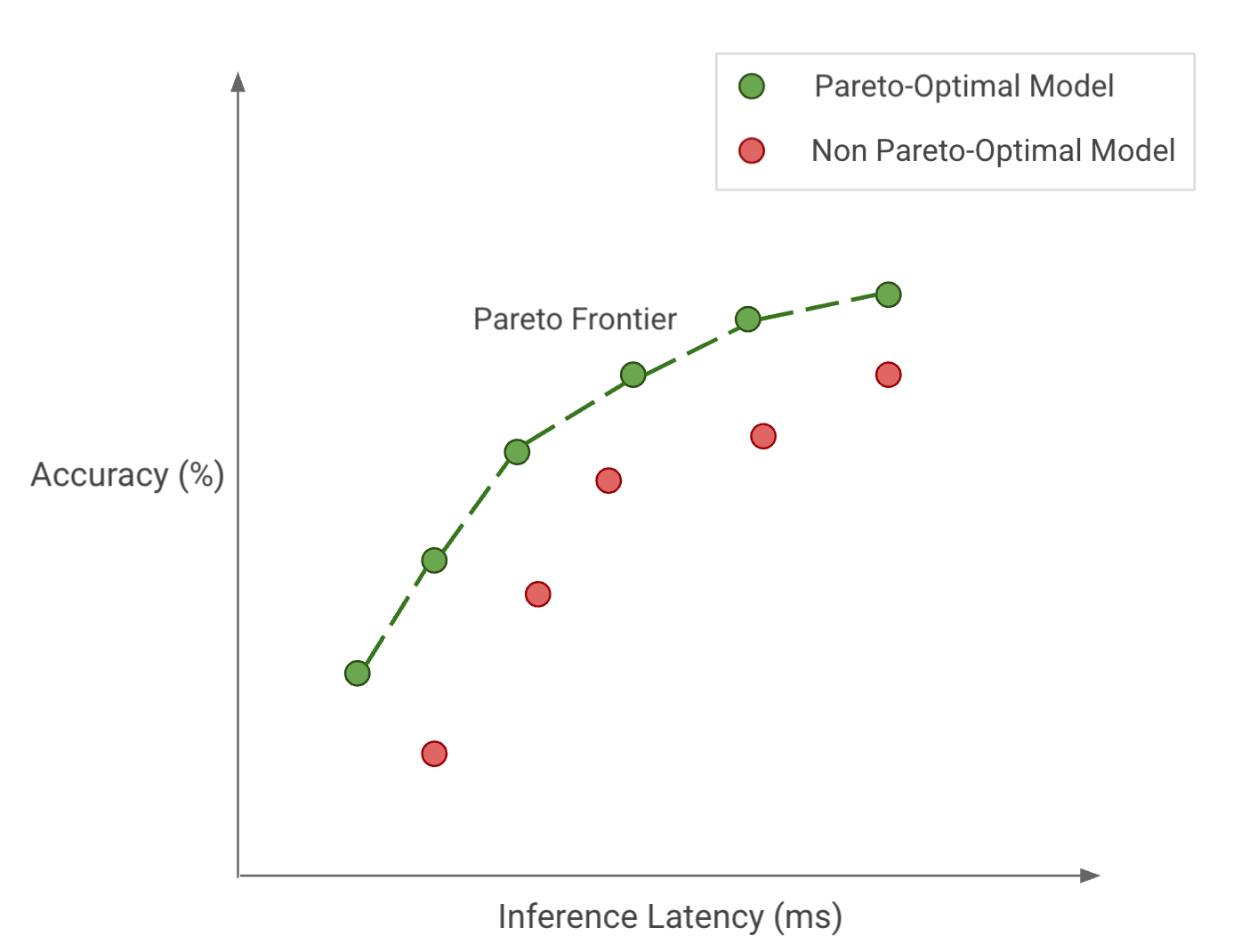}
  \caption{Pareto Optimality: Green dots represent pareto-optimal models (together forming the pareto-frontier), where none of the other models (red dots) get better accuracy with the same inference latency, or the other way around.}
  \label{fig:pareto-optimality}
\end{figure}

Regardless of what one might be optimizing for, we want to achieve \emph{pareto-optimality}. This implies that any model that we choose is the best for the tradeoffs that we care about. As an example in Figure \ref{fig:pareto-optimality}, the green dots represent pareto-optimal models, where none of the other models (red dots) get better accuracy with the same inference latency, or the other way around. Together, the pareto-optimal models (green dots) form our \emph{pareto-frontier}. The models in the pareto-frontier are by definition more efficient than the other models, since they perform the best for their given tradeoff. Hence, when we seek efficiency, we should be thinking about discovering and improving on the pareto-frontier.

To achieve this goal, we propose turning towards a collection of algorithms, techniques, tools, and infrastructure that work together to allow users to train and deploy \emph{pareto-optimal} models with respect to model quality and its footprint.

\section{A Mental Model}
In this section we present the mental model to think about the collection of algorithms, techniques, and tools related to efficient deep learning. We propose to structure them in five major areas, with the first four focused on modeling, and the final one around infrastructure and tools.

\begin{figure}[h]
  \centering
  \includegraphics[width=9cm]{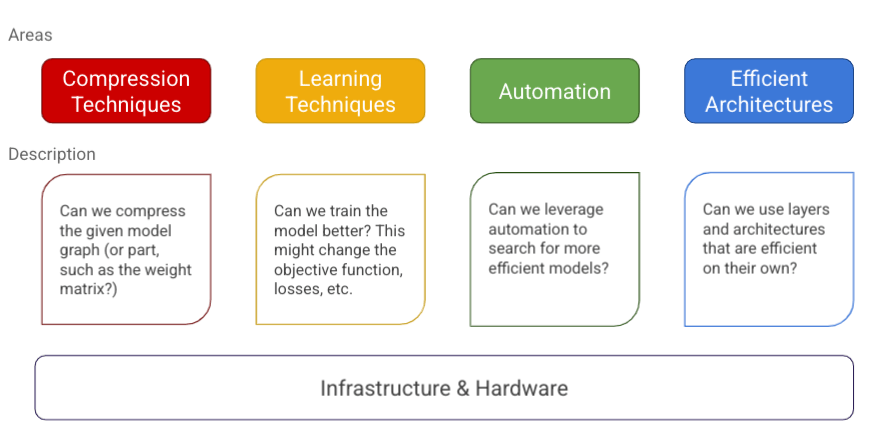}
  \caption{A mental model for thinking about algorithms, techniques, and tools related to efficiency in Deep Learning.}
  \label{efficiency-areas}
\end{figure}

\begin{enumerate}
\item \textbf{Compression Techniques}: These are general techniques and algorithms that look at optimizing the model's architecture, typically by compressing its layers. A classical example is quantization \cite{Jacob2018}, which tries to compress the weight matrices of a layer, by reducing its precision (eg., from 32-bit floating point values to 8-bit unsigned integers), with minimal loss in quality.

\item \textbf{Learning Techniques}: These are algorithms which focus on training the model differently (to make fewer prediction errors, require less data, converge faster, etc.). The improved quality can then be exchanged for a smaller footprint / a more efficient model by trimming the number of parameters if needed. An example of a learning technique is distillation \cite{Hinton2015}, which allows improving the accuracy of a smaller model by learning to mimic a larger model.

\item \textbf{Automation}: These are tools for improving the core metrics of the given model using automation. An example is hyper-parameter optimization (HPO) \cite{Golovin2017} where optimizing the hyper-parameters helps increase the accuracy, which could then be then exchanged for a model with lesser parameters. Similarly, architecture search \cite{Zoph2016} falls in this category too, where the architecture itself is tuned and the search helps find a model that optimizes both the loss / accuracy, and some other metric such as model latency, model size, etc. 

\item \textbf{Efficient Architectures}: These are fundamental blocks that were designed from scratch (convolutional layers, attention, etc.), that are a significant leap over the baseline methods used before them (fully connected layers, and RNNs respectively). As an example, convolutional layers introduced parameter sharing for use in image classification, which avoids having to learn separate weights for each input pixel, and also makes them robust to overfitting. Similarly, attention layers \cite{Bahdanau2014} solved the problem of Information Bottleneck in Seq2Seq models. These architectures can be used directly for efficiency gains. 

\item \textbf{Infrastructure}: Finally, we also need a foundation of infrastructure and tools that help us build and leverage efficient models. This includes the model training framework, such as Tensorflow \cite{Abadi2016}, PyTorch \cite{Paszke2019}, etc. (along with the tools required specifically for deploying efficient models such as Tensorflow Lite (TFLite), PyTorch Mobile, etc.). We depend on the infrastructure and tooling to leverage gains from efficient models. For example, to get both size and latency improvements with quantized models, we need the inference platform to support common neural network layers in quantized mode.
\end{enumerate}

We will survey each of these areas in depth in the following section.

\section{Landscape of Efficient Deep Learning}
\subsection{Compression Techniques}
Compression techniques as mentioned earlier, are usually generic techniques for achieving a more efficient representation of one or more layers in a neural network, with a possible quality trade off. The efficiency goal could be to optimize the model for one or more of the footprint metrics, such as model size, inference latency, training time required for convergence, etc. in exchange for as little quality loss as possible. In some cases if the model is over-parameterized, these techniques can improve model generalization.

\subsubsection{Pruning}
Given a neural network $f(X, W)$, where $X$ is the input and $W$ is the set of parameters (or weights), pruning is a technique for coming up with a minimal subset $W'$ such that the rest of the parameters of $W$ are pruned (or set to 0), while ensuring that the quality of the model remains above the desired threshold. After pruning, we can say the network has been made \emph{sparse}, where the sparsity can be quantified as the ratio of the number of parameters that were pruned to the number of parameters in the original network ($s = (1 - \frac{|W'|}{|W|})$). The higher the sparsity, the lesser the number of non-zero parameters in the pruned networks.

\begin{figure}[h]
  \centering
  \includegraphics[width=7.5cm]{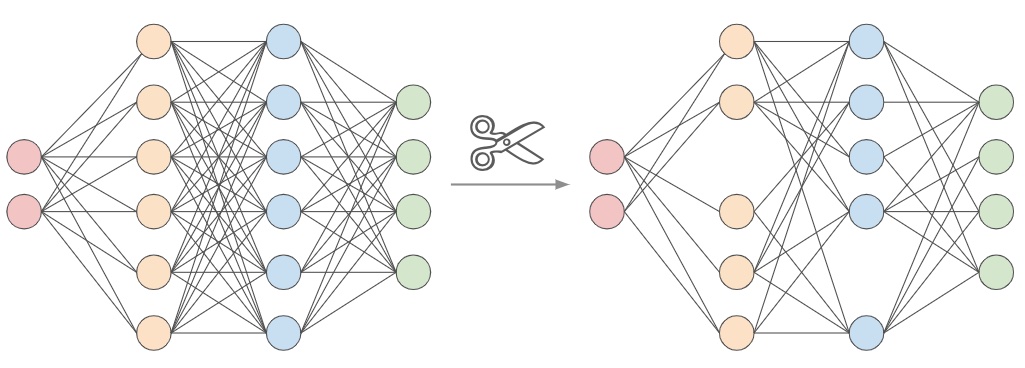}
  \caption{A simplified illustration of pruning weights (connections) and neurons (nodes) in a neural network comprising of fully connected layers.}
  \label{simple-pruning}
\end{figure}

Some of the classical works in this area are Optimal Brain Damage (OBD) by LeCun et al. \cite{Lecun1990}, and Optimal Brain Surgeon paper (OBD) by Hassibi et al. \cite{Hassibi1993}. These methods usually take a network that has been pre-trained to a reasonable quality and then iteratively prune the parameters which have the lowest `saliency' score, such that the impact on the validation loss is minimized. Once pruning concludes, the network is fine-tuned with the remaining parameters. This process is repeated a number of times until the desired number of original parameters are pruned (Algorithm~\ref{algo:pruning}).

\begin{algorithm}[H]
\small
\SetAlgoLined
\KwData{Pre-trained dense network with weights $W$, inputs $X$, number of pruning rounds $N$, fraction of parameters to prune per round $p$.}
\KwResult{Pruned network with weights $W'$.}
$W' \gets W$\;
\For{$i \gets 1$ \textbf{to} $N$} {
  $S \gets \texttt{compute\_saliency\_scores}(W')$\;

  $W' \gets W' - \texttt{select\_min\_k}\large(S, \frac{|W'|}{p}\large)$\;
  $W' \gets $\texttt{fine\_tune}($X$, $W'$)
}
return $W'$
 \caption{Standard Network Pruning with Fine-Tuning}
 \label{algo:pruning}
\end{algorithm}

OBD approximates the saliency score by using a second-derivative of the parameters ($\large \frac{\partial^2 L}{\partial w_{i}^2}$), where $L$ is the loss function, and $w_{i}$ is the candidate parameter for removal. The intuition is that the higher this value for a given parameter, the larger the change in the loss function's gradient if it were to be pruned.

For the purpose of speeding up the computation of the second-derivatives, OBD ignores cross-interaction between the weights ($\large \frac{\partial^2 L}{\partial w_{i} \partial w_{j}}$), and hence computes only the diagonal elements of the Hessian matrix. Otherwise, computing the full Hessian matrix is unwieldy for even a reasonable number of weights (with $n=10^4$, the size of the matrix is $10^4 \times 10^4 = 10^8$). In terms of results, LeCun et al. demonstrate that pruning reduced the parameters in a well-trained neural net by ~ 8x (combination of both automatic and manual removal) without a drop in classification accuracy.  

Across different pruning strategies, the core algorithm could remain similar, with changes in the following aspects.

\begin{itemize}
    \item \textbf{Saliency}: While \cite{Lecun1990, Hassibi1993} use second-order derivatives, other methods rely on simpler magnitude based pruning \cite{Han2015a, Han2015b}, or momentum based pruning \cite{Dettmers2019} etc. to determine the saliency score.
    \item \textbf{Structured v/s Unstructured}: The most flexible way of pruning is unstructured (or random) pruning, where all given parameters are treated equally. In structured pruning, parameters are pruned in blocks (such as pruning row-wise in a weight matrix, or pruning channel-wise in a convolutional filter \cite{Li2016a,Anwar2017,Molchanov2016,Liu2019}, etc.). The latter allows easier leveraging of inference-time gains in size and latency, since these blocks of pruned parameters can be intelligently skipped for storage and inference. Note that unstructured pruning can also be viewed as structured pruning with block size = 1.
    
    
    \item \textbf{Distribution}: The decision about how to distribute the sparsity budget (number of parameters to be pruned), could be made either by pooling in all the parameters from the network and then deciding which parameters to prune, or by smartly selecting how much to prune in each layer individually \cite{Dong2017, He2018}. \cite{Elsen2020,google-research2021GitHubFastConvnets} have found that some architectures like MobileNetV2, EfficientNet \cite{Tan2019} have thin first layers that do not contribute significantly to the number of parameters and pruning them leads to an accuracy drop without much gain. Hence, intuitively it would be helpful to allocate sparsity on a per-layer basis.  
    \item \textbf{Scheduling}: Another question is how much to prune, and when? Should we prune an equal number of parameters every round \cite{Lecun1990, Hassibi1993, Han2015a}, or should we prune at a higher pace in the beginning and gradually decrease \cite{Zhu2018, Dettmers2019}.
    \item \textbf{Regrowth}: Some methods allow regrowing pruned connections \cite{Evci2020, Dettmers2019} to keep the same level of sparsity through constant cycles of prune-redistribute-regrow. Dettmers et al. \cite{Dettmers2019} estimate training time speedups between 2.7x - 5.6x by starting and operating with a sparse model throughout. However there is a gap in terms of implementation of sparse operations on CPU, GPU, and other hardware. 
\end{itemize}

\renewcommand{\arraystretch}{1.0}
\begin{table}[]
\small
\begin{tabular}{llllll}
\hline
\multicolumn{1}{c}{Model Architecture} & Sparsity Type & Sparsity \% & FLOPs & Top-1 Accuracy \% & Source         \\ \hline
\multirow{6}{*}{MobileNet v2 - 1.0}      
 & Dense (Baseline) & 0\%  & 1x    & 72.0\% & Sandler et al. \cite{Sandler2018} \\ \cline{2-6} 
 & Unstructured & 75\% & 0.27x & 67.7\% & Zhu et al. \cite{Zhu2018} \\ \cline{2-6} 
 & Unstructured & 75\% & 0.52x & 71.9\% & Evci et al. \cite{Evci2020} \\ \cline{2-6} 
 & Structured (block-wise) & 85\% & 0.11x & 69.7\% & Elsen et al. \\ \cline{2-6} 
 & Unstructured & 90\% & 0.12x & 61.8\% & Zhu et al. \cite{Zhu2018} \\ \cline{2-6} 
 & Unstructured & 90\% & 0.12x & 69.7\% & Evci et al. \cite{Evci2020} \\ \hline
\end{tabular}
\caption{A sample of various sparsity results on the MobileNet v2 architecture with depth multiplier = 1.0.}
\label{tab:mnv2-sparsity}
\end{table}

\textbf{Beyond Model Optimization}: Frankle et al.’s \cite{Frankle2018} work on the Lottery Ticket Hypothesis took a different look at pruning, and postulated that within every large network lies a smaller network, which can be extracted with the original initialization of its parameters, and retrained on its own to match or exceed the performance of the larger network. The authors demonstrated these results on multiple datasets, but others such as \cite{Gale2019,Liu2018a} were not able to replicate this on larger datasets such as ImageNet \cite{Deng2009}. Rather Liu et al. \cite{Liu2018a} demonstrate that the pruned architecture with random initialization does no worse than the pruned architecture with the trained weights.

\textbf{Discussion}: There is a significant body of work that demonstrates impressive theoretical reduction in the model size (via number of parameters), or estimates the savings in FLOPs (Table~\ref{tab:mnv2-sparsity}). However, a large fraction of the results are on \emph{unstructured} pruning, where it is not currently clear how these improvements can lead to reduction in footprint metrics (apart from using standard file compression tools like GZip).

On the other hand, structured pruning with a meaningful block size is conducive to latency improvements. Elsen et al. \cite{Elsen2020,google-research2021GitHubFastConvnets} construct sparse convolutional networks that outperform their dense counterparts by $1.3$ - $2.4 \times$ with $\approx$ 66\% of the parameters, while retaining the same Top-1 accuracy. They do this via their library to convert from the NHWC (channels-last) standard dense representation to a special NCHW (channels-first) `Block Compressed Sparse Row' (BCSR) representation which is suitable for fast inference using their fast kernels on ARM devices, WebAssembly etc. \cite{XNNPACKAuthors2021a}. Although they also introduce some constraints on the kinds of sparse networks that can be accelerated \cite{XNNPACKAuthors2021b}. Overall, this is a promising step towards practical improvements in footprint metrics with pruned networks.

\subsubsection{Quantization}
Almost all the weights and activations of a typical network are in 32-bit floating-point values. One of the ideas of reducing model footprint is to reduce the precision for the weights and activations by \emph{quantizing} to a lower-precision datatype (often 8-bit fixed-point integers). There are two kinds of gains that we can get from quantization: (a) lower model size, and (b) lower inference latency. Often, only the model size is a constraint, and in this case we can employ a technique called weight quantization and get model size improvements \cite{TensorflowAuthors2021f}, where only the model weights are in reduced precision. In order to get latency improvements, the activations need to be in fixed-point as well (Activation Quantization \cite{Vanhoucke2011,Jacob2018}, such that all the operations in the quantized graph are happening in fixed-point math as well.

\begin{figure}[h]
  \centering
  \includegraphics[width=8cm]{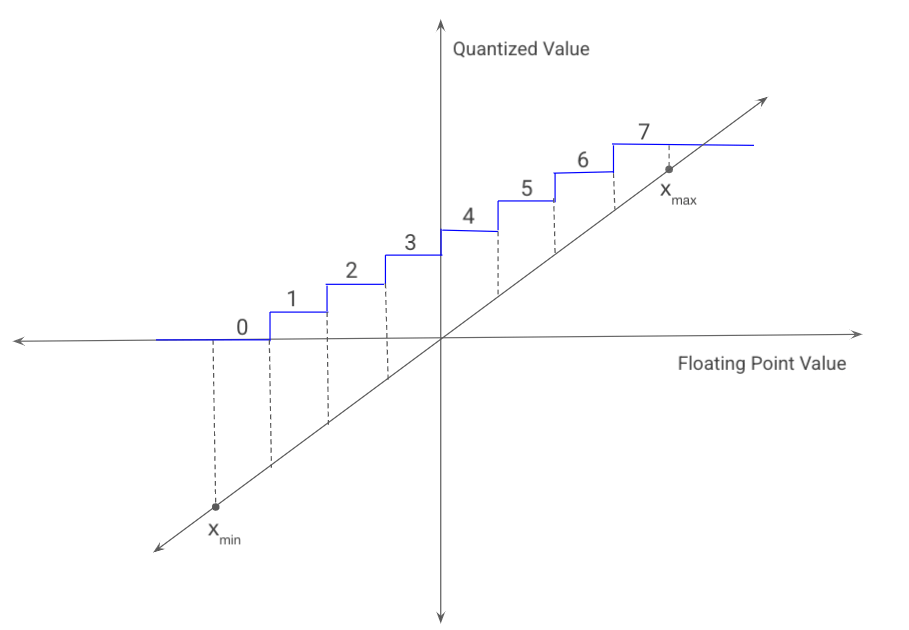}
  \caption{Quantizing floating-point continuous values to discrete fixed-point values. The continuous values are clamped to the range $x_{min}$ to $x_{max}$, and are mapped to discrete values in [$0$, $2^b - 1$] (in the above figure, $b = 3$, hence the quantized values are in the range [$0, 7$].}
  \label{fig:quantization}
\end{figure}

\textbf{Weight Quantization}: A simple \emph{scheme} for quantizing weights to get model size improvements (similar to \cite{Krishnamoorthi2018}) is as follows. Given a 32-bit floating-point weight matrix in a model, we can map the minimum weight value ($x_{min}$) in that matrix to $0$, and the maximum value ($x_{max}$) to $2^{b}-1$ (where $b$ is the number of bits of precision, and $b < 32$). Then we can linearly extrapolate all values between them to an integer value in [$0, 2^{b}-1$] (Figure ~\ref{fig:quantization}). Thus, we are able to map each floating point value to a fixed-point value where the latter requires a lesser number of bits than the floating-point representation. This process can also be done for signed $b$-bit fixed-point integers, where the output values will be in the range [-$2^{\frac{b}{2}} -1$, $2^{\frac{b}{2}} -1$]. One of the reasonable values of $b$ is $8$, since this would lead to a $32 / 8 = 4\times$ reduction in space, and also because of the near-universal support for \texttt{uint8\_t} and \texttt{int8\_t} datatypes.

During inference, we go in the reverse direction where we recover a lossy estimate of the original floating point value (\emph{dequantization}) using just the $x_{min}$ and $x_{max}$. This estimate is lossy since we lost $32 - b$ bits of information when did the rounding (another way to look at it is that a range of floating point values map to the same quantized value).

\cite{Jacob2018,Krishnamoorthi2018} formalize the quantization scheme with the following two constraints:
\begin{itemize}
    \item The quantization scheme should be linear (affine transformation), so that the precision bits are linearly distributed.
    \item $0.0$ should map exactly to a fixed-point value $x_{q_0}$, such that dequantizing $x_{q_0}$ gives us $0.0$. This is an implementation constraint, since $0$ is also used for padding to signify missing elements in tensors, and if dequantizing $x_{q_0}$ leads to a non-zero value, then it might be interpreted incorrectly as a valid element at that index.
\end{itemize}

The second constraint described above requires that $0$ be a part of the quantization range, which in turn requires updating $x_{min}$ and $x_{max}$, followed by clamping $x$ to lie in $[x_{min}, x_{max}]$. Following this, we can quantize $x$ by constructing a piece-wise linear transformation as follows:

\begin{equation}
\small
\label{eq:quantization-std}
    \textrm{quantize}(x) = x_q = \textrm{round}\bigg(\frac{x}{s}\bigg) + z
\end{equation}

$s$ is the floating-point \emph{scale} value (can be thought of as the inverse of the slope, which can be computed using $x_{min}$, $x_{max}$ and the range of the fixed-point values). $z$ is an integer \emph{zero-point} value which is the quantized value that is assigned to $x = 0.0$. This is the terminology followed in literature \cite{Jacob2018,Krishnamoorthi2018} (Algorithm \ref{algo:quantization}).

The dequantization step constructs $\hat{x}$, which is a lossy estimate of $x$, since we lose precision when quantizing to a lower number of bits. We can compute it as follows:

\begin{equation}
\small
\label{eq:dequantization-std}
    \textrm{dequantize}(x_q) = \hat{x} = s (x_q - z)
\end{equation}

 Since $s$ is in floating-point, $\hat{x}$ is also a floating-point value (Algorithm \ref{algo:dequantization}). Note that the quantization and dequantization steps can be performed for signed integers too by appropriately changing the value $x_{q_{min}}$ (which is the lowest fixed-point value in $b$-bits) in Algorithm~\ref{algo:quantization}.

\begin{minipage}[t]{7.5cm}
\null 
 \begin{algorithm}[H]
    \small
    \SetAlgoLined
    \KwData{Floating-point tensor to compress $\mathbf{X}$, number of precision bits $b$ for the fixed-point representation.}
    \KwResult{Quantized tensor $\mathbf{X_q}$.}
    $\textbf{X}_{min}, \textbf{X}_{max} \gets \textrm{min}(\mathbf{X}, 0), \textrm{max}(\mathbf{X}, 0)$\;
    $\mathbf{X} \gets \textrm{clamp}(\mathbf{X}, \textbf{X}_{min}, \textbf{X}_{max})$\;
    $s \gets \ddfrac{x_{max} - x_{min}}{2^b - 1}$\;
    $z \gets \textrm{round}\bigg(x_{q_{min}} - \ddfrac{x_{min}}{s}\bigg)$\;

    $\mathbf{X_q} \gets \textrm{round}\bigg(\ddfrac{\mathbf{X}}{s}\bigg) + z$\;
    return $\mathbf{X_q}$\;
    \caption{Quantizing a given weight matrix $\mathbf{X}$}
    \label{algo:quantization}
    \end{algorithm}
    \end{minipage}%
\hfill
\begin{minipage}[t]{6.5cm}
 \null
 \begin{algorithm}[H]
    \small
    \SetAlgoLined
    \KwData{Fixed-point matrix to dequantize $\mathbf{X_q}$, along with the scale $s$, and zero-point $z$ values which were calculated during quantization.} 
    \KwResult{Dequantized floating-point weight matrix $\widehat{\mathbf{X}}$.}
    $\widehat{\mathbf{X}} \gets s(\mathbf{X_q} - z)$\;
    return $\widehat{\mathbf{X}}$\;
    \caption{Dequantizing a given fixed-point weight matrix $\mathbf{X_q}$}
    \label{algo:dequantization}
  \end{algorithm}
\end{minipage}

We can utilize the above two algorithms for quantizing and dequantizing the model's weight matrices. Quantizing a pre-trained model's weights for reducing the size is termed as \emph{post-training quantization} in literature \cite{TensorflowAuthors2021f}. This might be sufficient for the purpose of reducing the model size when there is sufficient representational capacity in the model.

There are other works in literature \cite{Rastegari2016, Hubara2016, Li2016b} that demonstrate slightly different variants of quantization. XNOR-Net \cite{Rastegari2016}, Binarized Neural Networks \cite{Hubara2016} and others use $b=1$, and thus have weight matrices which just have two possible values $0$ or $1$, and the quantization function there is simply the $\textrm{sign}(x)$ function (assuming the weights are symmetrically distributed around $0$).

The promise with such extreme quantization approaches is the theoretical $32 / 1 = 32\times$ reduction in model size without much quality loss. Some of the works claim improvements on larger networks like AlexNet \cite{Krizhevsky2012}, VGG \cite{Simonyan2014}, Inception \cite{Szegedy2015} etc., which might already be more amenable to compression. A more informative task would be to demonstrate extreme quantization on smaller networks like the MobileNet family \cite{Sandler2018, Howard2019}. Additionally binary quantization (and other quantization schemes like ternary \cite{Li2016b}, bit-shift based networks \cite{Rastegari2016}, etc.) promise latency-efficient implementations of standard operations where multiplications and divisions are replaced by cheaper operations like addition, subtraction, etc. These claims need to be verified because even if these lead to theoretical reduction in FLOPs, the implementations still need support from the underlying hardware. A fair comparison would be using standard quantization with $b=8$, where the multiplications and divisions also become cheaper, and are supported by the hardware efficiently via SIMD instructions which allow for low-level data parallelism (for example, on x86 via the SSE instruction set, on ARM via the Neon \cite{Arm2021} intrinsics, and even on specialized DSPs like the Qualcomm Hexagon \cite{XNNPACKAuthors2021b}).

\textbf{Activation Quantization}: To be able to get \emph{latency improvements} with quantized networks, the math operations have to be done in fixed-point representations too. This means all intermediate layer inputs and outputs are also in fixed-point, and there is no need to dequantize the weight-matrices since they can be used directly along with the inputs.

Vanhoucke et al. \cite{Vanhoucke2011} demonstrated a $3 \times$ inference speedup using a fully fixed-point model on an x86 CPU, when compared to a floating-point model on the same CPU, without sacrificing accuracy. The weights are still quantized similar to post-training quantization, however all layer inputs (except the first layer) and the activations are fixed-point. In terms of performance, the primary driver for this improvement was the availability of fixed-point SIMD instructions in Intel's SSE4 instruction set \cite{ContributorstoWikimediaprojects2021b}, where commonly used building-block operations like the Multiply-Accumulate (MAC) \cite{Wikimedia2021MAC} can be parallelized. Since the paper was published, Intel has released two more iterations of these instruction sets \cite{ContributorstoWikimediaprojects2021c} which might further improve the speedups. 


\textbf{Quantization-Aware Training (QAT)}: The network that Vanhoucke et al. mention was a 5 layer feed-forward network that was post-training quantized. However post-training quantization can lead to quality loss during inference as highlighted in \cite{Krishnamoorthi2018,Jacob2018,Wang2020} as the networks become more complex. These could be because of: (a) outlier weights that skew the computation of the quantized values for the entire input range towards the outliers, leading to less number of bits being allocated to the bulk of the range, or (b) Different distribution of weights within the weight matrix, for eg. within a convolutional layer the distribution of weights between each filter might be different, but they are quantized the same way. These effects might be more pronounced at low-bit widths due to an even worse loss of precision. Wang et al. \cite{Wang2020} try to retain the post-training quantization but with new heuristics to allocate the precision bits in a learned fashion. Tools like the TFLite Converter \cite{TensorFlow2019PostTraining} augment post-training quantization with a representative dataset provided by the user, to actively correct for errors at different points in the model by comparing the error between the activations of the quantized and unquantized graphs.

Jacob et al. \cite{Jacob2018} propose (and further detailed by Krishnamoorthi et al. \cite{Krishnamoorthi2018}) a training regime which is \emph{quantization-aware}. In this setting, the training happens in floating-point but the forward-pass simulates the quantization behavior during inference. Both weights and activations are passed through a function that simulates this quantization behavior (\emph{fake-quantized} is the term used by many works \cite{Jacob2018,Krishnamoorthi2018}).

\begin{figure}%
    \centering
    \subfloat[\centering Quantization-Aware Training]{{\includegraphics[width=4.5cm]{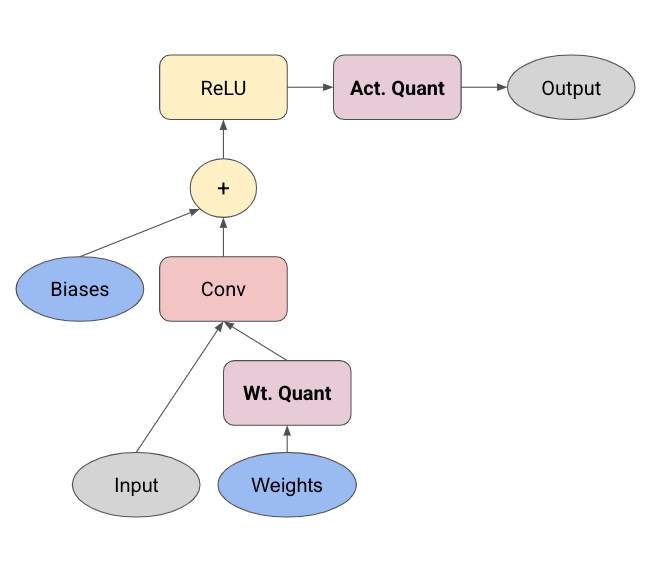} }}%
    \qquad
    \subfloat[\centering Final fixed-point inference graph]{{\includegraphics[width=4.5cm]{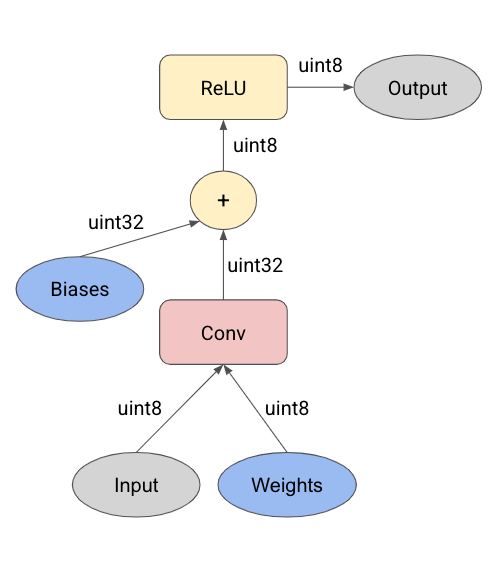} }}%
    \caption{(a) shows the injection of fake-quantization nodes to simulate quantization effect and collecting tensor statistics, for exporting a fully fixed-point inference graph. (b) shows the inference graph derived from the same graph as (a). Inputs and weights are in \texttt{uint8}, and results of common operations are in \texttt{uint32}. Biases are kept in \texttt{uint32} \cite{Jacob2018,Krishnamoorthi2018}}.%
    \label{fig:quantization-graphs}%
\end{figure}

Assuming $\mathbf{X}$ is the tensor to be fake-quantized, Jacob et al. \cite{Jacob2018} propose adding special quantization nodes in the training graph that collect the statistics (moving averages of $x_{min}$ and $x_{max}$) related to the weights and activations to be quantized (see Figure \ref{fig:quantization-graphs}(a) for an illustration). Once we have these values for each $\mathbf{X}$, we can derive the respective $\widehat{\mathbf{X}}$ using equations (\ref{eq:quantization-std} and \ref{eq:dequantization-std}) as follows. 

\begin{equation}
\small
\begin{split}
    \widehat{\mathbf{X}}{}& = \textrm{FakeQuant}(\mathbf{X}) \\
    & = \textrm{Dequantize}(\textrm{Quantize}(\mathbf{X})) \\
    & = s((\textrm{round}\bigg(\ddfrac{\textrm{clamp}(\mathbf{X}, x_{min}, x_{max})}{s}\bigg) + z) - z) \\
    & = s\bigg(\textrm{round}\bigg(\ddfrac{\textrm{clamp}(\mathbf{X}, x_{min}, x_{max})}{s}\bigg)\bigg) \\
\end{split}
\label{eq:fake-quant}
\end{equation}

Since the above equation is not directly differentiable because of the rounding behavior, to optimize a loss function $L$ w.r.t. $\mathbf{X}$, we can compute $\ddfrac{\partial{L}}{\partial{\mathbf{X}}}$ by chain-rule using the Straight-Through Estimator (STE) \cite{Bengio2013}. This allows us to make the staircase function differentiable with a linear approximation (See \cite{Krishnamoorthi2018} for details).

Quantization-Aware Training allows the network to adapt to tolerate the noise introduced by the clamping and rounding behavior during inference. Once the network is trained, tools such as the TFLite Model Converter \cite{TensorflowAuthors2021c} can generate the appropriate fixed-point inference model from a network annotated with the quantization nodes.

\textbf{Other Notable Works}:
Polino et al. \cite{Polino2018} allow non-uniform distribution of precision with learning a vector of quantization-points $p$, along with using distillation to further reduce loss of accuracy. The results for simpler datasets like CIFAR-10 are comparable to \cite{Krishnamoorthi2018,Jacob2018}. However, when working with ResNet architecture on the ImageNet dataset, they achieve lower model size and faster inference by using shallower student networks. This is not a fair comparison, since other works do not mix distillation along with quantization. Fan et al. \cite{Fan2020} demonstrate accuracy improvement on top of standard QAT (\cite{Jacob2018}) with $b < 8$. They hypothesize that the networks will learn better if the fake-quantization is not applied to the complete tensor at the same time to allow unbiased gradients to flow (instead of the STE approximation). Instead, they apply the fake-quantization operation stochastically in a block-wise manner on the given tensor. They also demonstrate improvements over QAT on 4-bit quantized Transformer and EfficientNet \cite{Tan2019} networks.

\textbf{Results}:
Refer to Table \ref{tab:mnv2-quantization} for a comparison between the baseline floating-point model, post-training quantized, and quantization-aware trained models \cite{TensorflowAuthors2021f}. The model with post-training quantization gets close to the baseline, but there is still a significant accuracy difference. The model size is $4 \times$ smaller, however the latency is slightly higher due to the need to dequantize the weights during inference. The model with 8-bit Quantization-Aware Training (QAT) gets quite close to the baseline floating point model while requiring $4 \times$ less disk space and being $1.64 \times$ faster.

\renewcommand{\arraystretch}{1.0}
\begin{table}[]
\small
\begin{tabular}{llllll}
\hline
Model Architecture                      & Quantization Type & Top-1 Accuracy & Size (MB) & Latency (ms, Pixel2) \\ \hline
\multirow{3}{*}{MobileNet v2-1.0 (224)} & Baseline          & 71.9\%         & 14        & 89                       \\ \cline{2-5} 
 & Post-Training Quantization  & 63.7\% & 3.6 & 98 \\ \cline{2-5} 
 & Quantization-Aware Training & 70.9\% & 3.6 & 54 \\ \hline
\end{tabular}
\caption{
A sample of various quantization results on the MobileNet v2 architecture for 8-bit quantization \cite{Tensorflow2021MOT}. We picked results on 8-bit, since from they can be readily used with hardware and software that exists today.
}
\label{tab:mnv2-quantization}
\end{table}

\textbf{Discussion}:
\begin{itemize}
    \item Quantization is a well-studied technique for model optimization and can help with very significant reduction in model size (often $4 \times$ when using 8-bit quantization) and inference latency.
    \item Weight quantization is straight-forward enough that it can be implemented by itself for reducing model size. Activation quantization should be strongly considered because it enables both latency reduction, as well as lower working memory required for intermediate computations in the model (which is essential for devices with low memory availability)
    \item When possible, Quantization-Aware Training should be used. It has been shown to dominate post-training quantization in terms of accuracy.
    \item However, tools like Tensorflow Lite have made it easy to rely on post-training quantization. \cite{TensorFlow2019PostTraining} shows that often there is minimal loss when using post-training quantization, and with the help of a representative dataset this is further shrunk down. Wherever there is an opportunity for switching to fixed-point operations, the infrastructure allows using them.
    \item For performance reasons, it is best to consider the common operations that follow a typical layer such as Batch-Norm, Activation, etc. and `fold' them in the quantization operations.
\end{itemize}

\subsubsection{Other Compression Techniques}
There are other compression techniques like Low-Rank Matrix Factorization, K-Means Clustering, Weight-Sharing etc. which are also actively being used for model compression \cite{Panigrahy2021} and might be suitable for further compressing hotspots in a model. 

\subsection{Learning Techniques}
Learning techniques try to train a model differently in order to obtain better quality metrics (accuracy, F1 score, precision, recall, etc.) while allowing supplementing, or in some cases replacing the traditional supervised learning. The improvement in quality can sometimes be traded off for a smaller footprint by reducing the number of parameters / layers in the model and achieving the same baseline quality with a smaller model. An incentive of paying attention to learning techniques is that they are applied only on the training, without impacting the inference.

\subsubsection{Distillation} 
Ensembles are well known to help with generalization \cite{Krogh1994,Hansen1990}. The intuition is that this enables learning multiple independent hypotheses, which are likely to be better than learning a single hypothesis. \cite{Dietterich2000} goes over some of the standard ensembling methods such as bagging (learning models that are trained on non-overlapping data and then ensembling them), boosting (learning models that are trained to fix the classification errors of other models in the ensemble), averaging (voting by all the ensemble models), etc.Bucila et al. \cite{Bucilua2006} used large ensembles to label synthetic data that they generated using various schemes. A smaller neural net is then trained to learn not just from the labeled data but also from this weakly labeled synthetic data. They found that single neural nets were able to mimic the performance of larger ensembles, while being $1000 \times$ smaller and faster. This demonstrated that it is possible to transfer the cumulative knowledge of ensembles to a single small model. Though it might not be sufficient to rely on just the existing labeled data. 

Hinton et al. \cite{Hinton2015}, in their seminal work explored how smaller networks (students) can be taught to extract `dark knowledge' from larger models / ensembles of larger models (teachers) in a slightly different manner. Instead of having to generate synthetic-data, they use the larger teacher model to generate \emph{soft-labels} on existing labeled data. The soft-labels assign a probability to each class, instead of hard binary values in the original data. The intuition is that these soft-labels capture the relationship between the different classes which the model can learn from. For example, a truck is more similar to a car than to an apple, which the model might not be able to learn directly from hard labels.

The student network learns to minimize the cross-entropy loss on these soft labels, along with the original ground-truth hard labels. Since the probabilities of the incorrect classes might be very small, the logits are scaled down by a `temperature' value $\geq 1.0$, so that the distribution is `softened'. If the input vector is $\mathbf{X}$, and the teacher model's logits are $\mathbf{Z^{(t)}}$, the teacher model's softened probabilities with temperature $T$ can be calculated as follows using the familiar softmax function:

\begin{equation}
    \small
    \mathbf{Y}_i^{(t)} = \ddfrac{\exp(\mathbf{Z_i^{(t)}} / T)}{\sum_{j=1}^{n} \exp(\mathbf{Z_j^{(t)}} / T)}
\end{equation}

Note that as $T$ increases, the relative differences between the various elements of $Y^{(t)}$ decreases. This happens because if all elements are divided by the same constant, the softmax function would lead to a larger drop for the bigger values. Hence, as the temperature $T$ increases, we see the distribution of $Y^{(t)}$ `soften' further.

When training along with labeled data ($\mathbf{X}$, $\mathbf{Y}$), and the student model's output ($\mathbf{Y^{(s)}}$), we can describe the loss function as:

\begin{equation}
    \small
    \begin{split}
        L& = \lambda_1 \cdot L_{\rm ground-truth} + \lambda_2 \cdot L_{\rm distillation} \\
        & = \lambda_1 \cdot \textrm{CrossEntropy}(\mathbf{Y}, \mathbf{Y^{(s)}}; \theta) + \lambda_2 \cdot \textrm{CrossEntropy}(\mathbf{Y^{(t)}}, \mathbf{Y^{(s)}}; \theta)
    \end{split}
\end{equation}

\begin{figure}[h]
  \centering
  \includegraphics[width=10cm]{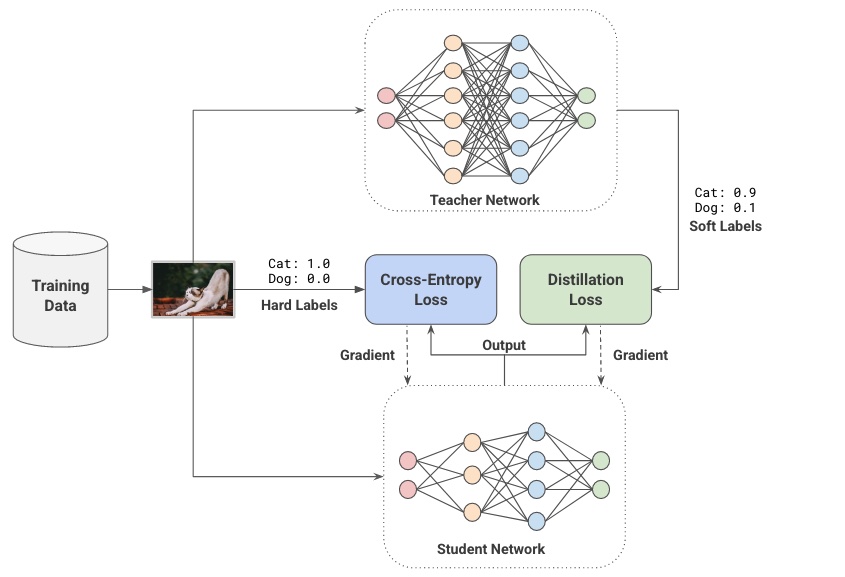}
  \caption{Distillation of a smaller student model from a larger pre-trained teacher model. Both the teacher and student models receive the same input. The teacher is used to generate `soft-labels' for the student, which gives the student more information than just hard binary labels. The student is trained using the regular cross-entropy loss with the hard labels, as well as using the distillation loss function which uses the soft labels from the teacher. In this setting, the teacher is frozen, and only the student receives the gradient updates. }
  \label{fig:distillation}
\end{figure}

$\textrm{CrossEntropy}$ is the cross-entropy loss function, which takes in the labels and the output. For the first loss term, we pass along the ground truth labels, and for the second loss term we pass the corresponding soft labels from the teacher model for the same input. $\lambda_1$ and $\lambda_2$ control the relative importance of the standard ground truth loss and the distillation loss respectively. When $\lambda_1 = 0$, the student model is trained with just the distillation loss. Similarly, when $\lambda_2 = 0$, it is equivalent to training with just the ground-truth labels. Usually, the teacher network is pre-trained and frozen during this process, and only the student network is updated. Refer to Figure \ref{fig:distillation} for an illustration of this process.

In the paper, Hinton et al. \cite{Hinton2015} were able to closely match the accuracy of a 10 model ensemble for a speech recognition task with a single distilled model. Urban et al. \cite{Urban2016} did a comprehensive study demonstrating that distillation significantly improves performance of shallow student networks as small as an MLP with one hidden layer on tasks like CIFAR-10. Sanh et al. \cite{Sanh2019} use the distillation loss for compressing a BERT \cite{Devlin2018} model (along with a cosine loss that minimizes the cosine distance between two internal vector representation of the input as seen by the teacher and student models). Their model retains 97\% of the performance of BERT-Base while being 40\% smaller and 60\% faster on CPU. 

It is possible to adapt the general idea of distillation to work on intermediate outputs of teachers and students. Zagoruyko et al. \cite{Zagoruyko2016} transfer intermediate `attention maps' between teacher and student convolutional networks. The intuition is to make the student focus on the parts of the image where the teacher is paying attention to. MobileBERT \cite{Sun2020} uses a progressive-knowledge transfer strategy where they do layer-wise distillation between the BERT student and teacher models, but they do so in stages, where the first $l$ layers are distilled in the $l$-th stage. Along with other architecture improvements, they obtain a 4.3$\times$ smaller and 5.5$\times$ faster BERT with small losses in quality. 

Another idea that has been well explored is exploiting a model trained in a supervised training to label unlabeled data. Blum et al. \cite{Blum1998} in their paper from 1998, report halving the error rate of their classifiers by retraining on a subset of pseudo-labels generated using the previous classifiers. This has been extended through distillation to use the teacher model to label a large corpus of unlabeled data, which can then be used to improve the quality of the student model \cite{Menghani2019,Xie2020,Yalniz2019}.  


Overall, distillation has been empirically shown to improve both the accuracy as well as the speed of convergence of student models across many domains. Hence, it enables training smaller models which might otherwise not be have an acceptable quality for deployment.

\textbf{Discussion}:
\begin{itemize}
    \item Distillation is an adaptable technique that needs minimal changes in the training infrastructure to be used. Even if the teacher model cannot be executed at the same time as the student model, the teacher model's predictions can be collected offline and treated as another source of labels.
    \item When there is sufficient label data, there is ample evidence that distillation is likely to improve the student model's predictions. If there is a large corpus of unlabeled data, the teacher model can be used to generate pseudo-labels on the unlabeled data, which can further improve the student model's accuracy.
    \item Strategies for intermediate-layer distillation have also shown to be effective in the case of complex networks. In such scenarios, a new loss term minimizing the difference between the outputs of the two networks at some semantically identical intermediate point(s) needs to be added.  
\end{itemize}

\somecomment{
Source: https://github.com/dkozlov/awesome-knowledge-distillation

1. Caruana
1. Hinton
1. DistillBERT
1. Patient Knowledge Distillation for BERT Model Compression
1. Learning Efficient Object Detection Models with Knowledge Distillation
}

\subsubsection{Data Augmentation}
When training large models for complex tasks in a supervised learning regime, the size of the training data corpus correlates with improvement in generalization. \cite{Sun2017} demonstrates logarithmic increase in the prediction accuracy with increase in the number of labeled examples. However, getting high-quality labeled data often requires a human in the loop and could be expensive.

Data Augmentation is a nifty way of addressing the scarcity of labeled data, by synthetically inflating the existing dataset through some \emph{augmentation methods}. These augmentation methods are transformations that can be applied cheaply on the given examples, such that the new label of the augmented example does not change, or can be cheaply inferred. As an example, consider the classical image classification task of labeling a given image to be a cat or a dog. Given an image of a dog, translating the image horizontally / vertically by a small number of pixels, rotating it by a small angle, etc. would not materially change the image, so the transformed image should still be labeled as `dog' by the classifier. This forces the classifier to learn a robust representation of the image that generalizes better across these transformations. 

The transformations as described above have long been demonstrated to improve accuracy of convolutional networks \cite{Simard2003,Cirecsan2011}. They have also been a core part of seminal works in Image Classification. A prime example is AlexNet \cite{Krizhevsky2012}, where such transformations were used to increase the effective size of the training dataset by 2048 $\times$, which won the ImageNet competition in 2012. Since then it has became common to use such transformations for Image Classification models (Inception \cite{Szegedy2015}, XCeption \cite{Chollet2017}, ResNet \cite{He2016}, etc.).

\begin{figure}[h]
  \centering
  \includegraphics[width=10cm]{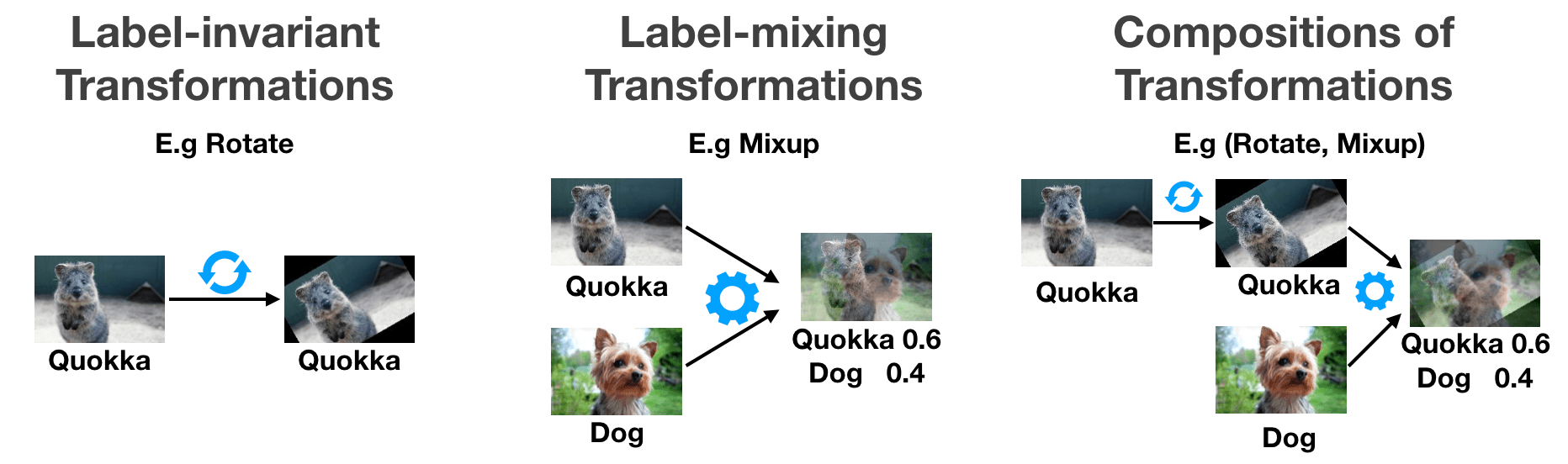}
  \caption{Some common types of data augmentations. Source: \cite{Li2020}}
  \label{fig:data-augmentation}
\end{figure}

We can categorize data-augmentation methods as follows (also refer to Figure \ref{fig:data-augmentation}):
\begin{itemize}
    \item \textbf{Label-Invariant Transformations}: These are some of the most common transformations, where the transformed example retains the original label. These can include simple geometric transformations such as translation, flipping, cropping, rotation, distortion, scaling, shearing, etc. However the user has to verify the label-invariance property with each transformation for the specific task at hand.
    \item \textbf{Label-Mixing Transformations}: Transformations such as Mixup \cite{Zhang2017}, mix inputs from two different classes in a weighted manner and treat the label to be a correspondingly weighted combination of the two classes (in the same ratio). The intuition is that the model should be able to extract out features that are relevant for both the classes. Other transformations like Sample Pairing also seem to help \cite{Inoue2018}.
    \item \textbf{Data-Dependent Transformations}: In this case, transformations are chosen such that they maximize the loss for that example \cite{Fawzi2016}, or are adversarially chosen so as to fool the classifier \cite{Gopalan2021}.
    \item \textbf{Synthetic Sampling}: These methods synthetically create new training examples. Algorithms like SMOTE \cite{Chawla2002} allow re-balancing the dataset to make up for skew in the datasets, and GANs can be used to synthetically create new samples \cite{Zhu2018} to improve model accuracy.
    \item \textbf{Composition of Transformations}: These are transformations that are themselves composed of other transformations, and the labels are computed depending on the nature of transformations that stacked.
\end{itemize}

\renewcommand{\arraystretch}{1.0}
\begin{table}[]
\small
\centering
\begin{tabular}{lL}
\hline
Transformation & Validation Accuracy Improvement (\%) \\ \hline
rotate         & 1.3                             \\ 
shear-x        & 0.9                             \\
shear-y        & 0.9                             \\
translate-x    & 0.4                             \\
translate-y    & 0.4                             \\
sharpness      & 0.1                             \\
autoContrast   & 0.1                             \\ \hline
\end{tabular}
\caption{
A breakdown of the contribution of various transformations on the validation accuracy of a model trained on the CIFAR-10 dataset. Source: \cite{Cubuk2020}.
}
\label{tab:vision-augmentation}
\end{table}

\textbf{Discussion}:
Apart from Computer Vision, Data-Augmentation has also been used in NLP, and Speech. In NLP, a common idea that has been used is `back-translation' \cite{Yu2018} where augmented examples are created by training two translation models, one going from the source language to the target language, and the other going back from the target language to the original source language. Since the back-translation is not exact, this process is able to generate augmented samples for the given input. Other methods like WordDropout \cite{Sennrich2016} stochastically set embeddings of certain words to zero. SwitchOut \cite{Wang2018} introduces a similarity measure to disallow augmentations that are too dissimilar to the original input. In Speech \cite{Hannun2014}, the input audio samples are translated to the left / right before being passed to the decoder.

While the augmentation policies are usually hand-tuned, there are also methods such as AutoAugment \cite{Cubuk2019} where the augmentation policy is learned through a Reinforcement-Learning (RL) based search, searching for the transformations to be applied, as well as their respective hyper-parameters. Though this is shown to improve accuracy, it is also complicated and expensive to setup a separate search for augmentation, taking as many as 15000 GPU hours to learn the optimal policy on ImageNet. The RandAugment \cite{Cubuk2020} paper demonstrated that it is possible to achieve similar results while reducing the search space to just two hyper-parameters (number of augmentation methods, and the strength of the distortion) for a given model and dataset.

Overall, we see that data-augmentation leads to better generalization of the given models. Some techniques can be specific for their domains RandAugment (Vision), BackTranslation and SwitchOut (NLP), etc. However, the core principles behind them make it likely that similar methods can be derived for other domains too (refer to our categorization of data-augmentation methods above).

\somecomment{
1. https://towardsdatascience.com/analyzing-data-augmentation-for-image-classification-3ed30aa61411 -- Get the figure with the embeddings of the augmented images.
1. 
}

\subsubsection{Self-Supervised Learning}
The Supervised-Learning paradigm relies heavily on labeled data. As mentioned earlier, it requires human intervention, and is expensive as well. To achieve reasonable quality on a non-trivial task, the amount of labeled data requires is large too. While techniques like Data-Augmentation, Distillation etc., help, they too rely on the presence of some labeled data to achieve a baseline performance.

\begin{figure}[h]
  \centering
  \includegraphics[width=6cm]{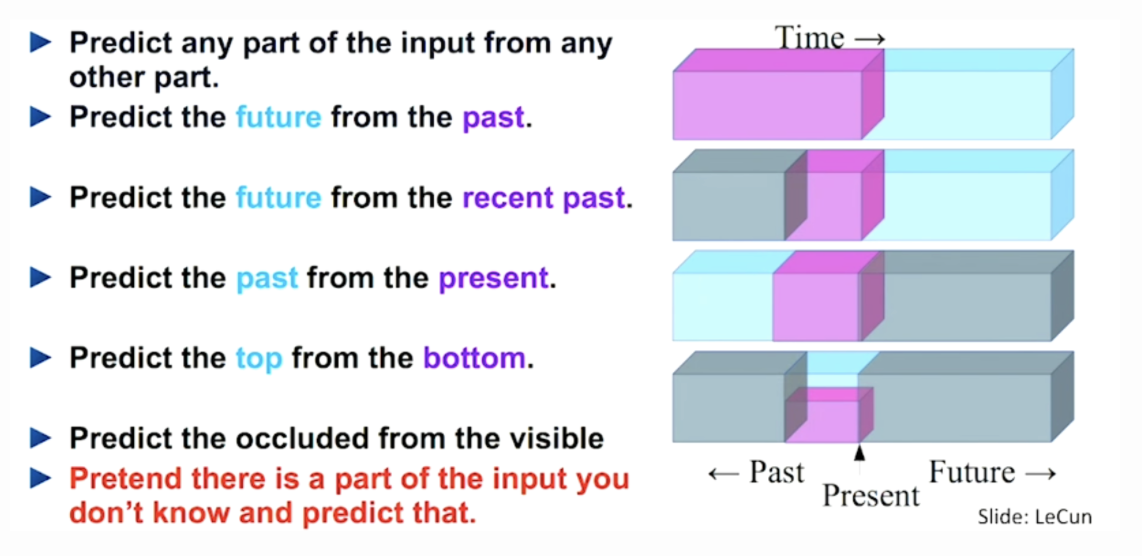}
  \caption{General theme of pretext tasks. Source: \cite{LeCun2018}}
  \label{fig:pre-training-tasks}
\end{figure}

Self-Supervised learning (SSL) avoids the need for labeled data to learn generalized representations, by aiming to extract more supervisory bits from each example. Since it focuses on learning robust representations of the example itself, it does not need to focus narrowly on the label. This is typically done by solving a \emph{pretext task} where the model pretends that a part / structure of the input is missing and learns to predict it (Refer to Figure \ref{fig:pre-training-tasks} for examples). Since unlabeled data is vast in many domains (Books, Wikipedia, and other text for NLP, Web Images \& Videos for Computer Vision, etc.), the model would not be bottlenecked by data for learning to solve these pretext tasks. 

Once the models learn generic representations that transfer well across tasks, they can be adapted to solve the target task by adding some layers that project the representation to the label space, and fine-tuning the model with the labeled data. Since the labeled data is not being used for learning rudimentary features, but rather how to map the high-level representations into the label space, the quantum of labeled data is going to be a fraction of what would have been required for training the model from scratch. From this lens, fine-tuning models pre-trained with Self-Supervised learning are \emph{data-efficient} (they converge faster, attain better quality for the same amount of labeled data when compared to training from scratch, etc.) (\cite{Howard2018,Devlin2018}).

\begin{figure}[h]
  \centering
  \includegraphics[width=5cm]{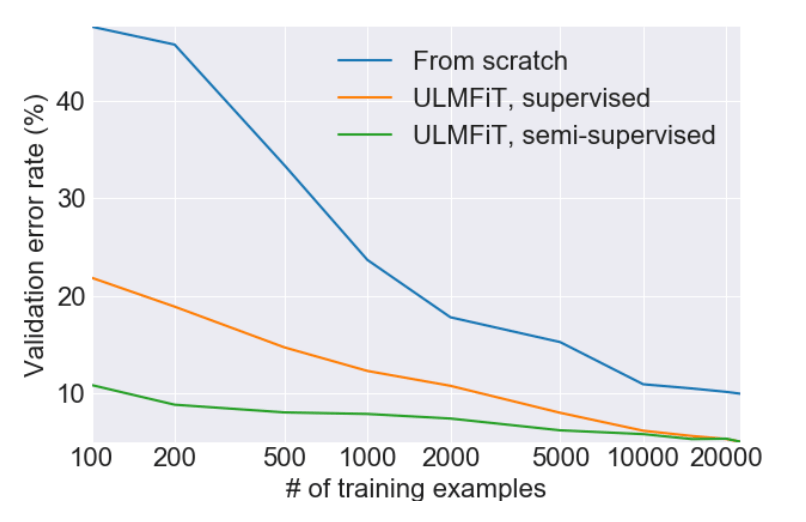}
  \caption{Validation Error w.r.t. number of training examples for different training methods on IMDb (from scratch, ULMFiT Supervised: pre-training with WikiText-103 and fine-tuning using labeled data, ULMFit Semi-Supervised: Pre-Training with WikiText-103 as well as unlabeled data from the target dataset and fine-tuning with labeled data). Source: \cite{Howard2018}}
  \label{fig:ulmfit}
\end{figure}

An example of this two step process of pre-training on unlabeled data and fine-tuning on labeled data has gained rapid acceptance in the NLP community. ULMFiT \cite{Howard2018} pioneered the idea of training a general purpose language model, where the model learns to solve the pretext task of predicting the next word in a given sentence, without the neeof an associated label. The authors found that using a large corpus of preprocessed unlabeled data such as the WikiText-103 dataset (derived from English Wikipedia pages) was a good choice for the pre-training step. This was sufficient for the model to learn general properties about the language, and the authors found that fine-tuning such a pre-trained model for a binary classification problem (IMDb dataset) required only 100 labeled examples ($\approx 10\times$ less labeled examples otherwise). Refer to Figure \ref{fig:ulmfit}. If we add a middle-step of pre-training using unlabeled data from the same target dataset, the authors report needing $\approx 20\times$ fewer labeled examples.

This idea of pre-training followed by fine-tuning is also used in BERT \cite{Devlin2018} (and other related models like GPT, RoBERTa, T5, etc.) where the pre-training steps involves learning to solve two tasks. Firstly, the Masked Language Model where about 15\% of the tokens in the given sentence are masked and the model needs to predict the masked token. The second task is, given two sentences $A$ and $B$, predict if $B$ follows $A$. The pre-training loss is the mean of the losses for the two tasks. Once pre-trained the model can then be used for classification or seq2seq tasks by adding additional layers on top of the last hidden layer. When it was published, BERT beat the State-of-the-Art on eleven NLP tasks.

\begin{figure}%
    \centering
    \subfloat[\centering Detecting relative order of patches. Source: \cite{Doersch2015}.]{{\includegraphics[width=5cm]{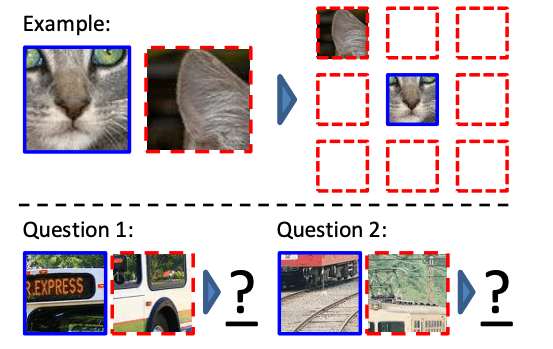} }}%
    \subfloat[\centering Predicting the degree of rotation of a given image.]{{\includegraphics[width=5cm]{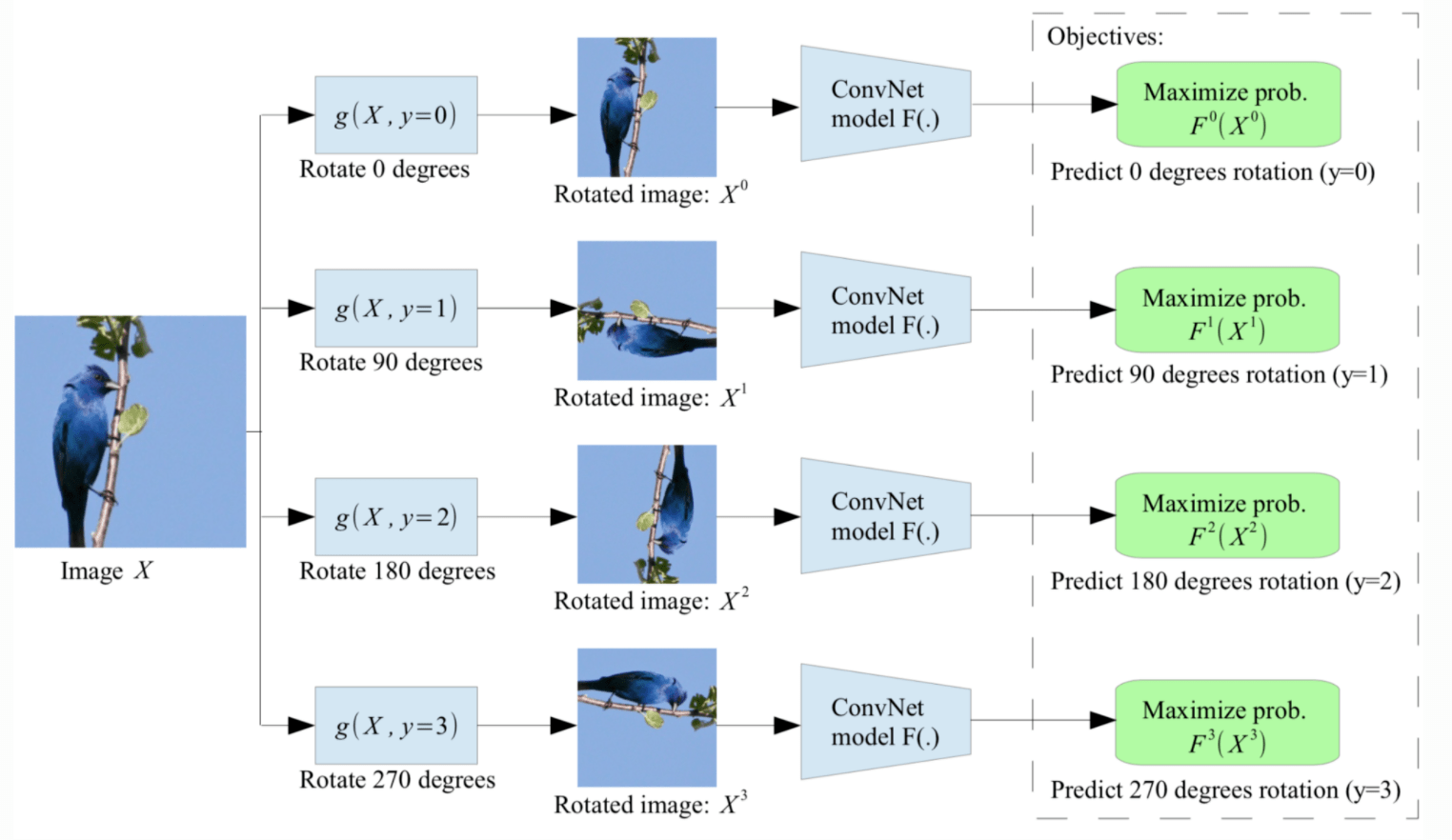} }}%
    \caption{Pretext tasks for vision problems.}%
    \label{fig:vision-pretext-tasks}%
\end{figure}

Similar to NLP, the pretext tasks in Vision have been used to train models that learn general representations. \cite{Doersch2015} extracts two patches from a training example and then trains the model to predict their relative position in the image (Refer to Figure \ref{fig:vision-pretext-tasks}(a)). They demonstrate that using a network pre-trained in this fashion improves the quality of the final object detection task, as compared to randomly initializing the network. Similarly, another task is to predict the degree of rotation for a given rotated image \cite{Gidaris2018}. The authors report that the network trained in a self-supervised manner this way can be fine-tuned to perform nearly as well as a fully supervised network.

Another common theme is Contrastive Learning, where the model is trained to distinguish between similar and dissimilar inputs. Frameworks such as SimCLR \cite{Chen2020a, Chen2020b}, try to learn representations $h_i$ and $h_j$ for two given inputs $\tilde{x_i}$ and $\tilde{x_j}$, where the latter two are differently augmented views of the same input, such that the cosine similarity of the projections of $h_i$ and $h_j$, $z_i$ and $z_j$ (using a separate function $g(.)$) can be maximized. Similarly, for dissimilar inputs the cosine similarity of $z_i$ and $z_j$ should be minimized. The authors report a Top-1 accuracy of $73.9\%$ on ImageNet with only 1\% labels (13 labels per class), and outperform the ResNet-50 supervised baseline with only 10\% labels.

\begin{figure}[h]
  \centering
  \includegraphics[width=4cm]{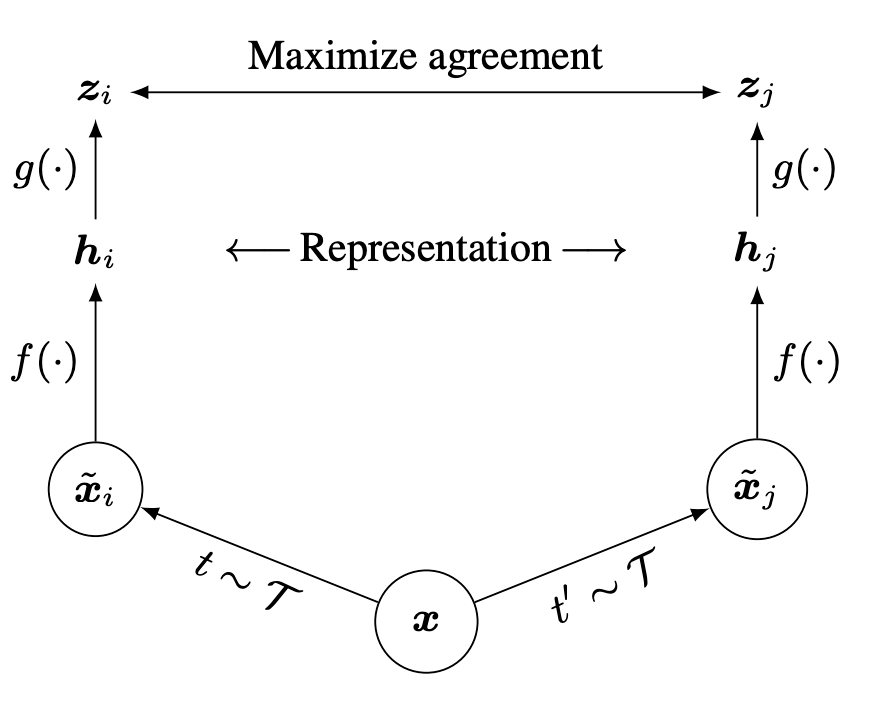}
  \caption{SimCLR framework for learning visual representations. Source: \cite{Chen2020a}}
  \label{fig:simclr}
\end{figure}

\textbf{Discussion}:
Self-Supervised Learning (SSL) has demonstrated significant success in the general representational learning with unlabeled data, followed by fine-tuning to adapt the model to the target task with a modest number of labeled examples. Yann LeCun has likened Self-Supervision as the cake, and Supervised Learning as the icing on top \cite{LeCun2018}, implying that SSL will be the primary way of training high-quality models in the future as we move beyond tasks where labeled data is abundant. 

With unlabeled data being practically limitless, SSL's success is dependent on creating useful pretext tasks for the domain of interest. As demonstrated across NLP \cite{Howard2018,Devlin2018}, Vision \cite{Chen2020a,Patrick2020}, Speech \cite{Glass2012}, etc., Self-Supervision is indeed not just helpful in speeding and improving convergence, but also enabling achieving high quality in tasks where it was intractable to get enough labeled samples. 

Practically, for someone training Deep Learning models on a custom task (say a speech recognition model for a remote African dialect), having a pre-trained checkpoint of a model trained in a self-supervised fashion (such as wav2vec \cite{Baevski2020}, which pre-trained in a similar way to BERT \cite{Devlin2018}), enables them to only spend an extremely tiny fraction of resources on both data labeling, as well as training to fine-tune to a good enough quality. In some cases, such as SimCLR \cite{Chen2020b}, SSL approaches have actually beaten previous supervised baselines with sophisticated models like ResNet-50. Hence, we are hopeful that SSL methods will be crucial for ML practitioners for training high-quality models cheaply.

\subsection{Automation}
It is possible to delegate some of the work around efficiency to automation, and letting automated approaches search for ways of training more efficient models. Apart from reducing work for humans, it also lowers the bias that manual decisions might introduce in model training, apart from systematically and automatically looking for optimal solutions. The trade-off is that these methods might require large computational resources, and hence have to be carefully applied.

\subsubsection{Hyper-Parameter Optimization (HPO)}
One of the commonly used methods that fall under this category is Hyper-Parameter Optimization (HPO) \cite{Yu2020}. Hyper-parameters such as initial learning rate, weight decay, etc. have to be carefully tuned for faster convergence \cite{Jordan2020}. They can also decide the network architecture such as the number of fully connected layers, number of filters in a convolutional layer, etc.


Experimentation can help us build an intuition for the \emph{range} in which these parameters might lie, but finding the best values requires a search for the exact values that optimize the given objective function (typically the loss value on the validation set). Manually searching for these quickly becomes tedious with the growth in the number of hyper-parameters and/or their possible values. Hence, let us explore possible algorithms for automating the search. To formalize this, let us assume without the loss of generalization, that we are optimizing the loss value on the given dataset's validation split. Then, let $\mathcal{L}$ be the loss function, $f$ be the model function that is learnt with the set of hyper-parameters ($\lambda$), $x$ be the input, and $\theta$ be the model parameters. With the search, we are trying to find $\lambda^{*}$ such that,

\begin{equation}
    \small
    \lambda^{*} = \argmin_{\lambda \in \Lambda}  \mathcal{L}(f_{\lambda}(x; \theta), y)
    \label{eq:hyper-param-opt}
\end{equation}


$\Lambda$ is the set of all possible hyper-parameters. In practice, the $\Lambda$ can be a very large set containing all possible combinations of the hyper-parameters, which would often be intractable since hyper-parameters like learning rate are real-valued. A common strategy is to approximate $\Lambda$ by picking a finite set of \emph{trials}, $S = \{\lambda^{(1)}, \lambda^{(2)}, ..., \lambda^{(n)}\}$, such that $S \in \Lambda$, and then we can approximate Equation (\ref{eq:hyper-param-opt}) with:

\begin{equation}
    \small
    \lambda^{*} \approx \argmin_{\lambda \in \{\lambda^{(1)}, ..., \lambda^{(n)}\}}  \mathcal{L}(f_{\lambda}(x; \theta), y)
    \label{eq:hyper-param-opt-approx}
\end{equation}

As we see, the choice of $S$ is crucial for the approximation to work. The user has to construct a range of reasonable values for each hyper-parameter $\lambda_{i} \in \lambda$. This can be based on prior experience with those hyper-parameters.

A simple algorithm for automating HPO is \textbf{Grid Search} (also referred to as Parameter Sweep), where $S$ consists of all the distinct and valid combinations of the given hyper-parameters based on their specified ranges. Each trial can then be run in parallel since each trial is independent of the others, and the optimal combination of the hyper-parameters is found once all the trials have completed. Since this approach tries all possible combinations, it suffers from the \emph{curse of dimensionality}, where the total number of trials grow very quickly.

Another approach is \textbf{Random Search} where trials are sampled randomly from the search space \cite{Bergstra2012}. Since each trial is independent of the others, it can still be executed randomly. However, there are few critical benefits of Random Search:
\begin{enumerate}
    \item Since the trials are i.i.d. (not the case for Grid Search), the resolution of the search can be changed on-the-fly (if the computational budget has changed, or certain trials have failed).
    \item Likelihood of finding the optimal $\lambda^{*}$ increases with the number of trials, which is not the case with Grid Search.
    \item If there are $K$ real-valued hyper-parameters, and $N$ total trials, grid search would pick $N^{\frac{1}{K}}$ for each hyper-parameter. However, not all hyper-parameters might be important. Random Search picks a random value for each hyper-parameter per trial. Hence, in cases with low effective dimensionality of the search space, Random Search performs better than Grid Search.
\end{enumerate}

\begin{figure}%
    \centering
    \subfloat[\centering Grid Search]{{\includegraphics[width=5cm]{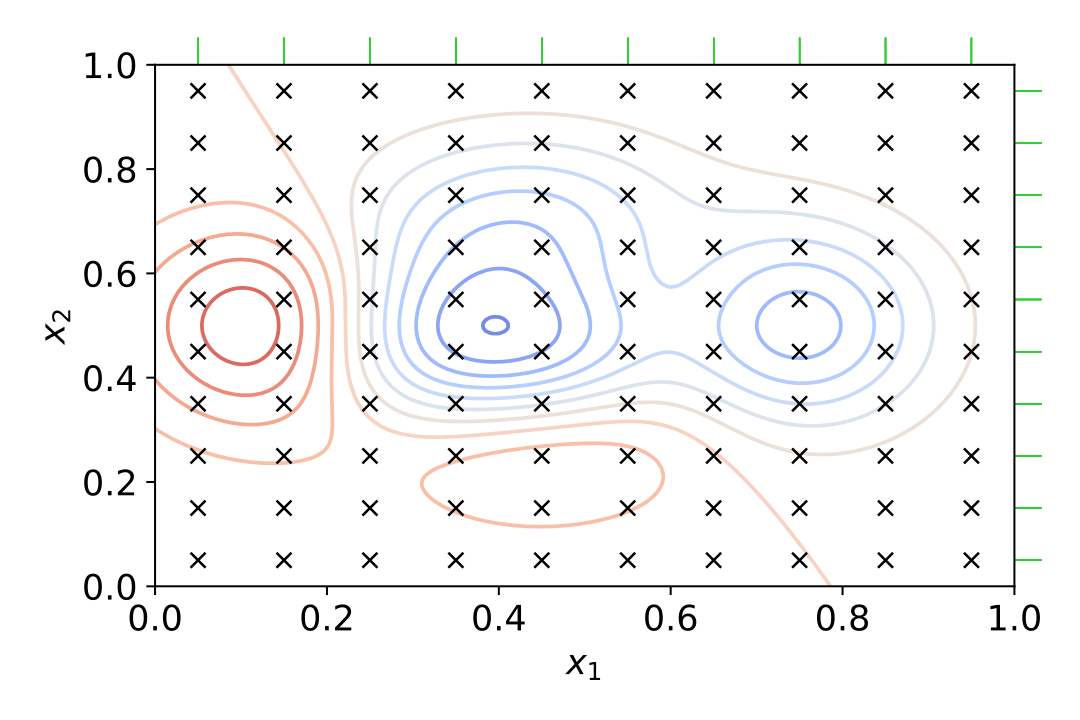} }}%
    \subfloat[\centering Random Search]{{\includegraphics[width=5cm]{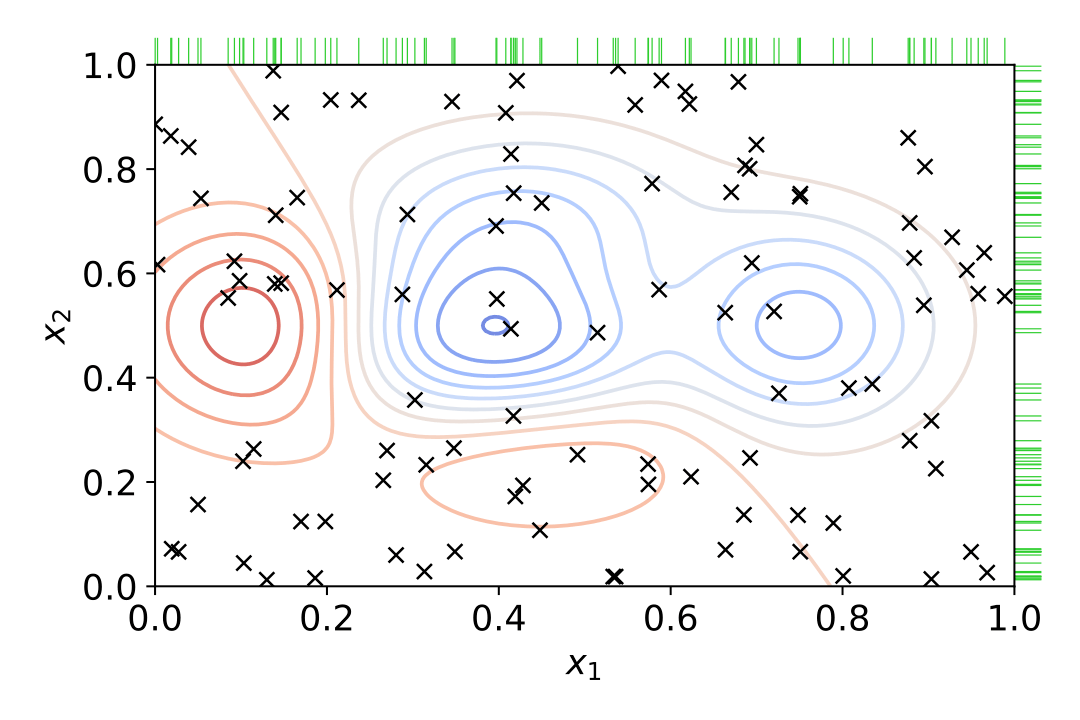} }}%
    \subfloat[\centering Bayesian Optimization]{{\includegraphics[width=5cm]{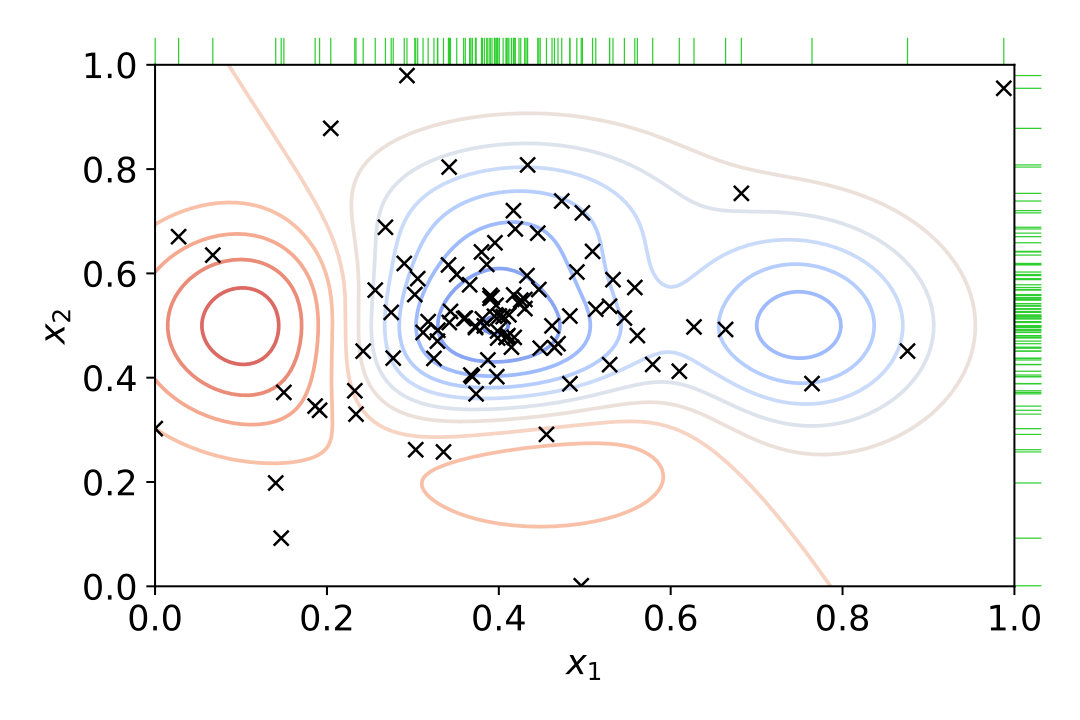} }}%
    \caption{Hyper-Parameter Search algorithms. Source: \cite{Wikimedia2021HPO}}%
    \label{fig:hpo-algorithms}%
\end{figure}

\textbf{Bayesian Optimization} (BO) based search \cite{Movckus1975, Agnihotri2020} is a \emph{model-based} sequential approach where the search is guided by actively estimating the value of the objective function at different points in the search space, and then spawning trials based on the information gathered so far. The estimation of the objective function is done using a \emph{surrogate function} that starts off with a prior estimate. The trials are created using an \emph{acquisition function} which picks the next trial using the surrogate function, the likelihood of improving on the optimum so far, whether to explore / exploit etc. As the trials complete, both these functions will refine their estimates. Since the method keeps an internal model of how the objective function looks and plans the next trials based on that knowledge, it is model-based. Also, since the selection of trials depends on the results of the past trials, this method is sequential. BO improves over Random Search in that the search is guided rather than random, thus fewer trials are required to reach the optimum. However, it also makes the search sequential (though it is possible to run multiple trials in parallel, overall it will lead to some wasted trials).

One of the strategies to save training resources with the above search algorithms is the \textbf{Early Stopping} of trials that are not promising. Google's Vizier \cite{Golovin2017} uses Median Stopping Rule for early stopping, where a trial is terminated if it's performance at a time step $t$ is below the the median performance of all trials run till that point of time.

Other algorithms for HPO include:
\begin{enumerate}
    \item \textbf{Population Based Training (PBT)} \cite{Jaderberg2017}: This method is similar to evolutionary approaches like genetic algorithms, where a fixed number of trials (referred to as the population) are spawned and trained to convergence. Each trial starts with a random set of hyper-parameters, and trained to a pre-determined number of steps. At this point, all trials are paused, and every trial's weights and parameters might be replaced by the weights and parameters from the `best' trial in the population so far. This is the \emph{exploitation} part of the search. For \emph{exploration}, these hyper-parameters are perturbed from their original values. This process repeats till convergence. It combines both the search and training in a fixed number of trials that run in parallel. It also only works with adaptive hyper-parameters like learning rate, weight-decay, etc. but cannot be used where hyper-parameters change the model structure. Note that the criteria for picking the `best' trial does not have to be differentiable. 
    \item \textbf{Multi-Armed Bandit Algorithms}: Methods like Successive Halving (SHA) \cite{Jamieson2016} and Hyper-Band \cite{Li2017} are similar to random search, but they allocate more resources to the trials which are performing well. Both these methods need the user to specify the total computational budget $B$ for the search (can be the total number of epochs of training, for instance). They then spawn and train a fixed number of trials with randomly sampled hyper-parameters while allocating the training budget. Once the budget is exhausted, the worse performing fraction ($\frac{\eta - 1}{\eta}$) of the trials are eliminated, and the remaining trials' new budget is multiplied by $\eta$. In the case of SHA, $\eta$ is 2, so the bottom $\frac{1}{2}$ of the trials are dropped, and the training budget for the remaining trials is doubled. For Hyper-Band $\eta$ is 3 or 4. Hyper-Band differs from SHA in that the user does not need to specify the maximum number of parallel trials, which introduces a trade-off between the total budget and the per-trial allocation.
\end{enumerate}

\textbf{HPO Toolkits}: There are several software toolkits that incorporate HPO algorithms as well as an easy to use interface (UI, as well as a way to specify the hyper-parameters and their ranges). Vizier \cite{Golovin2017} (an internal Google tool, also available via Google Cloud for blackbox tuning). Amazon offers Sagemaker \cite{Perrone2020} which is functionally similar and can also be accessed as an AWS service. NNI \cite{MicrosoftResearch2019NNI}, Tune \cite{Liaw2018}, Advisor \cite{Chen2021} are other open-source HPO software packages that can be used locally.

\subsubsection{Neural Architecture Search (NAS)}
Neural Architecture Search can be thought of an extension of Hyper-Parameter Optimization wherein we are searching for parameters that change the network architecture itself.

We find that there is consensus in the literature \cite{Elsken2019} around categorizing NAS as a system comprising of the following parts:
\begin{enumerate}
    \item \textbf{Search Space}: These are the operations that are allowed in the graph (Convolution ($1\times1, 3\times3, 5\times5$), Fully Connected, Pooling, etc.), as well as the semantics of how these operations and their outputs connect to other parts of the network.
    \item \textbf{Search Algorithm \& State}: This is the algorithm that controls the architecture search itself. Typically the standard algorithms that apply in HPO (Grid Search, Random Search, Bayesian Optimization, Evolutionary Algorithms), can be used for NAS as well. However, using Reinforcement Learning (RL) \cite{Zoph2016}, and Gradient Descent \cite{Liu2018b} are popular alternatives too. 
    \item \textbf{Evaluation Strategy}: This defines how we evaluate a model for fitness. It can simply be a conventional metric like validation loss, accuracy, etc. Or it can also be a compound metric, as in the case of MNasNet \cite{Tan2019} which creates a single custom metric based on accuracy as well as latency.
\end{enumerate}

\begin{figure}[h]
  \centering
  \includegraphics[width=7.5cm]{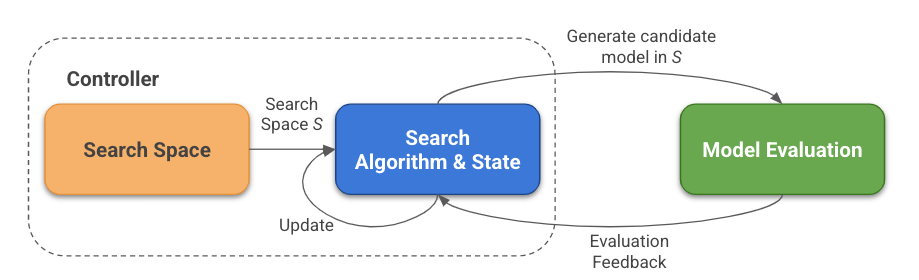}
  \caption{Neural Architecture Search: The controller can be thought of as a unit that encodes the search space, the search algorithm itself, and the state it maintains (typically the model that helps generate the candidates). The algorithm generates candidate models in the search space $S$, and receives an evaluation feedback. This feedback is used to update the state, and generate better candidate models.}
  \label{fig:nas-controller}
\end{figure}

The user is supposed to either explicitly or implicitly encode the search space. Together with the search algorithm, we can view this as a `controller' which generates sample candidate networks (Refer to Figure \ref{fig:nas-controller}). The evaluation stage will then train and evaluate these candidates for fitness. This fitness value is then passed as feedback to the search algorithm, which will use it for generating better candidates. While the implementation of each of these blocks vary, this structure is common across the seminal work in this area.

\begin{figure}[h]
  \centering
  \includegraphics[width=10cm]{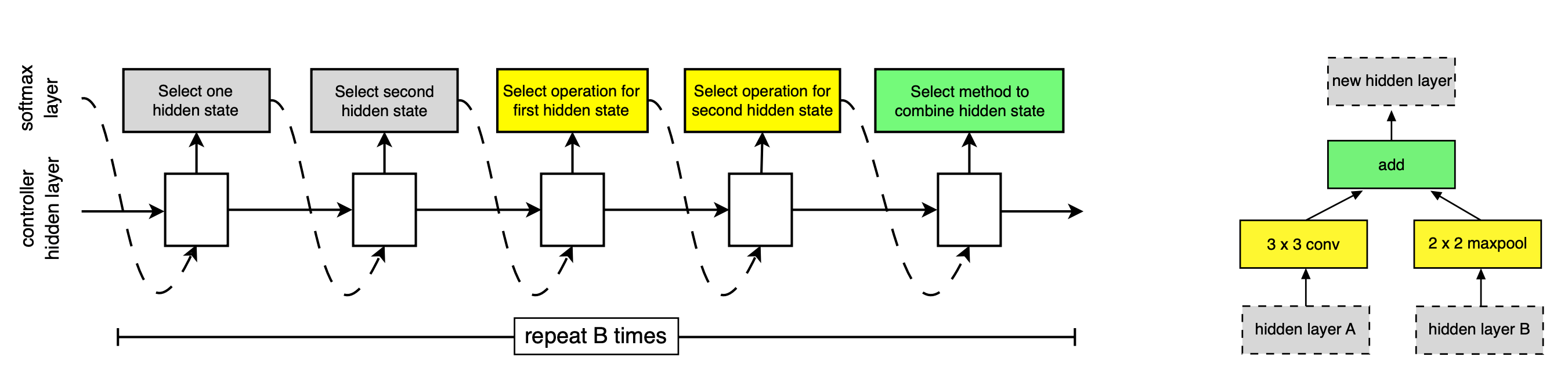}
  \caption{A NASNet controller generating the architecture, recursively making one decision at a time and generating a single block in the image (making a total of 5 decisions). Source: \cite{Zoph2018}.}
  \label{fig:nasnet-controller}
\end{figure}

Zoph et. al's paper from 2016 \cite{Zoph2016}, demonstrated that end-to-end neural network architectures can be generated using Reinforcement Learning. In this case, the controller is a Recurrent Neural Network, which generates the architectural hyper-parameters of a feed-forward network one layer at a time, for example, number of filters, stride, filter size, etc. They also support adding skip connections (refer Figure \ref{fig:nasnet-controller}). The network semantics are baked into the controller, so generating a network that behaves differently requires changing the controller. Also, training the controller itself is expensive (taking 22,400 GPU hours \cite{Zoph2018}), since the entire candidate network has to be trained from scratch for a single gradient update to happen. In a follow up paper \cite{Zoph2018}, they come up with a refined search space where instead of searching for the end-to-end architecture, they search for \emph{cells}: A `Normal Cell' that takes in an input, processes it, and returns an output of the same spatial dimensions. And a `Reduction Cell' that process its input, and returns an output whose spatial dimensions are scaled down by a factor of 2. Each cell is a combination of $B$ blocks. The controller's RNN generates one block at a time, where it picks outputs of two blocks in the past, the respective operations to apply on them, and how to combine them into a single output. The Normal and Reduction cells are stacked in alternating fashion ($N$ Normal cells followed by 1 Reduction cell, where $N$ is tunable) to construct an end-to-end network for CIFAR-10 and ImageNet. Learning these cells individually rather than learning the entire network seems to improve the search time by 7$\times$, when compared to the end-to-end network search in \cite{Zoph2016}, while beating the state-of-the-art in CIFAR-10 at that time.

Other approaches such as evolutionary techniques \cite{Real2019}, differentiable architecture search \cite{Liu2018b}, progressive search \cite{Liu2018c}, parameter sharing \cite{Pham2018}, etc. try to reduce the cost of architecture search (in some cases reducing the compute cost to a couple of GPU days instead of thousands of GPU days). These are covered in detail in \cite{Elsken2019}.

While most of the early papers focused on finding the architectures that performed best on quality metrics like accuracy, unconstrained by the footprint metrics. However, when focusing on efficiency, we are often interested in specific tradeoffs between quality and footprint. Architecture Search can help with multi-objective searches that optimize for both quality and footprint. MNasNet \cite{Tan2019} is one such work. It incorporates the model's latency on the target device into the objective function directly, as follows:

\begin{equation}
\small
\underset{m}{\operatorname{maximize}} \quad A C C(m) \times\left[\frac{L A T(m)}{T}\right]^{w}
\end{equation}

Where $m$ is the candidate model, $ACC$ is the accuracy metric, and $LAT$ is the latency of the given model on the desired device. $T$ is the target latency. $w$ is recommended to be $-0.07$. FBNet \cite{Wu2019} uses a similar approach with a compound reward function that has a weighted combination of the loss value on the validation set and the latency. However instead of measuring the latency of the candidate model on device, they use a pre-computed lookup table to approximate the latency to speed up the search process. They achieve networks that are upto $2.4\times$ smaller and $1.5\times$ faster than MobileNet, while finishing the search in 216 GPU Hours. Other works such as MONAS \cite{Hsu2018} use Reinforcement Learning to incorporate power consumption into the reward function along with hard constraints on the number of MAC operations in the model, and discover pareto-frontiers under the given constraints. 

\textbf{Discussion}: Automation plays a critical role in model efficiency. Hyper-Parameter Optimization (HPO) is now a natural step in training models and can extract significant quality improvements, while minimizing human involvement. In case the cost HPO becomes large, algorithms like Bayesian Optimization, Hyper-Band etc. with early stopping techniques can be used. HPO is also available in ready-to-use software packages like Tune \cite{Liaw2018}, Vizier via Google Cloud \cite{Golovin2017}, NNI \cite{MicrosoftResearch2019NNI}, etc. Similarly, recent advances in Neural Architecture Search (NAS) also make it feasible to construct architectures in a learned manner, while having constraints on both quality and footprint \cite{Tan2019}. Assuming several hundred GPU hours worth of compute required for the NAS run to finish, and an approx cost of \$3 GPU / hour on leading cloud computing services, this makes using NAS methods financially feasible and not similar in cost to manual experimentation with model architecture when optimizing for multiple objectives.

\subsection{Efficient Architectures}
Another common theme for tackling efficiency problems is to go back to the drawing board, and design layers and models that are efficient by design to replace the baseline. They are typically designed with some insight which might lead to a design that is better in general, or it might be better suited for the specific task. In this section, we lay out an examples of such efficient layers and models to illustrate this idea.

\subsubsection{Vision}
One of the classical example of efficient layers in the Vision domain are the Convolutional layers, which improved over Fully Connected (FC) layers in Vision models. FC layers suffer from two primary issues:
\begin{enumerate}
    \item FC layers ignore the spatial information of the input pixels. Intuitively, it is hard to build an understanding of the given input by looking at individual pixel values in isolation. They also ignore the spatial locality in nearby regions. 
    \item Secondly, using FC layers also leads to an explosion in the number of parameters when working with even moderately sized inputs. A $100\times100$ RGB image with 3 channels, would lead to each neuron in the first layer having $3\times10^4$ connections. This makes the network susceptible to overfitting also.
\end{enumerate}

Convolutional layers avoid this by learning ‘filters’, where each filter is a 3D weight matrix of a fixed size ($3\times3$, $5\times5$, etc.), with the third dimension being the same as the number of channels in the input. Each filter is convolved over the input to generate a feature map for that given filter. These filters learn to detect specific features, and convolving them with a particular input patch results in a single scalar value that is higher if the feature is present in that input patch. 

These learned features are simpler in lower layers (such as edges (horizontal, vertical, diagonal, etc.)), and more complex in subsequent layers (texture, shapes, etc.). This happens because the subsequent layers use the feature maps generated by previous layers, and each pixel in the input feature map of the $i$-th layer, depends on the past $i-1$ layers. This increases the \emph{receptive field} of the said pixel as $i$ increases, progressively increasing the complexity of the features that can be encoded in a filter. 

The core idea behind the efficiency of Convolutional Layers is that the same filter is used everywhere in the image, regardless of where the filter is applied. Hence, enforcing spatial invariance while sharing the parameters. Going back to the example of a $100\times100$ RGB image with 3 channels, a $5\times5$ filter would imply a total of $75$ ($5\times5\times3$) parameters. Each layer can learn multiple unique filters, and still be within a very reasonable parameter budget. This also has a regularizing effect, wherein a dramatically reduced number of parameters allow for easier optimization, and reducing the likelihood of overfitting. 


Convolutional Layers are usually coupled with Pooling Layers, which allow dimensionality reduction by subsampling the input (aggregating a sliding 2-D window of pixels, using functions like max, avg, etc.). Pooling would lead to smaller feature maps for the next layer to process, which makes it faster to process. LeNet5 \cite{Lecun1998} was the first Convolutional Network which included convolutional layers, pooling, etc. Subsequently, many iterations of these networks have been proposed with various improvements. AlexNet \cite{Krizhevsky2012}, Inception \cite{Szegedy2015}, ResNet \cite{He2016}, etc. have all made significant improvements over time on known image classification benchmarks using Convolutional Layers.

\textbf{Depth-Separable Convolutional Layers}: In the convolution operation, each filter is used to convolve over the two spatial dimensions and the third channel dimension. As a result, the size of each filter is $s_x \times s_y \times$ \texttt{input\_channels}, where $s_x$ and $s_y$ are typically equal. This is done for each filter, resulting in the convolution operation happening both spatially in the $x$ and $y$ dimensions, and depthwise in the $z$ dimension.

\begin{figure}[h]
  \centering
  \includegraphics[width=5cm]{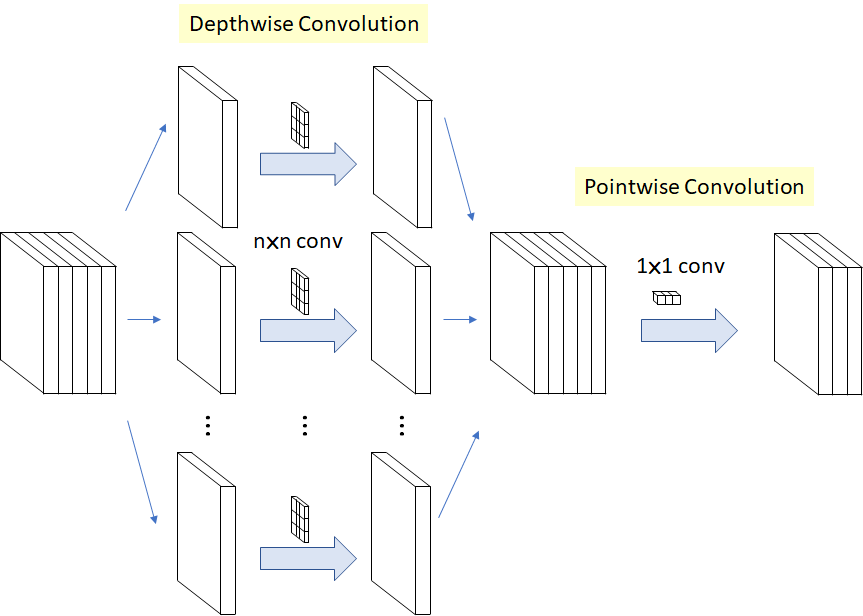}
  \caption{Depth-Separable Convolution. Source: \cite{Tsang2019}.}
  \label{fig:depthwise}
\end{figure}

Depth-separable convolution breaks this into two steps (Refer to Figure \ref{fig:depthwise}):
\begin{enumerate}
    \item Doing a point-wise convolution with $1 \times 1$ filters, such that the resulting feature map now has a depth of \texttt{output\_channels}.
    \item Doing a spatial convolution with $s_x \times s_y$ filters in the $x$ and $y$ dimensions.
\end{enumerate}
 
These two operations stacked together (without any intermediate non-linear activation) results in an output of the same shape as a regular convolution, with much fewer parameters ($1\times1\times$\texttt{input\_channels}$\times$ \texttt{output\_channels}$) + (s_x \times s_y \times$ \texttt{output\_channels}$)$, v/s $s_x\times s_y\times $ \texttt{input\_channels} $\times$ \texttt{output\_channels} for the regular convolution).  Similarly there is an order of magnitude less computation since the point-wise convolution is much cheaper for convolving with each input channel depth-wise (for more calculations refer to \cite{Sandler2018}). The Xception model architecture \cite{Chollet2017} demonstrated that using depth-wise separable convolutions in the Inception architecture, allowed reaching convergence sooner in terms of steps and a higher accuracy on the ImageNet dataset while keeping the number of parameters the same. 

The MobileNet model architecture \cite{Sandler2018} which was designed for mobile and embedded devices, also uses the depth-wise separable layers instead of the regular convolutional layers. This helps them reduce the number of parameters as well as the number of multiply-add operations by $7-10\times$ and allows deployment on Mobile for Computer Vision tasks. Users can expect a latency between 10-100ms depending on the model. MobileNet also provides a knob via the depth-multiplier for scaling the network to allow the user to trade-off between accuracy and latency. 

\subsubsection{Natural Language Understanding} \hfill
\\
\textbf{Attention Mechanism \& Transformer Family}:
One of the issues plaguing classical Sequence-to-Sequence (Seq2Seq) models for solving tasks such as Machine Translation (MT), was that of the information-bottleneck. Seq2Seq models typically have one or more encoder layers which encode the given input sequence ($\mathbf{x} = (x_1, x_2, ..., x_{T})$) into a fixed length vector(s) (also referred to as the context, $\mathbf{c}$), and one or more decoder layers which generate another sequence using this context. In the case of MT, the input sequence can be a sentence in the source language, and the output sequence can be the sentence in the target language.

\begin{figure}
\centering
\begin{minipage}{.55\textwidth}
  \centering
  \includegraphics[width=1.0\linewidth]{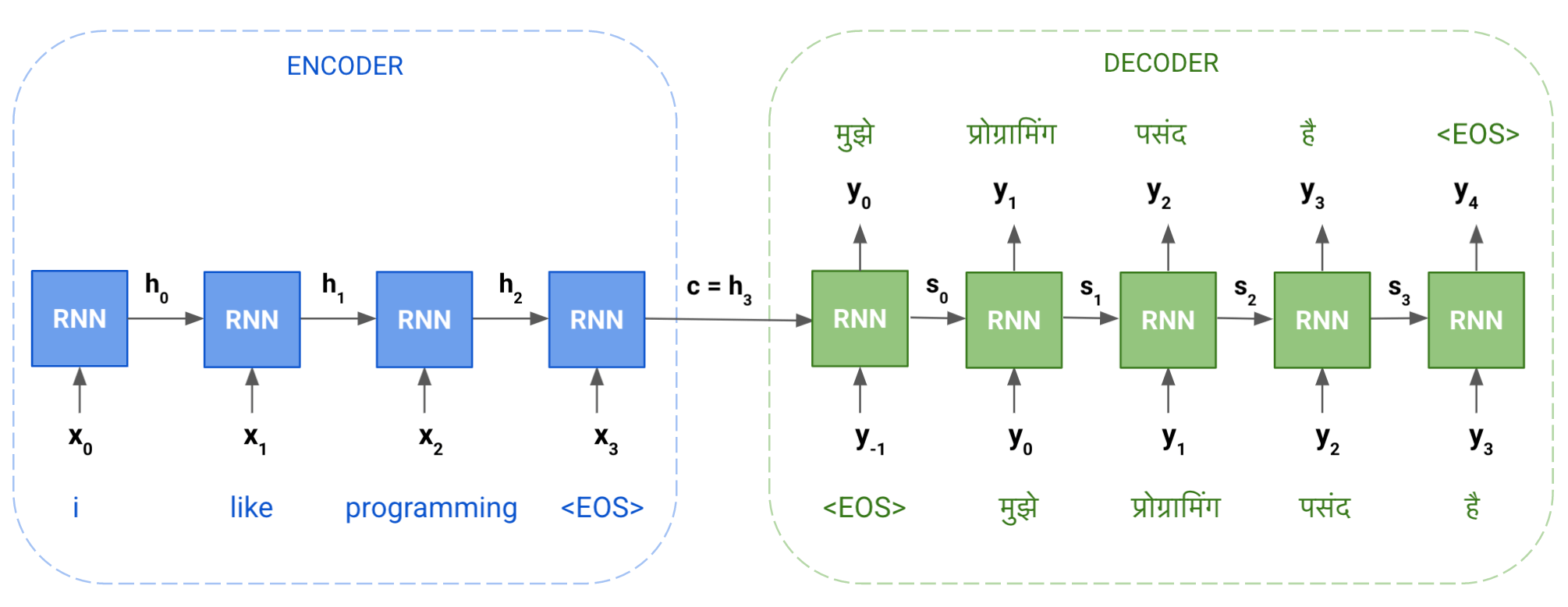}
  \captionof{figure}{Information Bottleneck in a Seq2Seq model for translating from English to Hindi. The context vector $c$ that the decoder has access to is fixed, and is typically the last hidden state ($h_T$).}
  \label{fig:information-bottleneck}
\end{minipage}%
\qquad
\begin{minipage}{.38\textwidth}
  \centering
  \includegraphics[width=.4\linewidth]{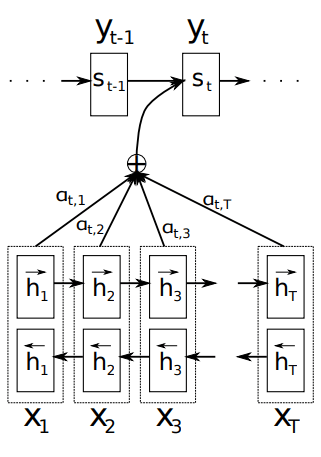}
  \captionof{figure}{Attention module learning a weighted context vector for each output token from the hidden states. Source: \cite{Bahdanau2014}.}
  \label{fig:bahdanau-attn}
\end{minipage}
\end{figure}

However, in classical Seq2Seq models such as \cite{Sutskever2014} the decoder layers could only see the hidden state of the final encoder step ($c = h_{T}$). This is a \emph{bottleneck} because the encoder block has to squash all the information about the sequence in a single context vector for all the decoding steps, and the decoder block has to somehow infer the entire encoded sequence from it (Refer to Figure \ref{fig:information-bottleneck}). It is possible to increase the size of the context vector, but it would lead to an increase in the hidden state of all the intermediate steps, and make the model larger and slower. 

The Attention mechanism was introduced in Bahdanau et al. \cite{Bahdanau2014} to be able to create a custom context vector for each output token, by allowing all hidden states to be visible to the decoder and then creating a weighted context vector, based on the output token's alignment with each input token. Essentially, the new weighted context vector is $c_i = \sum_{j}^{T} \alpha_{ij} . h_j$, where $\alpha_{ij}$ is the learned alignment (attention weight) between the decoder hidden state $s_{i-1}$ and the hidden state for the $j$-th token ($h_j$). $\alpha_{ij}$ could be viewed as how much attention should the $i$-th input token be given when processing the $j$-th input token. This model is generalized in some cases by having explicit Query ($Q$), Key ($K$), and Value ($V$) vectors. Where we seek to learn the attention weight distribution ($\mathbf{\alpha}$) between $Q$ and $K$, and use it to compute the weighted context vector ($\mathbf{c}$) over $V$. In the above encoder-decoder architecture, $Q$ is the decoder hidden state $s_{i-1}$, and $K = V$ is the encoder hidden state $h_j$. Attention has been used to solve a variety of NLU tasks (MT, Question Answering, Text Classification, Sentiment Analysis), as well as Vision, Multi-Modal Tasks etc. \cite{Chaudhari2019}. We refer the reader to \cite{Chaudhari2019} for further details on the taxonomy of attention models. 

\begin{figure}[h]
  \centering
  \includegraphics[width=7.5cm]{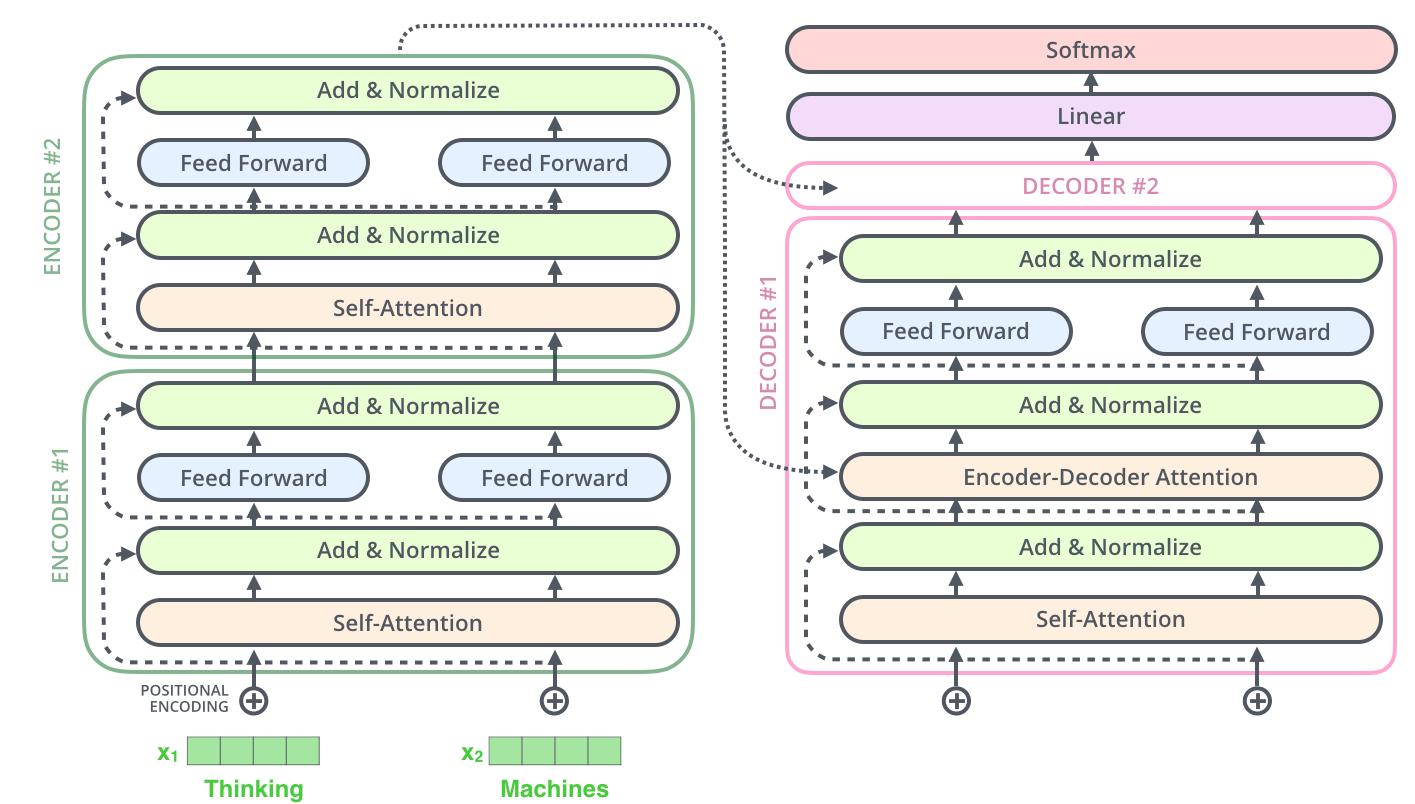}
  \caption{Transformer with its Encoder and Decoder blocks. Source: \cite{Alammar2021}.}
  \label{fig:transformer-encoder-decoder}
\end{figure}

The Transformer architecture \cite{Vaswani2017} was proposed in 2017, which introduced using Self-Attention layers for both the Encoder and the Decoder. They demonstrated that Attention layers could be used to replace traditional RNN based Seq2Seq models. The Self-Attention layer the query, key, and value vectors are all derived from the same sequence by using different projection matrices. 

Self-Attention also allows parallelizing the process of deriving relationships between the tokens in the input sequences. RNNs inherently force the process to occur one step at a time, i.e., learning long range dependencies is $O(n)$, where $n$ is the number of tokens. With Self-Attention, all tokens are processed together and pairwise relationships can be learnt in $O(1)$ \cite{Vaswani2017}. This makes it easier to leverage optimized training devices like GPUs and TPUs. The authors reported up to $300\times$ less training FLOPs as required to converge to a similar quality when compared to other recurrent and convolutional models. Tay et al. \cite{Tay2020} discuss the computation and memory efficiency of several Transformer variants and their underlying self-attention mechanisms in detail.

As introduced earlier, the BERT model architecture \cite{Devlin2018} beat the state-of-the-art in several NLU benchmarks. BERT is a stack of Transformer encoder layers that are pre-trained using a bi-directional masked language model training objective. It can also be used as a general purpose encoder which can then be used for other tasks. Other similar models like the GPT family \cite{Brown2020} have also been used for solving many NLU tasks.

\textbf{Random Projection Layers \& Models}
Pre-trained token representations such as word2vec \cite{Mikolov2017}, GLoVE \cite{Pennington2014}, etc. are common for NLU tasks. However, since they require a $d$-dimensional vector for storing each token, the total size consumed by the table quickly grows very large if the vocabulary size $V$ is substantial ($O(V.d)$).  

If model size is a constraint for deployment, we can either rely on compression techniques (as illustrated earlier) to help with Embedding Table compression, or evaluate layers and models that can work around the need for embedding tables. Random Projection based methods \cite{Ravi2017,Ravi2018,Kaliamoorthi2019,Kaliamoorthi2021} are one such family of models that do so. They propose replacing the embedding table and lookup by mapping the input feature $x$ (unicode token / word token, etc.), into a lower dimensional space. This is done using the random projection operator $\mathbb{P}$, such that $\mathbb{P}(x) \in \{0,1\}^{T.r}$, which can be decomposed into $T$ individual projection operations each generating an $r$-bit representation ($\mathbb{P}(x) = [\mathbb{P}_1(x), ..., \mathbb{P}_T(x)]$, where $\mathbb{P}_i(x) \in {0,1}^{r}$). $T$ and $r$ can be manually chosen.

Each random projection operation $\mathbb{P}_i$ is implemented using Locality Sensitive Hashing (LSH) \cite{Charikar2002,Ravi2018}, each using a different hash function (via different seeds). For theoretical guarantees about the Random Projection operation, refer to \cite{Charikar2002}, which demonstrates that the operation preserves the similarity between two points in the lower-dimensional space it maps these points to (this is crucial for the model to be learn the semantics about the inputs). If this relationship holds in the lower-dimensional space, the projection operation can be used to learn discriminative features for the given input. The core-benefit of the projection operation when compared to embedding tables is $O(T)$ space required instead of $O(V.d)$ ($T$ seeds required for $T$ hash functions). On the other hand, random-projection computation is $O(T)$ too v/s $O(1)$ for embedding table lookup. Hence, the projection layer is clearly useful when model size is the primary focus of optimization.

Across the various papers in the projection model family, there are subtle differences in implementation (computing complex features before (\cite{Ravi2018}) v/s after the projection operation (\cite{Kaliamoorthi2019,Sankar2019}), generating a ternary representation instead of binary (\cite{Kaliamoorthi2019,Kaliamoorthi2021}), applying complex layers and networks on top like Attention (\cite{Kaliamoorthi2019}), QRNN (\cite{Kaliamoorthi2021})), etc. 

\begin{figure}%
    \centering
    \subfloat[\centering PRADO Model. Source: \cite{Kaliamoorthi2019}.]{{\includegraphics[width=4cm]{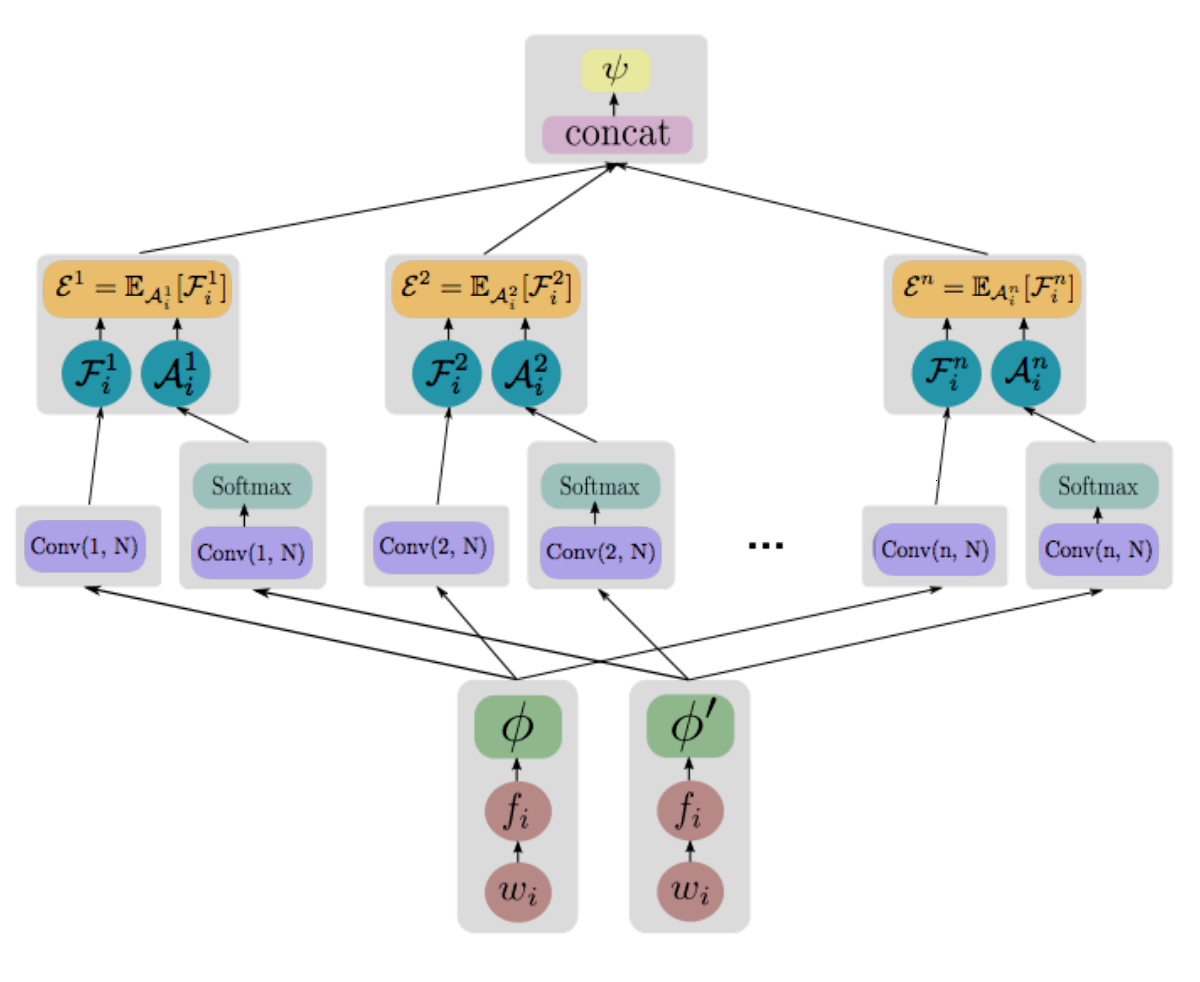} }}
    \subfloat[\centering PQRNN Model. Source: \cite{Kaliamoorthi2021}]{{\includegraphics[width=4cm]{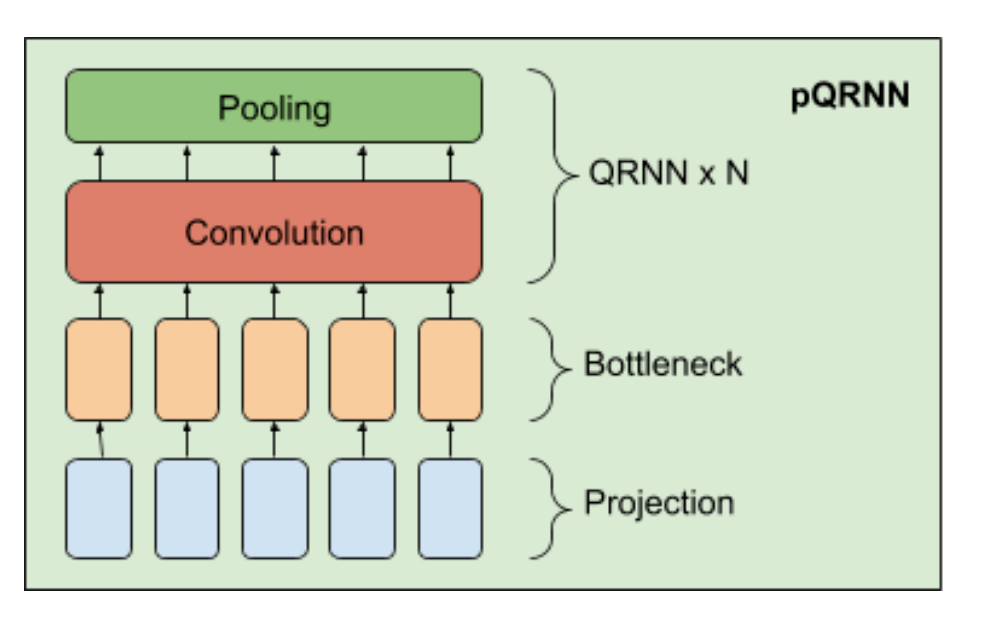} }}
    \subfloat[\centering Proformer Model. Source: \cite{Sankar2020}.]{{\includegraphics[width=3.5cm]{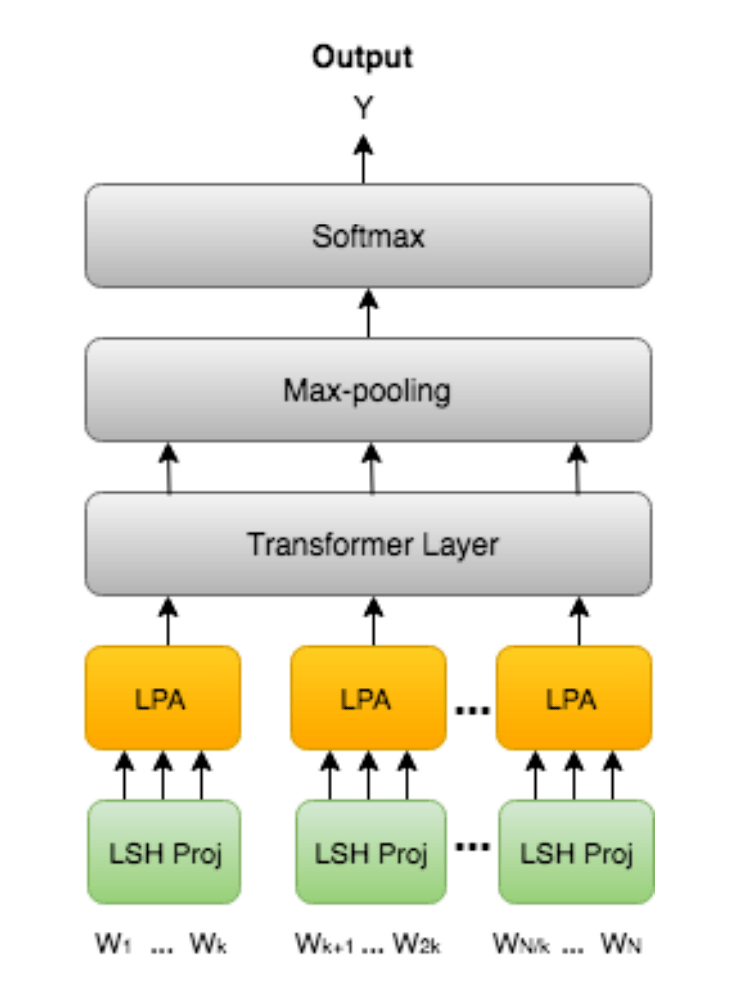} }}
    \caption{Collection of notable Random-Projection based models.}
    \label{fig:projection}
\end{figure}

Some of the Projection-based models (refer to Figure \ref{fig:projection}) have demonstrated impressive results on NLU tasks. PRADO (\cite{Kaliamoorthi2019}) generates n-gram features from the projected inputs, followed by having a Multi-Headed Attention layer on top. It achieved accuracies comparable to standard LSTM models, while being ~$100\times$ smaller, and taking 20-40 ms for inference on a Nexus 5X device. PQRNN \cite{Kaliamoorthi2021}, another Projection-based model that additionally uses a fast RNN implementation (QRNN) \cite{Bradbury2016} on top of the projected features. They report outperforming LSTMs while being $140\times$ smaller, and achieving $97.1\%$ of the quality of a BERT-like model while being $350\times$ smaller.

Proformer \cite{Sankar2020} introduces a Local Projected Attention (LPA) Layer, which combines the Projection operation with localized attention. They demonstrate  reaching $\approx$ 97.2\% BERT-base’s performance while occupying only 13\% of BERT-base’s memory. ProFormer also had 14.4 million parameters, compared to 110 million parameters of BERT-base.

\subsection{Infrastructure}
\begin{figure}[h]
  \centering
  \includegraphics[width=9cm]{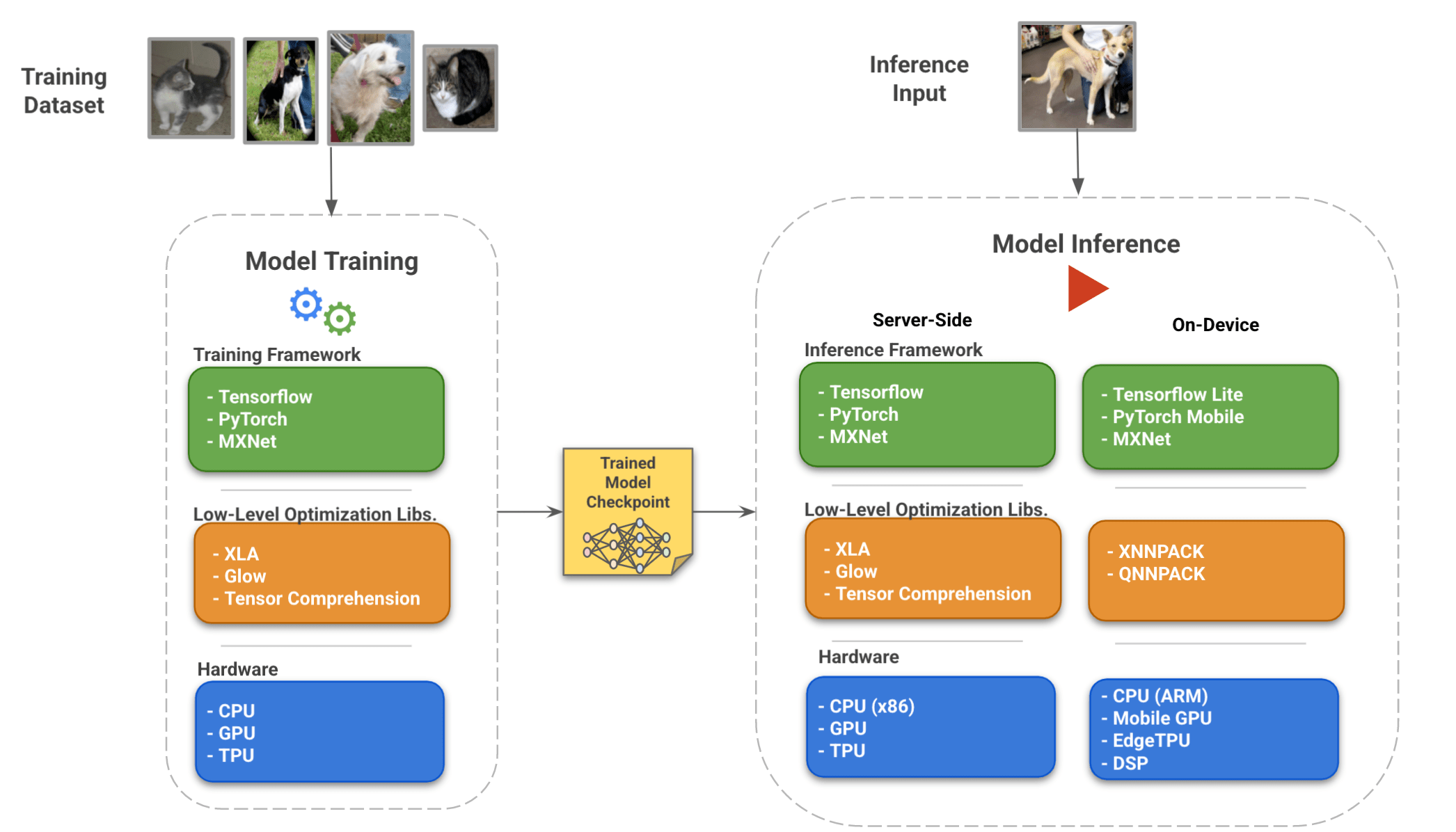}
  \caption{A visualization of the hardware and software infrastructure with emphasis on efficiency. On the left hand side is the model-training phase, which generates a trained model checkpoint. This model is then used on the inference side, which could either be server-side (conventional machines in cloud or on-prem), or on-device (mobile phones, IoT, edge devices, etc.).}
  \label{fig:infrastructure}
\end{figure}

In order to be able to train and run inference efficiently, there has to be a robust software and hardware infrastructure foundation. In this section we go over both these aspects. Refer to Figure \ref{fig:infrastructure} for a mental model of the software and hardware infrastructure, and how they interact with each other. In this section we provide a non-exhaustive but comprehensive survey of leading software and hardware infrastructure components that are critical to model efficiency.

\subsubsection{Tensorflow Ecosystem}
Tensorflow (TF) \cite{Abadi2016,TensorflowAuthors2021a} is a popular machine learning framework, that has been used in production by many large enterprises. It has some of the most extensive software support for model efficiency.

\textbf{Tensorflow Lite for On-Device Usecases}: Tensorflow Lite (TFLite) \cite{TensorflowAuthors2021b} is a collection of tools and libraries designed for inference in low-resource environments. At a high-level we can break down the TFLite project into two core parts:

\begin{itemize}
    \item \textbf{Interpreter and Op Kernels}: TFLite provides an interpreter for running specialized TFLite models, along with implementations of common neural net operations (Fully Connected, Convolution, Max Pooling, ReLu, Softmax, etc. each of which as an \emph{Op}). The implementation of such an operation is known as an \emph{Op Kernel}. Both the interpreter and Op Kernels are primarily optimized for inference on ARM-based processors as of the time of writing this paper. They can also leverage smartphone DSPs such as Qualcomm’s Hexagon \cite{XNNPACKAuthors2021b} for faster execution. The interpreter also allows the user to set multiple threads for execution.
    \item \textbf{Converter}: The TFLite converter\ccite{TensorflowAuthors2021c} as the name suggests is useful for converting the given TF model into a single flatbuffer\ccite{FlatBufferAuthors2021} file for inference by the interpreter. Apart from the conversion itself, it handles a lot of internal details like getting a graph ready for quantized inference, fusing operations\ccite{TensorflowAuthors2021d}, adding other metadata to the model, etc. With respect to quantization, it also allows post-training quantization as mentioned earlier\ccite{TensorflowAuthors2021f} with an optional representative dataset to improve accuracy.
\end{itemize}

\textbf{Other Tools for On-Device Inference}: TF Micro \cite{Warden2019}\ccite{TensorflowAuthors2021e} goes further, and consists of a slimmed down interpreter, and a smaller set of ops for inference on very low resource microcontrollers. TF Model Optimization toolkit \cite{TensorflowAuthors2021g} is a Tensorflow library for applying common compression techniques like quantization, pruning, clustering etc. TensorflowJS (TF.JS)\ccite{TensorflowAuthors2021h} is a library within the TF ecosystem that can be used to train and run neural networks within the browser or using Node.js \cite{NodeJsAuthors2021}. These models can also accelerated through GPUs via the WebGL interface \cite{ContributorstoWikimediaprojects2021a}. It supports both, importing models trained in TF, as well as creating new models from scratch in TF.JS.

\textbf{XLA for Server-Side Acceleration}: Typically a TF model graph is executed by TF's executor process and it uses standard optimized kernels for running it on CPU, GPU, etc. XLA \cite{TensorflowAuthors2021i} is a graph compiler that can optimize linear algebra computations in a model, by generating new kernels that are customized for the graph. These kernels are optimized for the model graph in question. For example, certain operations which can be fused together are combined in a single composite op. This avoids having to do multiple costly writes to RAM, when the operands can directly be operated on while they are still in cheaper caches. Kanwar et al. \cite{Kanwar2021a} report a 7$\times$ increase in training throughput, and 5$\times$ increase in the maximum batch size that can be used for BERT training. This allows training a BERT model for \$32 on Google Cloud.

\subsubsection{PyTorch Ecosystem}
PyTorch \cite{Paszke2019} is another popular machine-learning platform actively used by both academia and industry. It is often compared with Tensorflow in terms of usability and features.

\textbf{On-Device Usecases}: PyTorch also has a light-weight interpreter that enables running PyTorch models on Mobile \cite{PyTorchAuthors2021a}, with native runtimes for Android and iOS. This is analogous to the TFLite interpreter and runtime as introduced earlier. Similar to TFLite, PyTorch offers post-training quantization \cite{PyTorchAuthors2021b}, and other graph optimization steps such as constant folding, fusing certain operations together, putting the channels last (NHWC) format for optimizing convolutional layers\ccite{PyTorchAuthors2021c}.

\textbf{General Model Optimization}: PyTorch also offers the Just-in-Time (JIT) compilation facility \cite{PyTorchAuthors2021d}, which might seem similar to Tensorflow's XLA, but is actually a mechanism for generating a serializable intermediate representation (high-level IR, per \cite{Li2020}) of the model from the code in TorchScript \cite{PyTorchAuthors2021d}, which is a subset of Python. TorchScript adds constraints on the code that it can convert, such as type-checks, which allows it to sidestep some pitfalls of typical Python programming, while being Python compatible. It allows creating a bridge between the flexible PyTorch code for research and development, to a representation that can be deployed for inference in production. For example, exporting to TorchScript is a requirement to run on mobile devices \cite{PyTorchAuthors2021a}. This representation is analogous to the static inference mode graphs generated by TensorFlow. The alternative for XLA in the PyTorch world seem to be the Glow \cite{Rotem2018} and TensorComprehension \cite{Vasilache2018} compilers. They help in generating the lower-level intermediate representation that is derived from the higher-level IR (TorchScript, TF Graph). These low-level deep learning compilers are compared in detail in \cite{Li2020}.

PyTorch offers a model tuning guide \cite{PyTorchAuthors2021e}, which details various options that ML practitioners have at their disposal. Some of the core ideas in there are:
\begin{itemize}
    \item Turn on mixed-precision training \cite{PyTorchAuthors2021f} when using NVIDIA GPUs. This is described further in detail in the GPU sub-section in 3.5.4.
    \item Fusion of pointwise-operations (add, subtract, multiply, divide, etc.) using PyTorch JIT. Even though this should happen automatically, but adding the \texttt{torch.jit.script} decorator to methods which are completely composed of pointwise operations can force the TorchScript compiler to fuse them.
    \item Enabling buffer checkpointing allows keeping the outputs of only certain layers in memory, and computing the rest during the backward pass. This specifically helps with cheap to compute layers with large outputs like activations. A reduced memory usage can be exchanged for a larger batch size which improves utilization of the training platform (CPU, GPU, TPU, etc.). 
    \item Enabling device-specific optimizations, such as the cuDNN library, and Mixed Precision Training with NVIDIA GPUs (explained in the GPU subsection).
    \item Train with Distributed Data Parallel Training, which is suitable when there is a large amount of data and multiple GPUs are available for training. Each GPU gets its own copy of the model and optimizer, and operates on its own subset of the data. Each replicas gradients are periodically accumulated and then averaged.
\end{itemize}

\subsubsection{Hardware-Optimized Libraries}
We can further extract efficiency by optimizing for the hardware the neural networks run on. A prime deployment target is ARM's Cortex-family of processors. Cortex supports SIMD (Single-Instruction Multiple Data) instructions via the Neon \cite{Arm2021} architecture extension. SIMD instructions are useful for operating upon registers with vectors of data, which are essential for speeding up linear algebra operations through vectorization of these operations. QNNPACK \cite{Dukhan2020} and XNNPACK \cite{XNNPACKAuthors2021a} libraries are optimized for ARM Neon for mobile and embedded devices, and for x86 SSE2, AVX architectures, etc. QNNPACK supports several common ops in quantized inference mode for PyTorch. XNNPACK supports 32-bit floating point models and 16-bit floating point for TFLite. If a certain operation isn’t supported in XNNPACK, it falls back to the default implementation in TFLite.

Similarly, there are other low-level libraries like Accelerate for iOS \cite{AppleAuthors2021}, and NNAPI for Android \cite{AndroidDevelopers2021a} that try to abstract away the hardware-level acceleration decision from higher level ML frameworks. 

\subsubsection{Hardware}

\textbf{GPU}: Graphics Processing Units (GPUs) were originally designed for acclerating computer graphics, but began to be used for general-purpose usecases with the availability of the CUDA library \cite{ContributorstoWikimediaprojects2021e} in 2007, and libraries like like cuBLAS for speeding up linear algebra operations. In 2009, Raina et al. \cite{Raina2009} demonstrated that GPUs can be used to accelerate deep learning models. In 2012, following the AlexNet model's \cite{Krizhevsky2012} substantial improvement over the next entrant in the ImageNet competition further standardized the use of GPUs for deep learning models. Since then Nvidia has released several iterations of its GPU microarchitectures with increasing focus on deep learning performance. It has also introduced Tensor Cores \cite{NVIDIA2021a, Stosic2020} which are dedicated execution units in their GPUs, which are specialized for Deep Learning applications. TensorCores support training and inference in a range of precisions (fp32, TensorFloat32, fp16, bfloat16, int8, int4). As demonstrated earlier in quantization, switching to a lower precision is not always a significant trade-off, since the difference in model quality might often be minimal.

\begin{figure}[h]
    \centering
  \includegraphics[width=8.5cm]{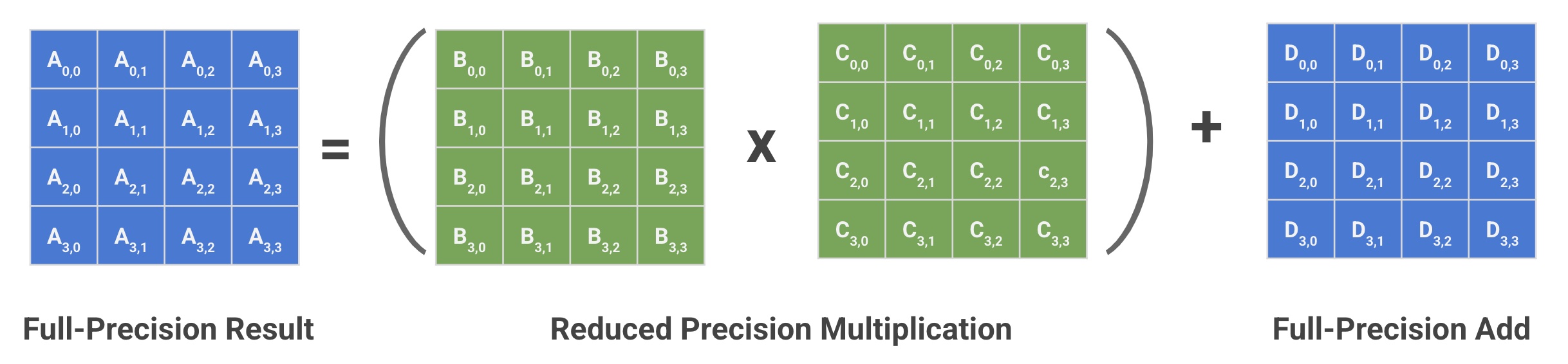}
  \caption{Reduced Precision Multiply-Accumulate (MAC) operation: An illustration of the $\mathbf{A} = (\mathbf{B} \times \mathbf{C}) + \mathbf{D}$ operation. $\mathbf{B}$ and $\mathbf{C}$ are in a reduced precision (fp16, bfloat16, TensorFloat32 etc.), while $\mathbf{A}$ and $\mathbf{D}$ are in fp32. The speedup comes from doing the expensive matrix-multiplication with a reduced precision format.}
  \label{fig:mac-reduced-precision}
\end{figure}

Tensor Cores optimize the standard Multiply-and-Accumulate (MAC) operation \cite{Wikimedia2021MAC}, $\mathbf{A} = (\mathbf{B} \times \mathbf{C}) + \mathbf{D}$. Where, $\mathbf{B}$ and $\mathbf{C}$ are in a reduced precision (fp16, bfloat16, TensorFloat32), while $\mathbf{A}$ and $\mathbf{D}$ are in fp32. The core speedup comes from doing the expensive matrix-multiplication in a lower-precision. The result of the multiplication is in fp32, which can be relatively cheaply added with $\mathbf{D}.$ When training with reduced-precision, NVIDIA reports between 1$\times$ to 15$\times$ training speedup depending on the model architecture and the GPU chosen \cite{Stosic2020}. Tensor Cores in NVidia's latest Ampere architecture GPUs also support faster inference with sparsity (specifically, structured sparsity in the ratio 2:4, where 2 elements out of a block of 4 elements are sparse) \cite{NVIDIA2021b}. They demonstrate an up to 1.5$\times$ speed up in inference time, and up to 1.8$\times$ speedup in individual layers. NVIDIA also offers the cuDNN libary \cite{NVIDIA2021b} that contains optimized versions of standard neural network operations such as fully-connected, convolution, batch-norm, activation, etc.

\begin{figure}[h]
  \centering
  \includegraphics[width=7cm]{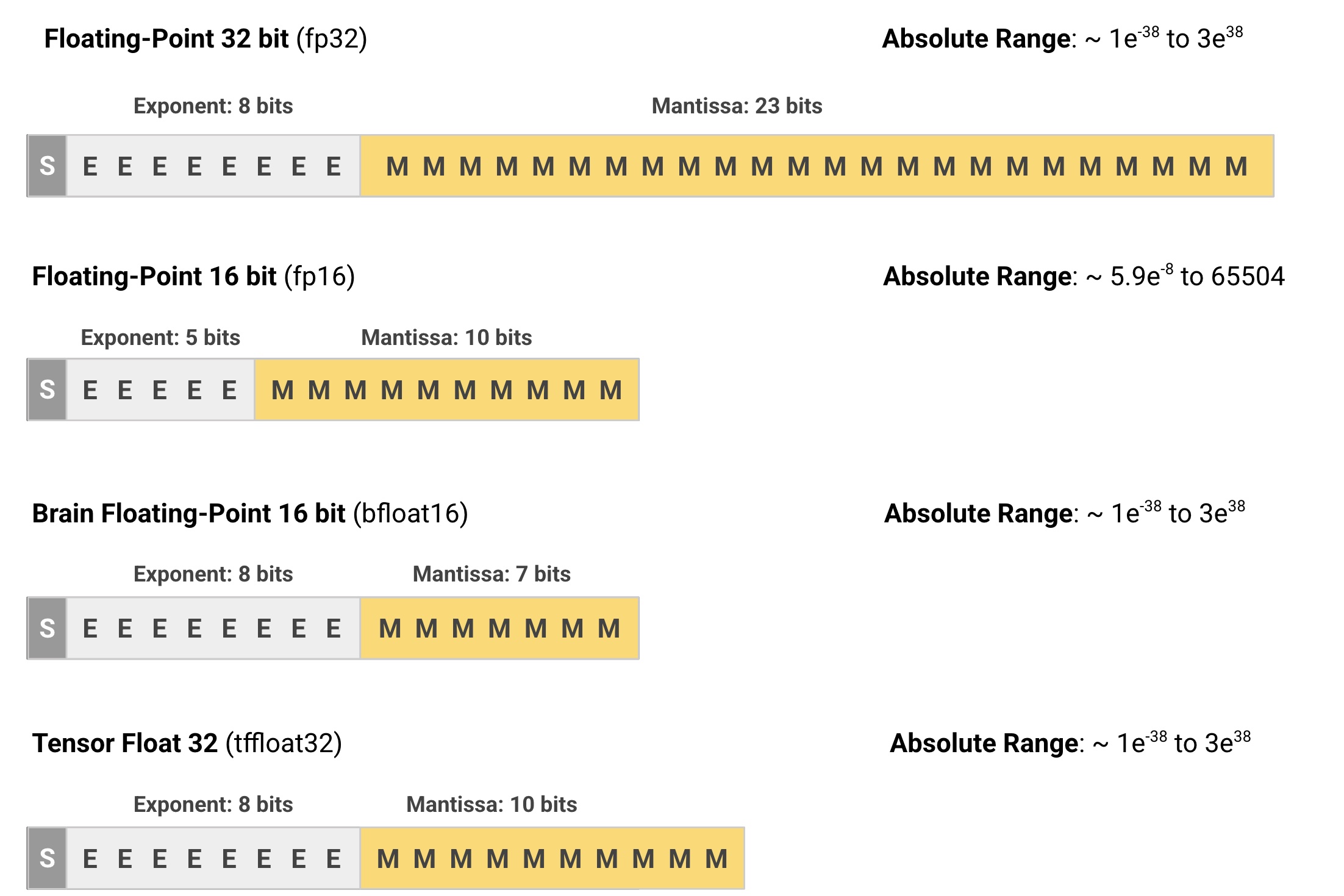}
  \caption{Common floating point format used in Training \& Inference: \texttt{fp32} is the standard 32-bit floating point number from IEEE-754 standard \cite{Wang2021a}. One bit is allocated for storing the sign. The exponent controls the range of the floating point value that can be expressed with that format, and the mantissa controls the precision. Note that \texttt{fp16} reduces the precision as well as range. The \texttt{bfloat16} format is a reasonable compromise because it keeps the same range as \texttt{fp32} while trading of precision to take up a total of 16 bits. NVidia GPUs also support Tensor Float 32 format that allocates 3 more bits to the mantissa than \texttt{bfloat16} to achieve better precision. However, it takes up a total of 19 bits which does not make it a trivially portable format.}
  \label{fig:data-types}
\end{figure}

\textbf{TPU}: TPUs are proprietary application-specific integrated circuits (ASICs) that Google has designed to accelerate deep learning applications with Tensorflow. Because they are not general purpose devices, they need not cater for any non-ML applications (which most GPUs have had to), hence they are finely tuned for parallelizing and accelerating linear algebra operations. The first iteration of the TPU was designed for inference with 8-bit integers, and was being used in Google for a year prior to their announcement in 2016 \cite{Jouppi2017}. Subsequent iterations of the TPU architectures enabled both training and inference with TPUs in floating point too. Google also opened up access to these TPUs via their Google Cloud service in 2018 \cite{Google2021a}.

\begin{figure}%
    \centering
    \subfloat[\centering A Systolic Array Cell implementing a Multiply-Accumulate (MAC) operation.]{{\includegraphics[width=4cm]{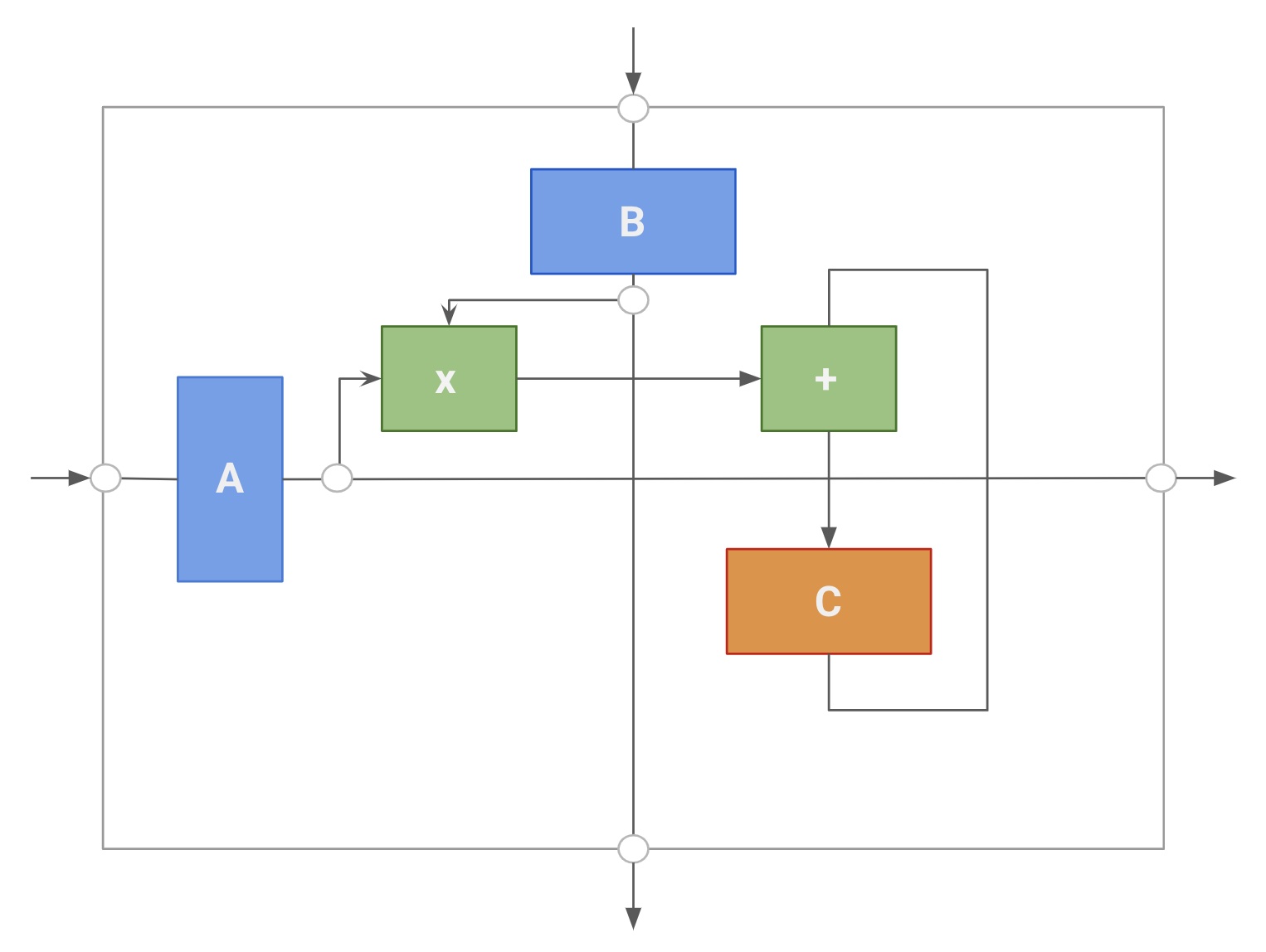} }}%
    \qquad
    \subfloat[\centering 4x4 Matrix Multiplication using Systolic Array]{{\includegraphics[width=5cm]{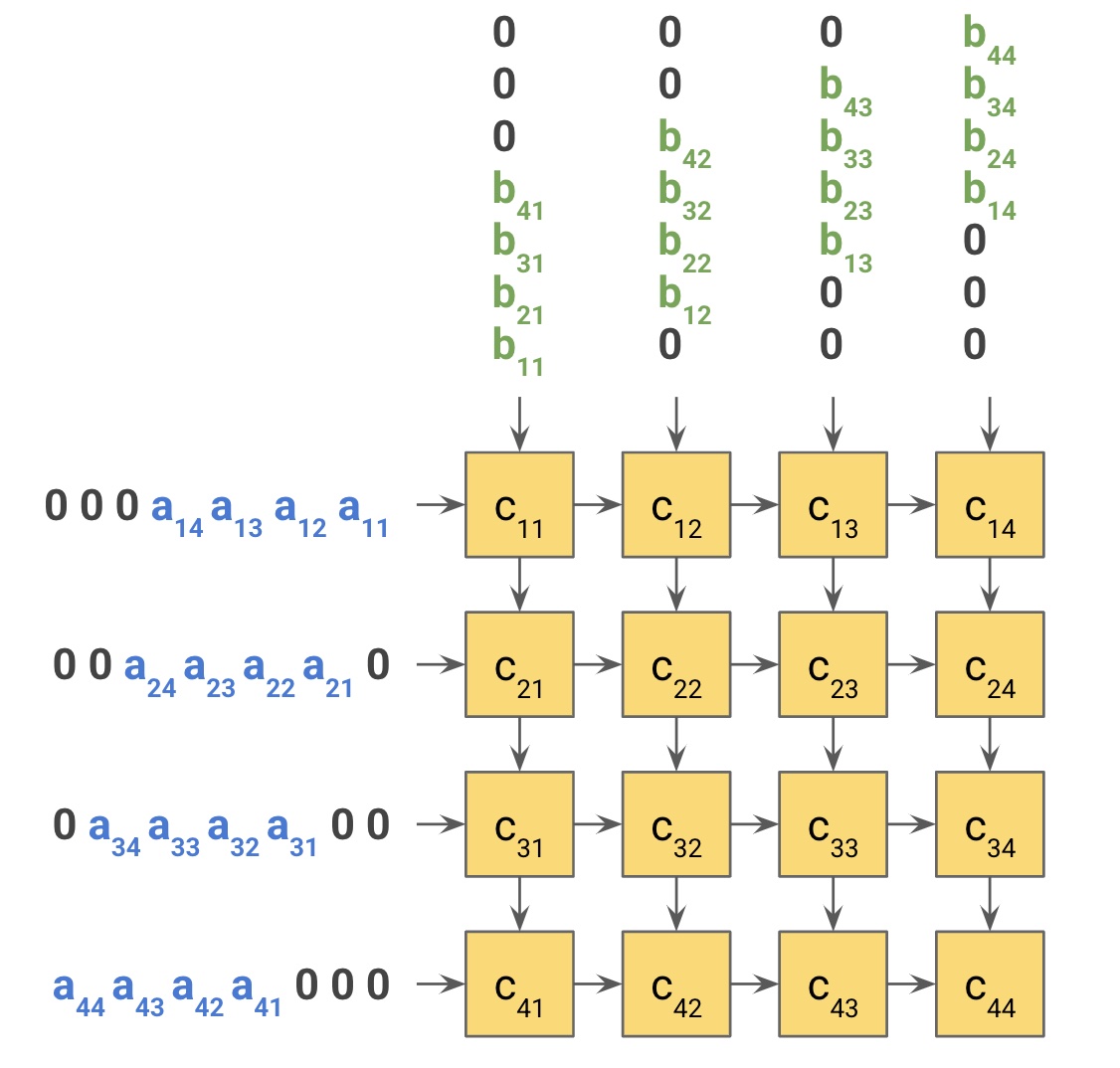} }}%
    \caption{Systolic Arrays in TPUs: Figure (a) shows a Systolic Array implementing a MAC operation, where the variables $A$ and $B$ are received by the cell, and $C$ is the resident memory.  $A$ is passed to the horizontally adjacent cell on the right, and $B$ is passed to the vertically adjacent cell below on the next clock tick. Figure (b) demonstrates how two 4$\times$4 matrices are multiplied using Systolic Arrays which is a mesh of cells constructed in Figure (a). The $i$-th row of array is fed the $i$-th column of $A$ (preceded by $i - 1$ 0s, which act as a delay). Similarly, the $i$-th column of the array is fed the $i$-th column of $B$ (preceded by $i - 1$ 0s). The corresponding $a_{ij}$ and $b_{jk}$ are passed to the neighboring cells on the next clock tick.}%
    \label{fig:systolic-array}%
\end{figure}

The core architecture of the TPU chips leverages the Systolic Array design \cite{Kung1980,Kung1982} (refer to Figure \ref{fig:systolic-array}), where a large computation is split across a mesh-like topology, where each cell computes a partial result and passes it on to the next cell in the order, every clock-step (in a rhythmic manner analogous to the systolic cardiac rhythm). Since there is no need to access registers for the intermediate results, once the required data is fetched the computation is not memory bound. Each TPU chip has two Tensor Cores (not to be confused with NVidia's Tensor Cores), each of which has a mesh of systolic arrays. There are 4 inter-connected TPU chips on a single TPU board. To further scale training and inference, a larger number of TPU boards can be connected in a mesh topology to form a 'pod'. As per publicly released numbers, each TPU chip (v3) can achieve 420 teraflops, and a TPU pod can reach 100+ petaflops \cite{Sato2021}.

TPUs have been used inside Google for applications like training models for Google Search, general purpose BERT models \cite{Devlin2018}, for applications like DeepMind's world beating AlphaGo and AlphaZero models \cite{Schrittwieser2020}, and many other research applications \cite{Tan2019}. They have also set model training time records in the MLPerf benchmarks. Similar to the GPUs, TPUs support the bfloat16 data-type \cite{Wang2021a} which is a reduced-precision alternative to training in full floating point 32-bit precision. XLA support allows transparently switching to bfloat16 without any model changes.

\textbf{EdgeTPU}: EdgeTPU is a custom ASIC chip designed by Google for running inference on edge devices, with low power requirements (4 Tera Ops / sec (TOPS) using 2 watts of power \cite{Google2021b}). Like the TPU, it is specialized for accelerating linear algebra operations, but only for inference and with a much lower compute budget. It is further limited to only a subset of operations \cite{Google2021c}, and works only with int8 quantized Tensorflow Lite models. Google releases the EdgeTPU using the Coral platform in various form-factors, ranging from a Raspberry-Pi like Dev Board to independent solderable modules \cite{Google2021d}. It has also been released with the Pixel 4 smartphones as the Pixel Neural Core \cite{Rakowski2019}, for accelerating on-device deep learning applications. The EdgeTPU chip itself is smaller than a US penny, making it amenable for deployment in many kinds of IoT devices.

\textbf{Jetson}: Jetson \cite{NVIDIA2021d} is a family of accelerators by Nvidia to enable deep learning applications for embedded and IoT devices. It comprises of the Nano, which is a low-powered "system on a module" (SoM) designed for lightweight deployments, as well as the more powerful Xavier and TX variants, which are based on the NVidia Volta and Pascal GPU architectures. As expected, the difference within the Jetson family is primarily the type and number of GPU cores on the accelerators. This makes the Nano suited for applications like home automation, and the rest for more compute intensive applications like industrial robotics.


\section{A Practitioner's Guide to Efficiency}

\begin{figure}[h]
  \centering
  \includegraphics[width=8.5cm]{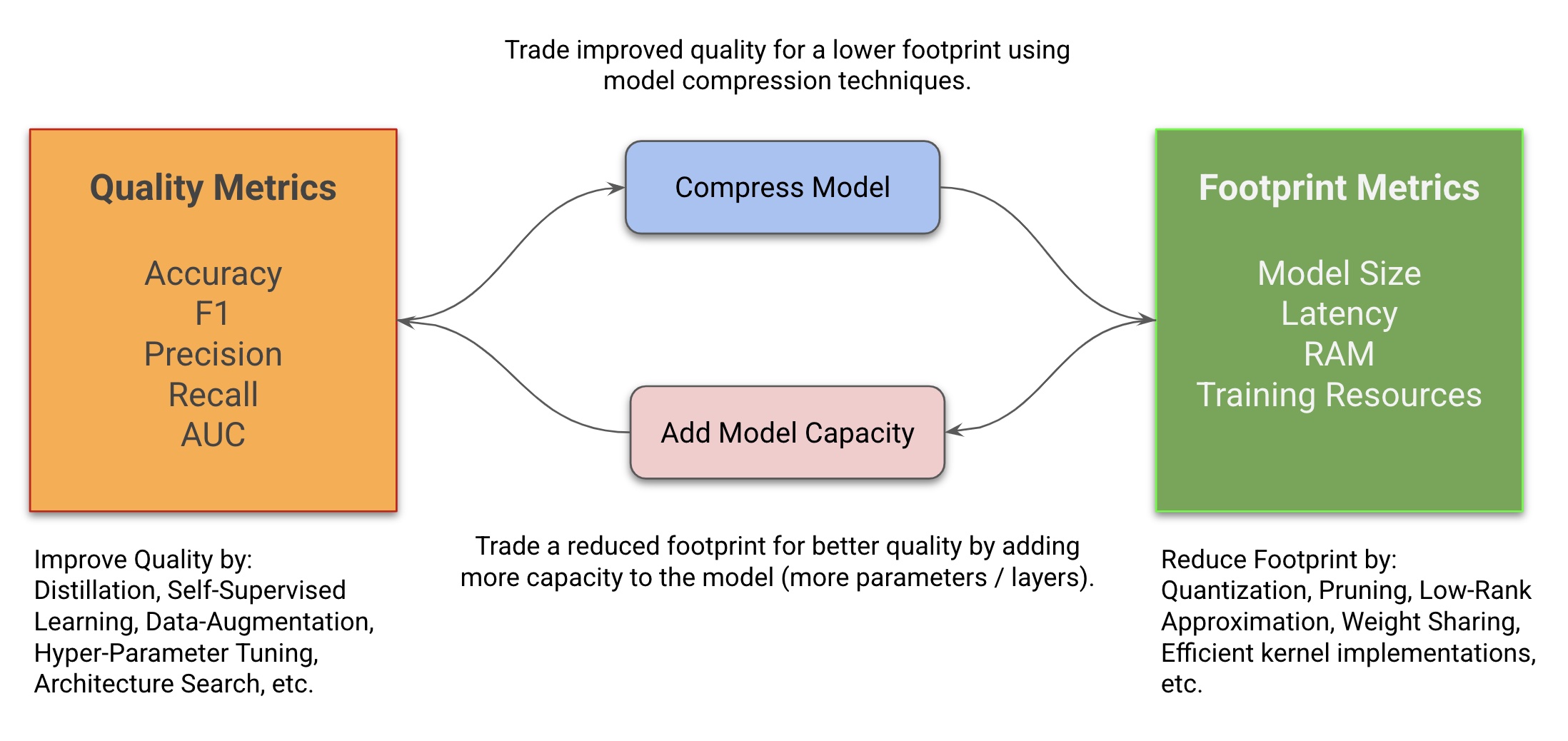}
  \caption{Trade off between Model Quality and Footprint: There exists a trade-off between model quality and model footprint. Model quality can be improved with techniques like distillation, data-augmentation, hyper-param tuning etc. Compression techniques can in turn help trade off some model quality for a better model footprint. Some / all of the improvement in footprint metrics can also be traded for better quality by simply adding more model capacity.}
  \label{fig:quality-compression-tradeoff}
\end{figure}

So far, we presented a broad set of tools and techniques in the Efficient Deep Learning landscape. In this section, we present a practical guide for practitioners to use, and how these tools and techniques work with each other. As mentioned earlier, what we seek are \emph{pareto-optimal} models, where we would like to achieve the best possible result in one dimension, while holding the other dimensions constant. Typically, one of these dimensions is \textbf{Quality}, and the other is \textbf{Footprint}. Quality related metrics could included Accuracy, F1, Precision, Recall, AUC, etc. While Footprint related metrics can include Model Size, Latency, RAM, etc.

Naturally, there exists a trade-off between Quality and Footprint metrics. A higher-capacity / deeper model is more likely to achieve a better accuracy, but at the cost of model size, latency, etc. On the other hand a model with lesser capcity / shallower, while possibly suitable for deployment, is also likely to be worse in accuracy. As illustrated in Figure \ref{fig:quality-compression-tradeoff}, we can traverse from a model with better quality metrics, and exchange some of the quality for better footprint by naively compressing the model / reducing the model capacity (\textbf{Shrink}). Similarly it is possible to naively improve quality by adding more capacity to the model (\textbf{Grow}). Growing can be addressed by the author of the model via appropriately increasing model capacity and tweaking other hyper-parameters to improve model quality. Shrinking can be achieved via Compression Techniques (Quantization, Pruning, Low-Rank Approximation, etc.), Efficient Layers \& Models, Architecture Search via Automation, etc. In addition, we can also \textbf{Improve} the quality metrics, while keeping the footprint same through Learning Techniques (Distillation, Data Augmentation, Self-Supervised Tuning), Hyper-Parameter Tuning, etc. (See Table \ref{tab:grow-shrink-improve} for more examples.)

\begingroup
\renewcommand{\arraystretch}{1.1} 
\begin{table}[]
\small
\begin{tabular}{l|l|l}
\hline
\multicolumn{1}{c|}{\begin{tabular}[c]{@{}c@{}}\textbf{Grow}\\ (Model Capacity)\end{tabular}} &
  \multicolumn{1}{c|}{\begin{tabular}[c]{@{}c@{}}\textbf{Shrink}\\ (Footprint)\end{tabular}} &
  \multicolumn{1}{c}{\begin{tabular}[c]{@{}c@{}}\textbf{Improve}\\ (Quality)\end{tabular}} \\ \hline
\multirow{4}{*}{\begin{tabular}[c]{@{}l@{}}Add layers, width, etc.\\ either manually or \\ using width / depth / \\ compound scaling \\ multipliers\end{tabular}} &
  \begin{tabular}[c]{@{}l@{}}Reduce layers, width, etc. \\ either manually  or using  \\ width / depth / compound \\ scaling multipliers\end{tabular} &
  \begin{tabular}[c]{@{}l@{}}Manual Tuning (Architecture / \\ Hyper-Parameters / \\ Features, etc.)\end{tabular} \\ \cline{2-3} 
 &
  \begin{tabular}[c]{@{}l@{}}\textbf{Compression Techniques}: \\ Quantization, Pruning, \\ Low-Rank Factorization, etc.\end{tabular} &
  \begin{tabular}[c]{@{}l@{}}\textbf{Learning Techniques}:\\ Data-Augmentation, Distillation, \\ Unsupervised Learning, etc.\end{tabular} \\ \cline{2-3} 
 &
  \begin{tabular}[c]{@{}l@{}}\textbf{Automation}:\\ Hyper-Param Optimization, \\ Architecture Search, etc.\end{tabular} &
  \begin{tabular}[c]{@{}l@{}}\textbf{Automation}:\\ Hyper-Param Optimization, \\ Architecture Search, etc.\end{tabular} \\ \cline{2-3} 
 &
  \begin{tabular}[c]{@{}l@{}}\textbf{Efficient Layers \& Models}:\\ Projection, PQRNN, (NLP),\\ Separable Convolution (Vision), \\ etc.\end{tabular} &
  \begin{tabular}[c]{@{}l@{}}\textbf{Efficient Layers \& Models}:\\ Transformers (NLP), \\ Vi-T (Vision), etc.\end{tabular} \\ \hline
\end{tabular}
\caption{Examples of techniques to use in the Grow, Shrink, and Improve phases.}
\label{tab:grow-shrink-improve}
\end{table}
\endgroup

Combining these three phases, we propose two strategies towards achieving pareto-optimal models:

\begin{enumerate}
    \item \textbf{Shrink-and-Improve for Footprint-Sensitive Models}: If as a practitioner, you want to reduce your footprint, while keeping the quality the same, this could be a useful strategy for on-device deployments and server-side model optimization. Shrinking should ideally be minimally lossy in terms of quality (can be achieved via learned compression techniques, architecture search etc.), but in some cases even naively reducing capacity can also be compensated by the Improve phase. It is also possible to do the Improve phase before the Shrink phase.
    \item \textbf{Grow-Improve-and-Shrink for Quality-Sensitive Models}: When you want to deploy models that have better quality while keeping the same footprint, it might make sense to follow this strategy. Here, the capacity is first added by growing the model as illustrated earlier. The model is then improved using via learning techniques, automation, etc. and then shrunk back either naively or in a learned manner. Alternatively, the model could be shrunk back either in a learned manner directly after growing the model too.
\end{enumerate}

We consider both these strategies as a way of going from a potentially non pareto-optimal model to another one that lies on the pareto-frontier with the trade-off that is appropriate for the user. Each efficiency technique individually helps move us closer to that target model.


\somecomment{

Modelling Techniques:
- Quality Impact v/s Footprint v/s Engg Resources Required v/s Training Resources v/s Inference Resources

1. Bottlenecking on Data: How much time do resources spend waiting for data?
1. When training memory is reduced, train with a larger batch size.
1. Leverage device specific optimizations (cuDNN, AMP)
1. Avoid moving data between accelerators and device (TPUs).
1. Parallelism (data parallelism: have multiple devices processing different data, works best when model fits in the device memory. try model parallelism otherwise).
1. For smaller model, dropout and L2 regularization is sufficient for avoiding over-fitting. However, distillation helps.
1. For larger models distillation isn't very helpful, augmentation is good.
}

\subsection{Experiments}
In order to demonstrate what we proposed above, we undertook the task of going through the exercise of making a given Deep Learning model efficient. Concretely, we had the following goals with this exercise:
\begin{enumerate}
    \item Achieve a new pareto-frontier using the efficiency techniques. Hence, demonstrating that these techniques can be used in isolation as well as in combination with other techniques, in the real-world by ML Practitioners.
    \item With various combinations of efficiency techniques and model scaling, demonstrate the tradeoffs for both `Shrink-and-Improve', and `Grow-Improve-and-Shrink' strategies for discovering and traversing the pareto-frontier. In other words, provide empirical evidence that it is possible for practitioners to either reduce model capacity to bring down the footprint (shrink) and then recover the model quality that they traded off (improve), or increase the model capacity to improve quality (growing) followed by model compression (shrinking) to improve model footprint.
\end{enumerate}

We picked the problem of classifying images in the CIFAR-10 dataset \cite{Krizhevsky2009} on compute constrained devices such as smartphones, IoT devices etc. We designed a deep convolutional architecture where we could scale the model capacity up or down, by increasing or decreasing the `width multiplier' ($w$) value. In the implementation, $w$ scales the number of filters for the convolutional layers (except the first two). Hence, using different values of $w$ in $[0.1, 0.25, 0.5, 0.75, 1.0]$ we obtain a family of models with different quality and footprint tradeoffs. We trained these models with some manual tuning to achieve a baseline of quality v/s footprint metrics. In this case, we measured quality through accuracy, and footprint through number of parameters, model size, and latency. In terms of techniques, we used Quantization for Shrinking, and Data Augmentation and Distillation for Improving. Many other techniques could be used to further drive the point home (Automation such as Hyper-Parameter Tuning, Efficient Layers such as Separable Convolutions), but were skipped to keep the interpretation of the results simpler. We used the Tensorflow-backed Keras APIs \cite{Chollet2020} for training, and the TFLite \cite{TensorflowAuthors2021b} framework for inference. The latencies were measured on three kinds of devices, low-end (Oppo A5), mid-end (Pixel 3XL), and high-end (Galaxy S10), in order of their increasing CPU compute power. The model size numbers reported are the sizes of the generated TFLite models, and the latency numbers are the average single-threaded CPU latency after warmup on the target device. The code for the experiments is available via an IPython notebook  \href{https://github.com/reddragon/efficient-dl-survey-paper/blob/main/CIFAR_10_End_to_End.ipynb}{here}.

\begin{figure}
    \centering
    \subfloat[\centering Number of Params v/s Accuracy]{{\includegraphics[width=4.5cm]{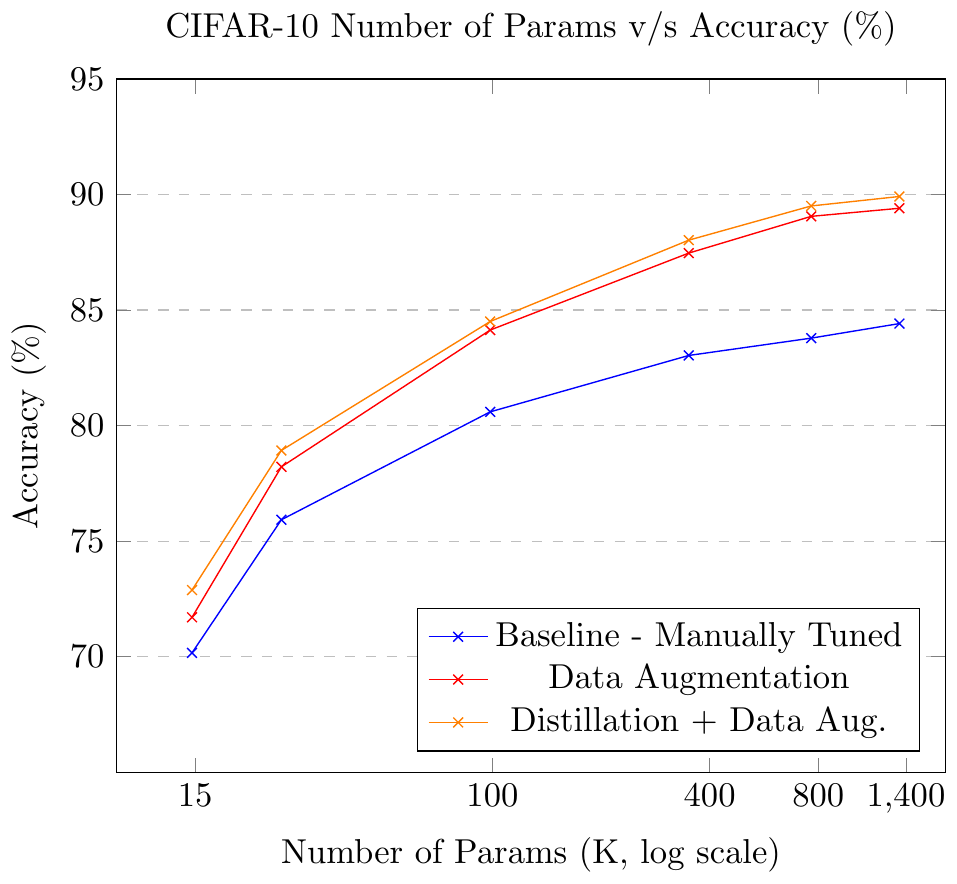} }}
    \qquad
    \subfloat[\centering Model Size v/s Accuracy]{{\includegraphics[width=4.55cm]{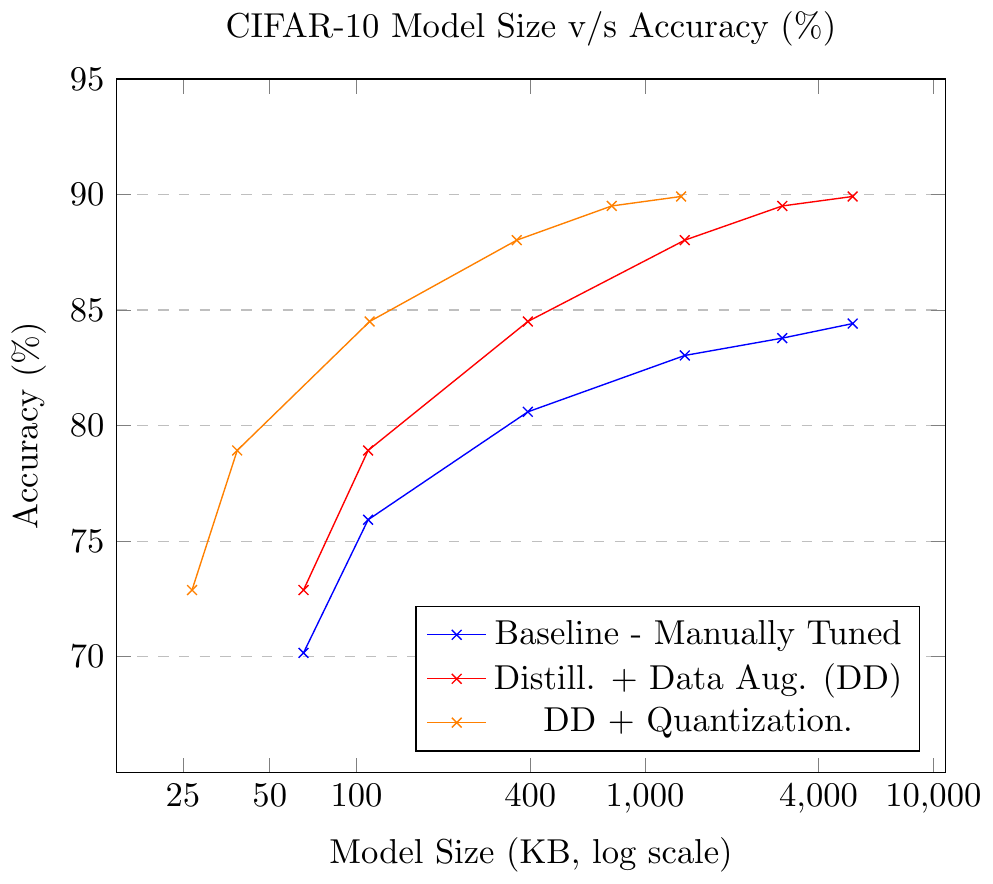} }}
    \caption{Change in Accuracy with respect to Number of Params and Model Size. Each point on a curve is a model from Table \ref{tab:float-model-cifar10} in figure (a) and from Table \ref{tab:quant-model-cifar10} in figure (b).}
    \label{fig:params-size}
\end{figure}

\begin{figure}
    \centering
    \subfloat[\centering Low-End Device Latency]{{\includegraphics[width=4.5cm]{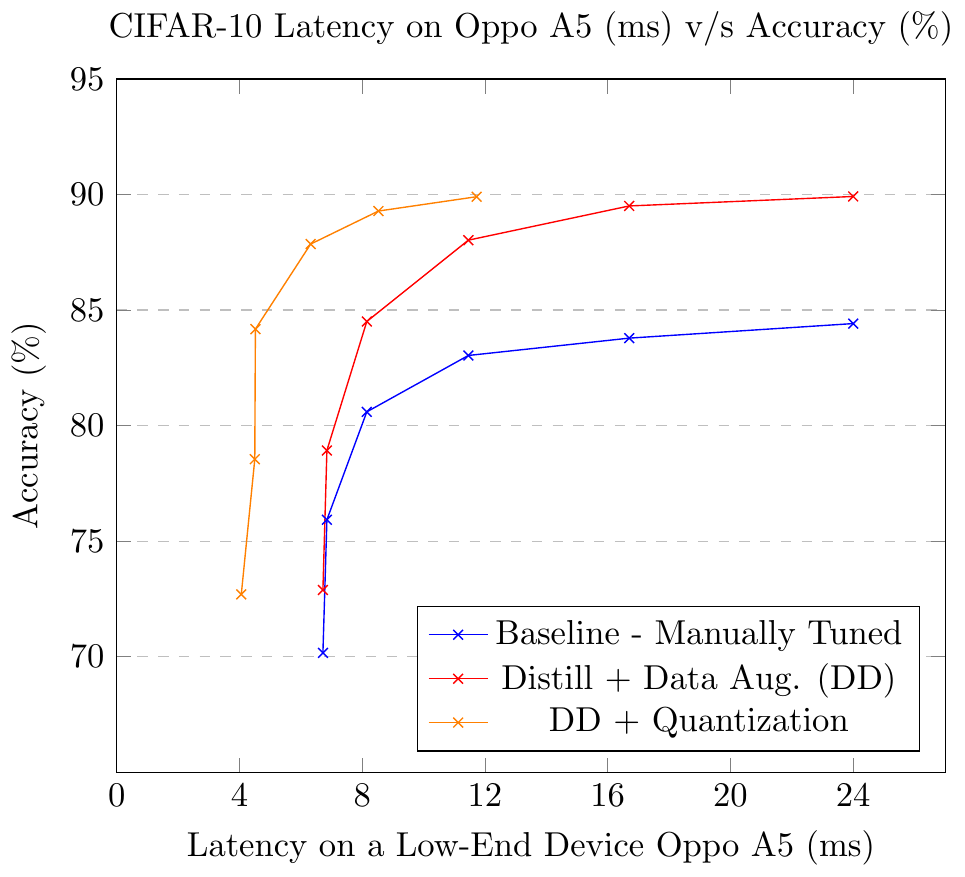} }}
    \subfloat[\centering Mid-Tier Device Latency]{{\includegraphics[width=4.5cm]{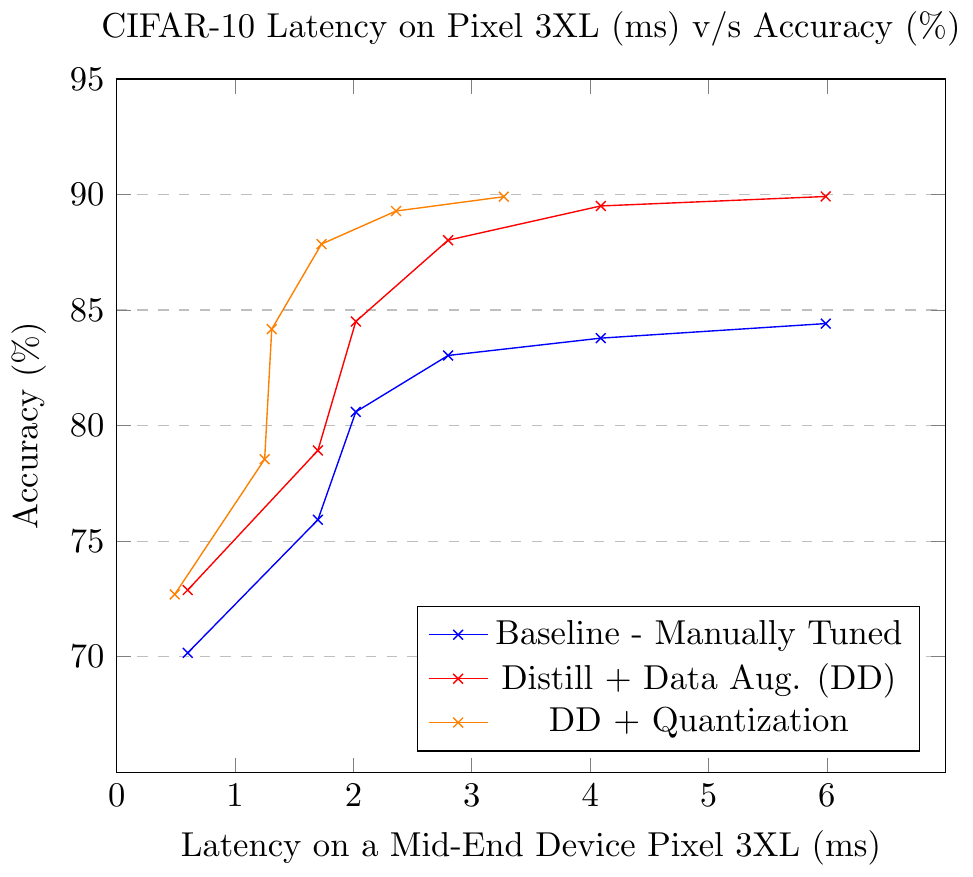} }}
    \subfloat[\centering High-End Device Latency]{{\includegraphics[width=4.5cm]{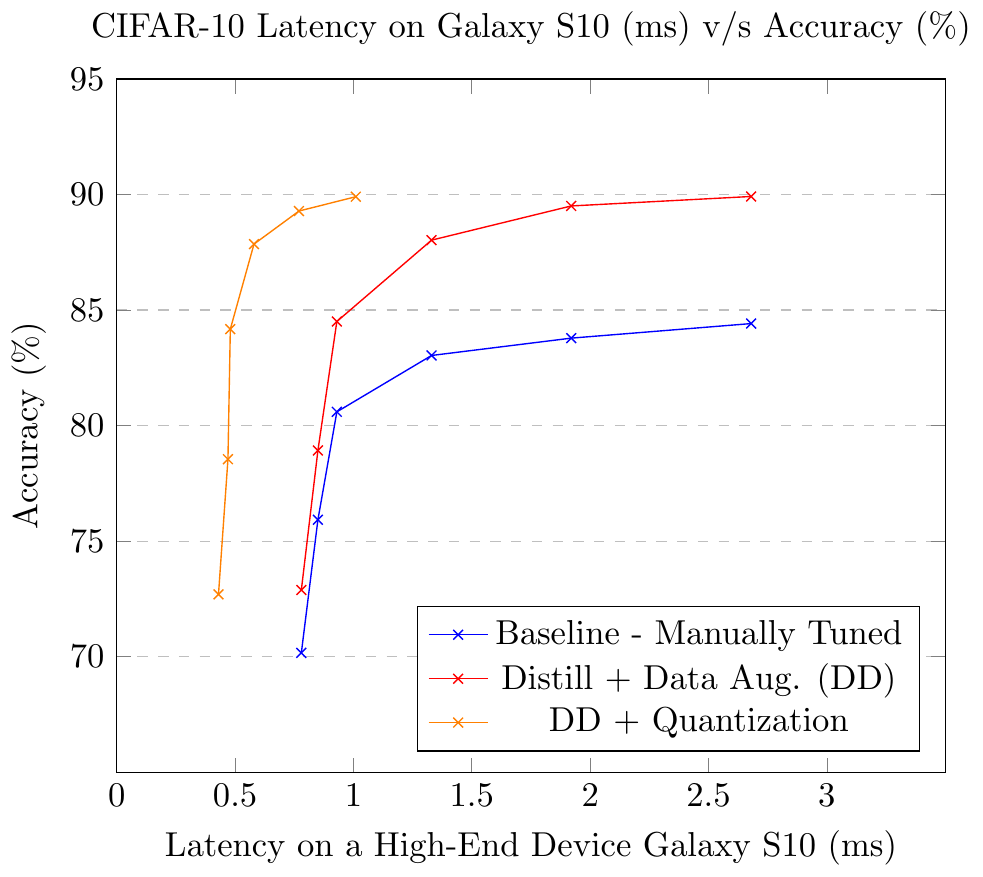} }}
    \caption{Average latency of models on different devices (low-, mid-, and high-end smartphones). The orange curve denotes the quantized models in addition to being trained with distillation and data augmentation.}
    \label{fig:latency-plots}
\end{figure}

Table \ref{tab:float-model-cifar10} compiles the results for 6 width-multipliers in increasing order, ranging from $0.05$ to $1.0$. Between the smallest to the largest models, the number of params grows by $\approx 91.4 \times$, and the model size grows by $\approx 80.2 \times$. The latency numbers also grow between $3.5 - 10 \times$ based on the device. Within the same row, footprint metrics will not change since we are not changing the model architecture. In Table \ref{tab:float-model-cifar10} we purely work with techniques that will improve the model quality (Data Augmentation and Distillation). Table \ref{tab:quant-model-cifar10} reports the numbers for the Quantized versions of the corresponding models in Table \ref{tab:float-model-cifar10}. We use Quantization for the Shrink phase, to reduce model size by $\approx 4 \times$, and reduce the average latency by $1.5 - 2.65 \times$. Figures \ref{fig:params-size} and \ref{fig:latency-plots} plot the notable results from Tables \ref{tab:float-model-cifar10} and \ref{tab:quant-model-cifar10}.

\begingroup
\begin{table}[]
\small
\centering
\begin{tabular}{ccccccccc}
\midrule
\multicolumn{1}{c}{\multirow{2}{*}{\begin{tabular}[c]{@{}c@{}}Width \\ Multiplier\end{tabular}}} &
  \multicolumn{1}{c}{\multirow{2}{*}{\begin{tabular}[c]{@{}c@{}}\# Params\\ (K)\end{tabular}}} &
  \multicolumn{1}{c}{\multirow{2}{*}{\begin{tabular}[c]{@{}c@{}}Model \\ Size\\ (KB)\end{tabular}}} &
  \multicolumn{3}{c}{\begin{tabular}[c]{@{}c@{}}Accuracy\\ (\%)\end{tabular}} &
  \multicolumn{3}{c}{Average Latency (ms)} \\
  \cmidrule(r){4-6}
  \cmidrule(r){7-9}
\multicolumn{1}{c}{} &
  \multicolumn{1}{c}{} &
  \multicolumn{1}{c}{} &
  \multicolumn{1}{c}{Baseline} &
  \multicolumn{1}{c}{Augmentation} &
  \multicolumn{1}{c}{\begin{tabular}[c]{@{}c@{}}Augmentation \\ + Distillation\end{tabular}} &
  \multicolumn{1}{c}{\begin{tabular}[c]{@{}c@{}}Oppo \\ A5\end{tabular}} &
  \multicolumn{1}{c}{\begin{tabular}[c]{@{}c@{}}Pixel \\ 3XL\end{tabular}} &
  \multicolumn{1}{c}{\begin{tabular}[c]{@{}c@{}}Galaxy \\ S10\end{tabular}} \\
\midrule \\
0.05 & 14.7    & 65.45   & 70.17 & 71.71 & 72.89 & 6.72  & 0.6  & 0.78 \\
0.1  & 26      & 109.61  & 75.93 & 78.22 & 78.93 & 6.85  & 1.7  & 0.85 \\
0.25 & 98.57   & 392.49  & 80.6  & 84.14 & 84.51 & 8.15  & 2.02 & 0.93 \\
0.5  & 350.05  & 1374.11 & 83.04 & 87.47 & 88.03 & 11.46 & 2.8  & 1.33 \\
0.75 & 764.87  & 2993.71 & 83.79 & 89.06 & 89.51 & 16.7  & 4.09 & 1.92 \\
1    & 1343.01 & 5251.34 & 84.42 & 89.41 & 89.92 & 24    & 5.99 & 2.68 \\
\midrule
\end{tabular}
\caption{Quality and Footprint metrics for Floating-Point models for the CIFAR-10 dataset.}
\label{tab:float-model-cifar10}
\end{table}
\endgroup

\begingroup
\setlength{\tabcolsep}{5pt} 
\renewcommand{\arraystretch}{0.9} 
\begin{table}[]
\small
\centering
\begin{tabular}{ccccccccc}
\midrule
\multicolumn{1}{c}{\multirow{2}{*}{\begin{tabular}[c]{@{}c@{}}Width \\ Multiplier\end{tabular}}} &
  \multicolumn{1}{c}{\multirow{2}{*}{\begin{tabular}[c]{@{}c@{}}\# Params\\ (K)\end{tabular}}} &
  \multicolumn{1}{c}{\multirow{2}{*}{\begin{tabular}[c]{@{}c@{}}Model \\ Size\\ (KB)\end{tabular}}} &
  \multicolumn{3}{c}{\begin{tabular}[c]{@{}c@{}}Accuracy\\ (\%)\end{tabular}} &
  \multicolumn{3}{c}{Average Latency (ms)} \\
  \cmidrule(r){4-6}
  \cmidrule(r){7-9}
\multicolumn{1}{c}{} &
  \multicolumn{1}{c}{} &
  \multicolumn{1}{c}{} &
  \multicolumn{1}{c}{Baseline} &
  \multicolumn{1}{c}{Augmentation} &
  \multicolumn{1}{c}{\begin{tabular}[c]{@{}c@{}}Augmentation \\ + Distillation\end{tabular}} &
  \multicolumn{1}{c}{\begin{tabular}[c]{@{}c@{}}Oppo \\ A5\end{tabular}} &
  \multicolumn{1}{c}{\begin{tabular}[c]{@{}c@{}}Pixel \\ 3XL\end{tabular}} &
  \multicolumn{1}{c}{\begin{tabular}[c]{@{}c@{}}Galaxy \\ S10\end{tabular}} \\
\midrule \\
0.05 & 14.7    & 26.87   & 69.9  & 71.72 & 72.7  & 4.06  & 0.49 & 0.43 \\
0.1  & 26      & 38.55   & 75.98 & 78.19 & 78.55 & 4.5   & 1.25 & 0.47 \\
0.25 & 98.57   & 111     & 80.76 & 83.98 & 84.18 & 4.52  & 1.31 & 0.48 \\
0.5  & 350.05  & 359.31  & 83    & 87.32 & 87.86 & 6.32  & 1.73 & 0.58 \\
0.75 & 764.87  & 767.09  & 83.6  & 88.57 & 89.29 & 8.53  & 2.36 & 0.77 \\
1    & 1343.01 & 1334.41 & 84.52 & 89.28 & 89.91 & 11.73 & 3.27 & 1.01 \\
\midrule
\end{tabular}
\caption{Quality and Footprint metrics for \emph{Quantized} models for the CIFAR-10 dataset. Each model is the quantized equivalent of the corresponding model in Table \ref{tab:float-model-cifar10}.}
\label{tab:quant-model-cifar10}
\end{table}
\endgroup

\subsection{Discussion}
Let us try to interpret the above data to validate if our strategies can be used practically.

\textbf{Shrink-and-Improve for Footprint-Sensitive Models}: Refer to Table \ref{tab:float-model-cifar10} and Figure \ref{fig:params-size}. If our goal was to deploy the model with Width Multiplier ($w$) = $1.0$ and accuracy $84.42\%$, but the bottleneck was the model size (5.25 MB) and latency on a low-end device (24 ms on Oppo A5). This is the classic case of the footprint metrics not meeting the bar, hence we could apply the Shrink-and-Improve strategy, by first naively scaling our model down to a Width Multiplier ($w$) of $0.25$. This smaller model when manually tuned, as seen in Table \ref{tab:float-model-cifar10}, achieves an accuracy of $80.76\%$. However, when we use a combination of Data Augmentation \& Distillation from a separately trained larger teacher model with an accuracy of $90.86\%$, the accuracy of the smaller model improves to $84.18\%$, very close to the target model that we want to deploy. The size of this smaller model is 392.49 KB, which is $13.8 \times$ smaller, and the latency is 8.15 ms, which is $2.94 \times$ faster at a comparable accuracy. It is possible to further compress this model by using Quantization for some additional shrinking. The same smaller model ($w = 0.25$) when Quantized in Table \ref{tab:quant-model-cifar10}, is 111 KB in size (\textbf{$47.3 \times$ smaller}) and has a latency of 4.52 ms (\textbf{$5.31 \times$ faster}), while retaining an accuracy of $84.18\%$. It is possible to do this for other pairs of points on the curves.

\textbf{Grow-Improve-Shrink for Quality-Sensitive Models}: Assuming our goal is to deploy a model that has footprint metrics comparable to the model with $w = 0.25$ (392.49 KB model size, 0.93 ms on a high-end Galaxy S10 device), but an accuracy better than the baseline $80.6\%$ (refer to Table \ref{tab:float-model-cifar10}). In this case, we can choose to first grow our model to $w = 0.5$. This instantly blows up the model size to 1.37 MB ($3.49 \times$ bigger), and latency to 1.33 ms ($1.43 \times$ slower). However, we ignore that for a bit and improve our model's quality to $88.03\%$ with Data Augmentation \& Distillation. Then using Quantization for shrinking (refer to Table \ref{tab:quant-model-cifar10}), we can get a model that is 359.31 KB in size (32 KB smaller) and has a 0.58 ms latency on Galaxy S10 ($1.6 \times$ faster), with an accuracy of $87.86\%$, an absolute 7.10\% increase in accuracy while keeping the model size approximately same and making it $1.6 \times$ faster. It is also possible to apply this strategy to other pairs of models.

Thus, we've verified that the above two strategies can work both ways, whether your goal is to optimize for quality metrics or footprint metrics. We were also able to visually inspect through Figures \ref{fig:params-size} and \ref{fig:latency-plots} that efficiency techniques can improve on the pareto frontiers constructed through manual tuning. To contain the scope of experimentation, we selected two sets of efficiency techniques (Compression Techniques (Quantization), and Learning Techniques (Data Augmentation \& Distillation). Hence, it would be useful to explore other techniques as well such as Automation (for Hyper-Parameter Tuning to further improve on results), and Efficient Layers \& Models (Separable Convolution as illustrated in MobileNet \cite{Sandler2018} could be used in place of larger convolutional layers). Finally, we would also like to emphasize paying attention to performance of Deep Learning models (optimized or not) on underrepresented classes and out-of-distribution data to ensure model fairness, since quality metrics alone might not be sufficient for discovering deeper issues with models \cite{Hooker2020}.

\section{Conclusion}
In this paper, we started with demonstrating the rapid growth in Deep Learning models, and motivating the fact that someone training and deploying models today has to make either implicit or explicit decisions about efficiency. However, the landscape of model efficiency is vast. 

To help with this, we laid out a mental model for the readers to wrap their heads around the multiple focus areas of model efficiency and optimization. The surveys of the core model optimization techniques give the reader an opportunity to understand the state-of-the-art, apply these techniques in the modelling process, and/or use them as a starting point for exploration. The infrastructure section also lays out the software libraries and hardware which make training and inference of efficient models possible. 

Finally, we presented a section of explicit and actionable insights supplemented by code, for a practitioner to use as a guide in this space. This section will hopefully give concrete and actionable takeaways, as well as tradeoffs to think about when optimizing a model for training and deployment. To conclude, we feel that with this survey we have equipped the reader with the necessary understanding to break-down the steps required to go from a sub-optimal model to one that meets their constraints for both quality as well as footprint.

\section{Acknowledgements}
We would like to thank the Learn2Compress team at Google Research for their support with this work. We would also like to thank Akanksha Saran and Aditya Sarawgi for their help with proof-reading and suggestions for improving the content.

\bibliographystyle{ACM-Reference-Format}
\small
\bibliography{paper}

\end{document}